\documentclass[review]{elsarticle} 

\usepackage{lineno,hyperref}
\modulolinenumbers[5]

\usepackage{times}
\usepackage{epsfig}
\usepackage{graphicx}
\usepackage{amsmath}
\usepackage{amssymb}
\usepackage{amsthm}
\usepackage{wrapfig, balance}

\usepackage[american]{babel}
\usepackage{microtype}
\usepackage{balance}
\usepackage{url}
\usepackage{color}
\usepackage{color}
\usepackage{hyperref}
\hypersetup{
colorlinks=true,
linkcolor=red,
filecolor=magenta,      
urlcolor=black,
}
\usepackage{slashbox}
\usepackage{multirow}
\usepackage{enumitem}

\newcommand{\etal}{{\it et al. }}

\newcommand{\wrt}{{\it w.r.t. }}

\DeclareMathOperator*{\argmin}{arg\,min}

\newtheorem{defi}{Definition}
\newtheorem{thm}{Theorem}
\newtheorem{cor}{Corollary}

\usepackage{algorithm}
\usepackage{algorithmic}
\usepackage[ruled,vlined,algo2e]{algorithm2e}
\usepackage{slashbox}

\journal{Journal of \LaTeX\ Templates}

\begin{document}

\begin{frontmatter}
	
\title{Training Deep Neural Networks via Branch-and-Bound}


\author[address1]{Yuanwei Wu\corref{cor1}}
\ead{wuyuanwei2010@gmail.com}

\author[address2]{Ziming Zhang}
\ead{zzhang15@wpi.edu}

\author[address3]{Guanghui Wang}
\ead{wangcs@ryerson.ca}

\cortext[cor1]{Corresponding author}

\address[address1]{Department of Electrical Engineering \& Computer Science, University of Kansas, Lawrence, KS, USA}
\address[address2]{Department of Electrical \& Computer Engineering, Worcester Polytechnic Institute, Worcester, MA, USA}
\address[address3]{Department of Computer Science, Ryerson University, Toronto, ON, CA}

\begin{abstract}
In this paper, we propose BPGrad, a novel approximate algorithm for deep nueral network training, based on adaptive estimates of feasible region via branch-and-bound. The method is based on the assumption of Lipschitz continuity in objective function, and as a result, it can adaptively determine the step size for the current gradient given the history of previous updates. We prove that, by repeating such a branch-and-pruning procedure, it can achieve the optimal solution within finite iterations. A computationally efficient solver based on BPGrad has been proposed to train the deep neural networks. Empirical results demonstrate that BPGrad solver works well in practice and compares favorably to other stochastic optimization methods in the tasks of object recognition, detection, and segmentation. The code is available at \url{https://github.com/RyanCV/BPGrad}. 
\end{abstract}

\begin{keyword}
Branch and pruning, adaptive solver, deep neural network training, nonconvex optimization, deep learning
\end{keyword}
	
\end{frontmatter}


\section{Introduction}
Deep learning (DL) dramatically improved the state-of-the-art performance in segmentation \cite{he2021sosd,wu2019unsupervised_R6D}, classification \cite{cen2021deep,wu2019unsupervised_RLQ}, object detection \cite{wu2020recent,ma2020mdfn_pr}, 
image generation~\cite{xu2021drb}, speech recognition~\cite{hinton2012deep}, and natural language processing~\cite{devlin2018bert}. It is well known that the empirical success of DL stems mainly from better network architectures~\cite{he2016deep,Zhang_2020_WACV}, the availability of massive dataset like ImageNet~\cite{imagenet_cvpr09}, and increasing computation power of GPUs.

However, the reasons for such huge success of DL still keep elusive theoretically. Researchers start to understand DL from the perspective of optimization such as the optimality of learned models~\cite{haeffele2017global,yun2018global} recently. It has been proved that under certain (very restrictive) conditions, the critical points in DL can actually achieve global optimality, even though its objective is highly nonconvex. Such theoretical results may partially explain why such deep models work well in practical and broad applications.  

\begin{figure}
	\begin{center}
		\includegraphics[width=.65\linewidth]{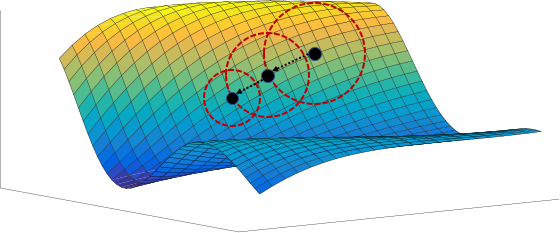}
		\caption{Illustration of the workflow of BPGrad, where each black dot denotes the solution at each iteration (\textit{i.e.,} branch), each directed dotted line denotes the current gradient, and each red dotted circle denotes the region wherein there should be no solutions achieving global optimality (\textit{i.e.,} pruning). BPGrad can automatically estimate the scales of these regions based on the function evaluation and the Lipschitz condition.}
		\label{fig:BPGrad_demo}
	\end{center}
\end{figure}

Global optimality is always desirable and preferred for optimization, which helps the generalization of learned models. In recent years, there are a substantial amount of work focusing on the theoretical analysis of the relations between global optimality and generalization in deep learning, such as~\cite{haeffele2017global,yun2018global,lin2018resnet,liang2018adding,du2018gradient,zou2018stochastic,allen2018learning,allen2018convergence,zhou2018sgd}. All these papers above indicate that global optimality in deep learning improves the generalization.

From the algorithmic perspective, however, locating global optimality in DL is extremely challenging due to its high dimensionality and non-convexity. To our best knowledge, currently there are no DL solvers intentionally developed for this purpose, including stochastic gradient descent (SGD)~\cite{bottou2016optimization}, Adagrad~\cite{duchi2011adaptive}, Adadelta~\cite{zeiler2012adadelta}, RMSProp~\cite{Tieleman2012} and Adam~\cite{kingma2014adam}. Instead, regularization is often used to smooth the objective in DL so that the solvers can converge to some geometrically wider and flatter regions in the parameter space where good model solutions may exist~\cite{zhang2015deep,chaudhari2016entropy,zhang2017Convergent}. These solutions, however, may not necessarily be the global optimum.  
\begin{figure}
	\begin{center}
		\includegraphics[width=0.41\linewidth]{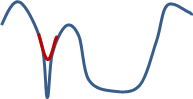} 
		\caption{Illustration of Lipschitz continuity as regularization (red) to smooth a function (blue).}
		\label{fig:Lipschitz}
	\end{center} 
	
\end{figure}

Inspired by the techniques of global optimization for nonconvex functions, we propose a novel approximation algorithm, {\em BPGrad}, which aims to locate the global optimality in DL via branch and pruning (BP)~\cite{sotiropoulos2001branch}. BP is a well-known algorithm developed to search for global solutions for nonconvex optimization problems. Its basic idea is to effectively and gradually shrink the gap between the lower and upper bounds of the global optimum by efficiently branching and pruning the parameter space. Fig.~\ref{fig:BPGrad_demo} illustrates the optimization procedure in BPGrad algorithm.  

In order to branch and prune the parameter space, we assume that the objective functions in DL are Lipschitz continuous~\cite{erikssonapplied} or can be approximated by Lipschitz functions, a fairly weak constraint as it always holds in DL~\cite{goodfellow2016deep}. In fact, the Lipschitz condition provides us a natural way to estimate the lower bound in BP for locating the global optimum (see Sec.~\ref{subsec:Algorithm}). It turns out as well that the Lipschitz condition can serve as regularization if needed, as illustrated in Fig.~\ref{fig:Lipschitz}, to improve the generalization as demonstrated in \cite{chaudhari2016entropy}. {\em In this sense, our BPGrad algorithm/solver essentially aims to locate global optimality in the smoothed objective functions for DL.}

From the perspective of optimization, our algorithm shares similarities with the work~\cite{MalherbeICML17} on global optimization of general Lipschitz functions (not specifically for DL). In~\cite{MalherbeICML17} a uniform sampler is utilized to maximize the lower bound of the maximizer (equivalently minimizing the upper bound of the minimizer) subject to the Lipschitz condition. Convergence properties {\em w.h.p.}~are derived. In contrast, our approach considers estimating both the lower and upper bounds of the global optimum, and employs the gradients as guidance to effectively sample the parameter space for pruning. Theoretical analysis and experiments show that our algorithm can converge within finite iterations.

From the empirical solver perspective, our solver shares similarities with the work~\cite{koushik2016improving} on improving SGD using the feedback from the objective. Specifically,~\cite{koushik2016improving} tracks the relative changes in the objective with a running average, and uses it to adaptively tune the learning rate in SGD. No theoretical analysis, however, is provided for justification. In contrast, our solver does use the feedback from the objective function to determine the learning rate adaptively but based on the rescaled distance between the feedback and the current lower bound estimation. Both theoretical and empirical justifications are established in our work.

In summary, the contributions of this paper are twofold: (1) We propose a novel approximation algorithm with branch and bounding towards global optimization in deep learning application, and a theoretical proof is proposed to show that our BPGrad algorithm has the capability of converging to global optimality within finite iterations; (2) Our solver BPGrad achieves competing results over the state-of-the-art adaptive solvers on the tasks of object recognition, detection, and segmentation. A preliminary result of this study was published previously~\cite{Zhang_2018_CVPR}. This paper is a substantial extension with more technical details and experimental results.

\section{Related Work}\label{sec:related_work} 
\subsection{Deep Learning Solvers} 
SGD is one of the most widely used solvers for object recognition~\cite{krizhevsky2012imagenet,simonyan2014very,he2016deep}, object detection~\cite{girshick2014rich,ren2015faster,he2017mask}, and object segmentation~\cite{long2015fully}.

In general, SGD suffers from slow convergence, and thus its learning rate needs to be carefully tuned. To improve the efficiency of SGD, several DL solvers with adaptive learning rates have been proposed, including Momentum~\cite{qian1999momentum}, Adagrad~\cite{duchi2011adaptive}, Adadelta~\cite{zeiler2012adadelta}, RMSProp~\cite{Tieleman2012} and Adam~\cite{kingma2014adam}. As stated in \cite{ruder2016overview}, these solvers are able to escape the saddle points and often yield faster convergence empirically by integrating the advantages from both stochastic and batch methods where small mini-batches are used to adopt historical gradient information to automatically adjust the learning rate. 

Adagrad is well suited to deal with sparse data, as it adapts the learning rate to the parameters, performing smaller updates on frequent parameters and larger updates on infrequent parameters. However, it suffers from shrinking on the learning rate, which motivates Adadelta, RMSProp and Adam. Adadelta accumulates squared gradients to be fixed values rather than over time in Adagrad, RMSProp updates the parameters based on the rescaled gradients, and Adam does the same based on the estimated mean and variance of the gradients. Mukkamala and Hein in~\cite{mukkamala2017variants} proposed variant solvers of RMSProp and Adagrad with logarithmic regret bounds. Berrada~\etal proposed deep frank-wolfe for neural network optimization~\cite{berrada2019deep} and introduced an adaptive learning-rate optimization algorithm in the interpolation setting~\cite{berrada2020training}. Readers may refer to~\cite{ruder2016overview} for a comprehensive review on the gradient descent based optimization algorithms.

\subsection{Global Optimality in Deep Learning} 
The empirical loss minimization problem in DL is high-dimensional and nonconvex with potentially numerous local minima and saddle points. Earlier work on training neural networks~\cite{blum1989training} showed that it is difficult to find the global optima because in the worst case even learning a simple 3-node neural network is NP-complete. In spite of the challenges in training deep models, researchers have attempted to provide empirical as well as theoretical justification for the success of these models \textit{w.r.t.} global optimality in learning~\cite{zhang2016understanding, lin2018resnet,liang2018adding, du2018gradient, zou2018stochastic, allen2018learning, allen2018convergence, brutzkus2017globally, nguyen2017loss, yun2018global, haeffele2017global}. Several recent works have also studied on how to overcome poor local optima using the global loss structures~\cite{freeman2016topology, kawaguchi2016deep, choromanska2015loss, dauphin2014identifying, pmlr-v49-lee16, hand2017global, li2017visualizing}. Some other works explore the local structures of minima to study the differences between sharp and wide local minima during training found by SGD and its variants~\cite{hochreiter1997flat, keskar2016large, keskar2016large, pmlr-v70-dinh17b, soudry2016no, draxler2018essentially, chaudhari2016entropy}.

\subsection{Branch, Bound and Pruning}
Branch-and-bound (B\&B)~\cite{land1960automatic} is one of the promising methods for global optimization in nonconvex problems. The basic idea of B\&B is to recursively divide the feasible set of a problem into disjoint subsets (``branching"), where each node represents a subproblem that only conducts searches on the subset of that node. The key idea is to keep the track of bounds on the minimum, and use these bounds to ``prune" the search space, removing candidate solutions that cannot be optimal provably. To our best knowledge, currently no DL solvers are developed based on B\&B, while ours is.

\section{BPGrad Algorithm for Deep Learning}\label{sec:BPGrad_algo}
\subsection{Notation} 
Let $\mathcal{X}\subseteq\mathbb{R}^d$ be the parameters space, $\mathbf{x}\in\mathcal{X}$ be the parameters of a given neural network, and $(\omega, y)\in \Omega \times \mathcal{Y}$ be a pair of a data sample $\omega$ and its associated label $y$. Let $\phi: \Omega\times \mathcal{X} \rightarrow \mathcal{Y}$ denote the nonconvex mapping function defined by the network, and $f$ be the objective function with Lipschitz constant $L\geq0$ to train the network. For all $\mathbf{x}=(x_1,\cdots,x_d)\in\mathbb{R}^d$, let $\|\mathbf{x}\|_2=(\sum_{i=1}^{d}x_i^2)^{1/2}$ denote the standard $\ell_2$-norm, $\nabla f$ be the gradient of $f$ over parameters $\mathbf{x}$\footnote{We assume $\nabla f\neq \mathbf{0}$ {\em w.l.o.g.}, and empirically we can randomly sample a non-zero direction for update wherever $\nabla f=\mathbf{0}$.}, $\nabla \tilde{f}=\frac{\nabla f}{\|\nabla f\|_2}$ be the {\em normalized} gradient (i.e., the direction of the gradient), and $f^\ast$ be the global minimum.

\subsection{Problem Statement}
Given a deep network, the task of training process is to learn the parameters by minimizing the following objective function $f$:
\begin{align}\label{eqn:f}
\min_{\mathbf{x}\in\mathcal{X}} f(\mathbf{x}) \equiv \mathbb{E}_{(\omega\times y)\in\Omega\times\mathcal{Y}}\Big[\mathcal{L}(y, \phi(\omega, \mathbf{x}))\Big] + \mathcal{R}(\mathbf{x}),
\end{align}
where $\mathbb{E}$ is the expectation over data pairs, $\mathcal{L}$ is the loss function (\textit{e.g.,} cross entropy loss) to evaluate the differences between the ground-truth labels and the predicted labels of given data samples, and $\mathcal{R}$ is a form of regularization over parameters designed to prevent overfitting (\textit{e.g.,} weight decay via $\ell_2$ regularization). We make assumptions throughout the paper as follows.

\begin{enumerate}
	\item[{\em F1.}] $f$ is lower-bounded by 0, {\it i.e.} $f(\mathbf{x})\geq0, \forall \mathbf{x}\in\mathcal{X}$;
	\item[{\em F2.}] $f$ is differentiable for every $\mathbf{x}\in\mathcal{X}$;
	\item[{\em F3.}] $f$ is Lipschitz continuous, or can be approximated by Lipschitz functions, with constant $L\geq 0$.
\end{enumerate}

\subsection{Algorithm} \label{subsec:Algorithm}
Our BPGrad algorithm relies on the following assumption:
\begin{defi}[Lipschitz Continuity~\cite{erikssonapplied}]\label{def:Lipschitz}
	A function f : $\mathbb{R}^m \rightarrow \mathbb{R}$ is {\em Lipschitz continuous} if there exists a Lipschitz constant $L\geq0$ such that 
	\begin{align}\label{eqn:LF}
	|f(\mathbf{x}_1)-f(\mathbf{x}_2)|\leq L\|\mathbf{x}_1-\mathbf{x}_2\|_2, \forall \mathbf{x}_1 , \mathbf{x}_2 \in \mathcal{X}.
	\end{align}
\end{defi}

\subsubsection{Lower and Upper Bound Estimation}\label{sssec:lub}
Consider the situation where samples $\mathbf{x}_1, \cdots, \mathbf{x}_t\in\mathcal{X}$ exist for evaluation by function $f$ with Lipschitz constant $L$, whose global minimum $f^\ast$ is reached by the sample $\mathbf{x}^\ast$. 
Then based on Eq.~(\ref{eqn:LF}) and simple algebra, we can obtain
\begin{align}\label{eqn:LUB_f}
\max_{i=1,\cdots,t}\Big\{f(\mathbf{x}_i)-L\|\mathbf{x}_i-\mathbf{x}^\ast\|_2\Big\}\leq f^\ast \leq \min_{i=1,\cdots,t}f(\mathbf{x}_i).
\end{align}

This provides us both the lower and upper bounds of the global minimum. The upper bound is tractable, however, the lower bound is {\em intractable}. The intractability comes from the fact that the optimal sample $\mathbf{x}^\ast$ is unknown, and thus makes the lower bound in Eq.~(\ref{eqn:LUB_f}) empirically unusable. To address this problem, we propose a novel tractable estimator, $\rho\min_{i=1,\cdots,t}f(\mathbf{x}_i)$ $(0\leq \rho <1)$, for the lower bound. This estimator intentionally introduces a gap from the upper bound, which will be reduced by either decreasing the upper bound or increasing $\rho$. As proved in Thm.~\ref{thm:f} (see Sec.~\ref{ssec:analysis}), when the parameter space $\mathcal{X}$ is fully covered by the samples $\{\mathbf{x}_i\}$, this estimator will become the lower bound of $f^\ast$.

In summary, we define our lower and upper bound estimators for the global minimum as $\rho\min_{i=1,\cdots,t}f(\mathbf{x}_i)$ and $\min_{i=1,\cdots,t}f(\mathbf{x}_i)$, respectively.


\subsubsection{Branch and Pruning} \label{subsub:B&P}
Based on our estimators, we propose a novel approximation algorithm, called BPGrad, towards global optimization in DL via branch and pruning. The implementation of the algorithm is shown in Alg.~\ref{alg:BnB}, where the predefined constant $\epsilon\geq0$ controls the precision of the solution with a default value of $\epsilon=10^{-4}$. It determines how many iterations are required to satisfy the precision during the optimization. 

\noindent
{\bf Branch:} 
The {\em inner} loop in Alg.~\ref{alg:BnB} conducts the branch operation to split the parameter space recursively by {\em sampling}. To this end, we need a mapping between the parameter space and the bounds. Considering the lower bound in Eq.~(\ref{eqn:LUB_f}), we propose sampling $\mathbf{x}_{t+1}\in\mathcal{X}$ based on the previous samples $\mathbf{x}_1,\cdots,\mathbf{x}_t\in\mathcal{X}$ so that it satisfies

\begin{align}\label{eqn:sampling_rule}
\hspace{-2mm}\max_{i=1,\cdots,t}\Big\{f(\mathbf{x}_i)-L\|\mathbf{x}_i-\mathbf{x}_{t+1}\|_2\Big\}\leq\rho\min_{i=1,\cdots,t}f(\mathbf{x}_i).
\end{align}  

Note that an equivalent constraint has been used in~\cite{MalherbeICML17}. To improve the efficiency of sampling while keep decreasing the objective, we propose a strategy of sampling along the directions of (stochastic) gradients with small distortion. Though gradients only encode local structures of (nonconvex) functions in a high dimensional space, they are good indicators for locating local minima~\cite{pmlr-v49-lee16, panageas2016gradient}. Specifically, we formulate it as a minimization problem to generate samples from the parameter space:
\begin{align}\label{eqn:GD-sampler}
&\min_{\mathbf{x}_{t+1}\in\mathcal{X}, \eta_t\geq0}\left\|\mathbf{x}_{t+1} - \left(\mathbf{x}_t - \eta_t\nabla\tilde{f}(\mathbf{x}_t)\right)\right\|_2^2 + \gamma\eta_t^2, \\
&\mbox{s.t.} \; \max_{i=1,\cdots,t}\Big\{f(\mathbf{x}_i)-L\|\mathbf{x}_i-\mathbf{x}_{t+1}\|_2\Big\}\leq\rho\min_{i=1,\cdots,t}f(\mathbf{x}_i), \nonumber
\end{align}
where $\gamma\geq0$ is a predefined constant controlling the trade-off between the distortion and the step size $\eta_t\geq0$. That is, under the condition in Eq.~(\ref{eqn:sampling_rule}), the objective in Eq.~(\ref{eqn:GD-sampler}) aims to generate a sample that has small distortion from an anchor point, whose step size is small and, due to the locality property of gradients, along the direction of the gradient.

Note that other reasonable objective functions may also be utilized here for the sampling purpose as long as the condition in Eq.~(\ref{eqn:sampling_rule}) is satisfied. More efficient sampling objectives will be investigated in our future work. 

\noindent
{\bf Pruning:} 
In fact, Eq.~(\ref{eqn:sampling_rule}) specifies that new samples should be generated outside the union of a set of balls defined by previous samples. To precisely describe this requirement, we introduce a new concept of removable solution space as bellow.

\begin{defi}[Removable Parameter Space (RPS)]\label{def:RPS}
	We define the RPS, denoted as $\mathcal{X}_{R}$, as
	\begin{align}\label{eqn:ball}
	\mathcal{X}_R(t) \stackrel{\mbox{\em def}}{=} \cup_{j=1,\cdots,t} \mathcal{B}\left(\mathbf{x}_j, r_j\right),
	\end{align}
	where $\mathcal{B}(\mathbf{x}_j, r_j)=\{\mathbf{x}\mid \|\mathbf{x} - \mathbf{x}_j\|_2 < r_j, \mathbf{x}\in\mathcal{X}\}, \forall j$ defines a ball centered at sample $\mathbf{x}_j\in\mathcal{X}$ with radius $r_j=\frac{1}{L}\left[f(\mathbf{x}_j)-\rho\min_{i=1,\cdots,t}f(\mathbf{x}_i)\right], \forall j$.
\end{defi}

\begin{algorithm}[t]
	\SetAlgoLined
	\SetKwInOut{Input}{Input}\SetKwInOut{Output}{Output}
	\Input{objective function $f$ with Lipschitz constant $L\geq0$, precision $\epsilon\geq 0$}
	\Output{minimizer $\mathbf{x}^\ast$}
	\BlankLine
	Randomly initialize $\mathbf{x}_1$, $t\leftarrow 1$, $\rho\leftarrow 0$;
	
	\While{$\min_{i=1,\cdots,t}f(\mathbf{x}_i)\geq \frac{\epsilon}{1-\rho}$}{
		\While{$\exists \mathbf{x}_{t+1}\in\mathcal{X}$ satisfies Eq.~(\ref{eqn:sampling_rule})}{
			Compute $\mathbf{x}_{t+1}$ by solving Eq.~(\ref{eqn:GD-sampler});
			
			$t\leftarrow t+1$;
		}
		Increase $\rho$ such that $0\leq\rho<1$ still holds;
	}
	\Return $\mathbf{x}^\ast=\mathbf{x}_{i^\ast}$ where $i^\ast\in\argmin_{i=1,\cdots,t} f(\mathbf{x}_i)$;
	\caption{General BPGrad Algorithm}
	\label{alg:BnB}
\end{algorithm}

RPS specifies a region wherein the function evaluations of all the points cannot be smaller than the lower bound estimator conditioning on the Lipschitz continuity assumption. Therefore, when the lower bound estimator is higher than the global minimum $f^\ast$, we can safely remove all the points in RPS without evaluation. Parameter $\rho$ controls such confidence or tolerance. However, when it becomes smaller than $f^\ast$, there is a risk of missing the global solution. To address this issue, we propose the {\em outer} loop in Alg.~\ref{alg:BnB} to increase the lower bound, so as to draw more samples which may further decrease the upper bound later.

The implementation of Alg.~\ref{alg:BnB} involves a sequential procedure of branching and pruning, which starts at an initial point $x_1$ by evaluating the function $f(x_1)$, calculating the radius $r_1=\frac{1}{L}[f(x_1)-\rho \min_{i=1}f(x_1)]$, then at each step $t\geq1$ to draw a new sample $\mathbf{x}_{t+1}\sim\mathcal{X}\setminus\mathcal{X}_R(t)$ which depends on the previous evaluations $\{(x_j, r_j, f(x_j))\}_{j=1,\cdots,t}$, and finally evaluate the objective function $f(x_{t+1})$ at this point. To illustrate its effectiveness in an interpretable domain, we have applied the Alg.~\ref{alg:BnB} to solve a problem of controllable complexity in Sec.~\ref{subsub:demo_Alg1}.

\begin{figure}
	\begin{center}
		\includegraphics[width=0.9\linewidth]{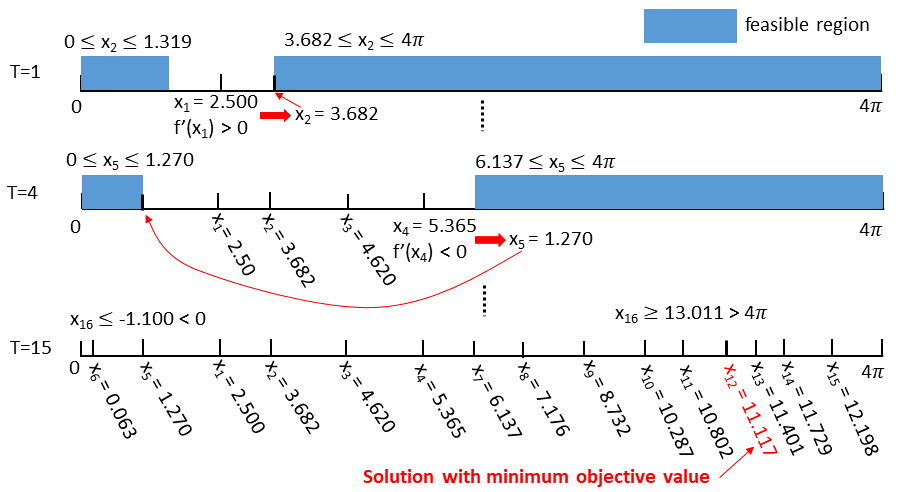}
		\caption{Illustration of BPGrad from Alg.~\ref{alg:BnB} in $1D$ space.}
		\label{fig:illustration_BPGrad_Alg1} 
	\end{center}
\end{figure}

\subsubsection{Illustration of Alg.~\ref{alg:BnB} in One-Dimensional Space} \label{subsub:demo_Alg1}

In Fig.~\ref{fig:illustration_BPGrad_Alg1}, we show an example of using Alg.~\ref{alg:BnB} to solve the following nonconvex function:
\begin{align}\label{equ:f3_1d_demo_BPGrad_Alg1}
f(x)=x\sin(x) + 15, \forall x\in[0, 4\pi].
\end{align}

This function has a local and a global minimum at $x_{lmin}=4.813$ and $x_{gmin}=11.086$, respectively. The Lipschitz constant and initial point (at $T=1$) are set to $L=4\pi$ and $x_1=2.5$. Given $x_1$, using Alg. \ref{alg:BnB} we can obtain two feasible sets for $x_2$ as shown in Fig.~\ref{fig:illustration_BPGrad_Alg1}. In branching, due to the gradient $f'(x_1)>0$, we select the solution in the right set, leading to $x_2=3.682$. Then the infeasible set is pruned from the parameter space. Similarly, at iteration $T=4$, since the gradient $f'(x_4)<0$, the solution is in the left set with $x_5=1.27$. By alternating the procedures of branching and pruning, we eventually have searched all the parameter space within $16$ iterations, and found the global minimum at $x_{12}=11.117$. The error of this solution \wrt the ground-truth is $0.031$. 

\subsection{Theoretical Analysis}\label{ssec:analysis}
\begin{thm}[Lower \& Upper Bounds]\label{thm:f}
	Whenever \\ $\mathcal{X}_R(t)\equiv\mathcal{X}$ holds, the samples generated by Alg.~\ref{alg:BnB} satisfies
	\begin{align}\label{eqn:rho}
	\rho\min_{i=1,\cdots,t}f(\mathbf{x}_i) \leq f^\ast \leq \min_{i=1,\cdots,t}f(\mathbf{x}_i).
	\end{align}
\end{thm}

\begin{proof}
	Since $f^\ast$ is the global minimum, it always holds that $f^\ast \leq \min_{i=1,\cdots,t}f(\mathbf{x}_i)$. When $\mathcal{X}_R(t)\equiv\mathcal{X}$, suppose $\rho\min_{i=1,\cdots,T}f(\mathbf{x}_i) > f^\ast$ holds, then there would exist at least one point (i.e.\ global minimum) left for sampling, contradicting to the condition of $\mathcal{X}_R(t)\equiv\mathcal{X}$. We then complete the proof.
\end{proof}

\begin{cor}[Approximation Error Bound]
	Given that both $\min_{i=1,\cdots,t}f(\mathbf{x}_i)\leq \frac{\epsilon}{1-\rho}$ and $\mathcal{X}_R(t)\equiv\mathcal{X}$ hold, it is satisfied that
	\begin{align}
	\min_{i=1,\cdots,t}f(\mathbf{x}_i) - f^\ast \leq \epsilon.
	\end{align}
\end{cor}

\begin{thm}[Convergence within Finite Samples]\label{thm:number}
	The total number of samples, $T$, in Alg.~\ref{alg:BnB} is upper bounded by:
	\begin{align}
	T \leq \left[\frac{2L}{(1-\rho)f_{\min}}\right]^d \cdot \frac{V_{\mathcal{X}}}{C},
	\end{align}
	where $V_{\mathcal{X}}$ denotes the volume of the space $\mathcal{X}$, $C=\frac{\pi^{\frac{d}{2}}}{\Gamma\left(\frac{d}{2}+1\right)}$ denotes a constant, and $f_{\min} = \min_{i=1,\cdots,T}f(\mathbf{x}_i)$ denotes the minimum evaluation.
\end{thm}
\begin{proof} 
	Given $\forall j, \forall t$ such that $1\leq j\leq t\leq T-1$, we have
	\begin{align}
	\hspace{-5mm}\|\mathbf{x}_{t+1}-\mathbf{x}_j\|_2 & \geq \frac{1}{L}\left[f(\mathbf{x}_j) - \rho\min_{i=1,\cdots,t}f(\mathbf{x}_i)\right]  \\
	\hspace{-5mm} & \geq \frac{1-\rho}{L}\cdot\min_{i=1,\cdots,t}f(\mathbf{x}_i) \geq \frac{(1-\rho)f_{\min}}{L}. \nonumber
	\end{align}
	This allows us to generate two balls $\mathcal{B}\left(\mathbf{x}_{t+1}, \frac{(1-\rho)f_{\min}}{2L}\right)$ and $\mathcal{B}\left(\mathbf{x}_j, \frac{(1-\rho)f_{\min}}{2L}\right)$ so that they have no overlap with each other. As a result, we can generate $T$ balls with radius of $\frac{(1-\rho)f_{\min}}{2L}$ and no overlaps, and their accumulated volume should be no bigger than $V_{\mathcal{X}}$, \textit{i.e.,}
	\begin{align}
	V_{\mathcal{X}} \geq \sum_{t=1}^T V_{\mathcal{B}\left(\mathbf{x}_t, \frac{(1-\rho)f_{\min}}{2L}\right)} = C\left[\frac{(1-\rho)f_{\min}}{2L}\right]^d T.
	\end{align}
	Further using simple algebra we can complete the proof.
\end{proof}

\begin{algorithm}[t]
	\SetAlgoLined
	\SetKwInOut{Input}{Input}\SetKwInOut{Output}{Output}
	\Input{number of samples $T$, objective function $f$ with Lipschitz constant $L\geq0$, momentum $0\leq\mu\leq 1$, parameter $\rho\geq0$}
	\Output{minimizer $\mathbf{x}^\ast$}
	\BlankLine
	$\mathbf{v}_1\leftarrow\mathbf{0}$, and randomly initialize $\mathbf{x}_1$;
	
	\For{$t\leftarrow 1$ \KwTo $T-1$}{
		
		$\mathbf{v}_{t+1}\leftarrow\mu\mathbf{v}_t - \frac{f(\mathbf{x}_t) - \rho\min_{i=1,\cdots,t}f(\mathbf{x}_i)}{L}\cdot\frac{\nabla f(\mathbf{x}_t)}{\|\nabla f(\mathbf{x}_t)\|_2}$;
		
		$\mathbf{x}_{t+1}\leftarrow\mathbf{x}_{t} + \mathbf{v}_{t+1}$;
		
	}
	\Return $\mathbf{x}^\ast=\mathbf{x}_T$;
	\caption{BPGrad based Solver for Deep Learning}\label{alg:BnB-GD}
\end{algorithm}

\section{Approximate DL Solver based on BPGrad}\label{sec:BPGrad_solver}
Although the BPGrad algorithm has nice theoretical properties for global optimization, we still need to solve the following problems in order to apply the Alg.~\ref{alg:BnB} to deep learning applications.

\begin{enumerate}[noitemsep]
	\item[{\em P1.}] From Thm.~\ref{thm:number} we can see that, due to the high dimensionality of the parameter space in DL, it is impractical to draw sufficient samples to cover the entire space.
	\item[{\em P2.}] Solving Eq.~(\ref{eqn:GD-sampler}) involves the knowledge of previous samples, which incurs a significant amount of computational and storage burden for deep learning.
	\item[{\em P3.}] Computing $f(\mathbf{x}_t)$ and $\nabla\tilde{f}(\mathbf{x}_t), \forall \mathbf{x}_t\in\mathcal{X}$ is time-consuming, especially for large-scale data.
\end{enumerate}

To address {\em P1}, in practice we manually set the maximum numbers of iterations for both inner and outer loops in Alg.~\ref{alg:BnB}.

To address {\em P2}, we further make some extra assumptions to simplify the sampling procedure based on Eq.~(\ref{eqn:GD-sampler}) as follows:

\begin{enumerate}[noitemsep]
	\item[{\em A1.}] Minimizing distortion is more important than minimizing step sizes, i.e.\ $\gamma\ll1$;
	\item[{\em A2.}] $\mathcal{X}$ is sufficiently large where $\exists\eta_t\geq0$ so that $\mathbf{x}_{t+1}=\mathbf{x}_t-\eta_t\nabla\tilde{f}(\mathbf{x}_t)\in\mathcal{X}\setminus\mathcal{X}_R(t)$ always holds;
	\item[{\em A3.}] $\eta_t\geq0$ is always sufficiently small for local update.
	\item[{\em A4.}] $\mathbf{x}_{t+1}$ can be sampled only based on $\mathbf{x}_t$ and $\nabla\tilde{f}(\mathbf{x}_t)$.
\end{enumerate}

By imposing these assumptions on Eq.~(\ref{eqn:GD-sampler}), we can directly compute the solution as follows:
\begin{align}
\eta_t = \frac{1}{L}\left[f(\mathbf{x}_t) - \rho\min_{i=1,\cdots,t}f(\mathbf{x}_i)\right].
\label{eqn:eta_t}
\end{align}

To address {\em P3}, we utilize mini-batches to estimate $f(\mathbf{x}_t)$ and $\nabla\tilde{f}(\mathbf{x}_t)$ efficiently in each iteration.

\begin{figure}[t]
	\begin{minipage}[b]{0.49\columnwidth}
		\begin{center}
			\centerline{\includegraphics[width=\columnwidth]{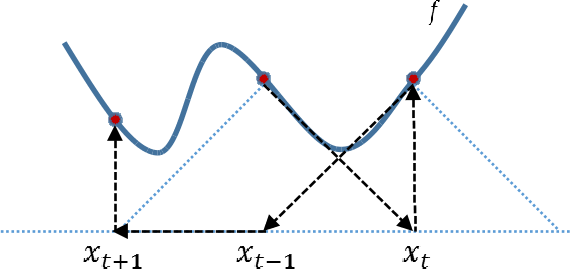}}
			\centerline{(a)	Sampling using Eq.~(\ref{eqn:GD-sampler})}
		\end{center}
	\end{minipage}
	\begin{minipage}[b]{0.49\columnwidth}
		\begin{center}
			\centerline{\includegraphics[width=\columnwidth]{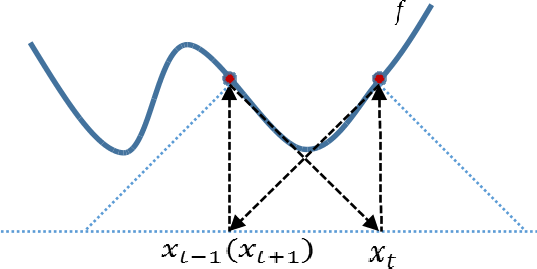}}
			\centerline{(b) Sampling using Eq.~(\ref{eqn:eta_t})}
		\end{center}
	\end{minipage}	 
	\vspace{-8mm}
	\caption{1D illustration of the difference in sampling between {\bf (a)} using Eq.~(\ref{eqn:GD-sampler}) and {\bf (b)} using Eq.~(\ref{eqn:eta_t}). Here the solid blue lines denote function $f$, the black dotted lines denote the sampling paths starting from $\mathbf{x}_{t-1} \rightarrow \mathbf{x}_t \rightarrow \mathbf{x}_{t+1}$, and each triangle surrounded by blue dotted lines denotes the RPS of each sample. It can be seen that (b) suffers from being stuck locally, while (a) can avoid the locality based on the RPS.}
	\label{fig:sample}
\end{figure}

In summary, we present our BPGrad solver in Alg.~\ref{alg:BnB-GD} by modifying Alg.~\ref{alg:BnB} for the sake of fast sampling as well as low memory footprint in training deep models, however, there is a risk of being stuck in local regions. Fig.~\ref{fig:sample} illustrates such a scenario using a $1$D example. In Fig.~\ref{fig:sample} (b) the sampling method falls into a loop because it does not consider the history of samples except for the current one. In contrast, the sampling method in Fig.~\ref{fig:sample} (a) is able to keep generating new samples by avoiding the RPS of previous samples with more computation and storage as expected.

\subsection{Theoretical Analysis}
\begin{thm}[Global Property Preservation]\label{thm:fast_sample}
	Let $\mathbf{x}_{t+1} = \mathbf{x}_t - \eta_t\nabla\tilde{f}(\mathbf{x}_t)$ where $\eta_t$ is computed using Eq.~(\ref{eqn:eta_t}). Then $\mathbf{x}_{t+1}$ satisfies Eq.~(\ref{eqn:sampling_rule}) if it holds that
	\begin{align}\label{eqn:fast_condition}
	\hspace{-3mm}\left\langle\mathbf{x}_i-\mathbf{x}_t, \nabla\tilde{f}(\mathbf{x}_t)\right\rangle \geq \frac{f(\mathbf{x}_i) - f(\mathbf{x}_t)}{L}, \forall i=1,\cdots,t,
	\end{align}
	where $\langle\cdot,\cdot\rangle$ denotes the inner product between two vectors.
\end{thm}
\begin{proof}
	Based on Eq.~(\ref{eqn:LF}), Eq.~(\ref{eqn:eta_t}), and Eq.~(\ref{eqn:fast_condition}), we have
	\begin{align}
	& \|\mathbf{x}_i-\mathbf{x}_{t+1}\|_2 \nonumber \\
	& = \left(\|\mathbf{x}_i-\mathbf{x}_{t}\|_2^2+\eta_t^2+2\eta_t\left\langle\mathbf{x}_i-\mathbf{x}_t, \nabla\tilde{f}(\mathbf{x}_t)\right\rangle\right)^{\frac{1}{2}} \nonumber\\
	&\geq \frac{1}{L}\left[f(\mathbf{x}_i) - \rho\min_{i=1,\cdots,t}f(\mathbf{x}_i)\right], \forall i=1,\cdots,t,
	\end{align}
	which is essentially equivalent to Eq.~(\ref{eqn:sampling_rule}) based on algebra. We then can complete the proof.
\end{proof}

\begin{cor}\label{cor:1}
	Suppose that a monotonically decreasing sequence $\{f(\mathbf{x}_i)\}_{i=1,\cdots,t}$ is generated to minimize function $f$ by sampling using Eq.~(\ref{eqn:eta_t}). Then the condition in Eq.~(\ref{eqn:fast_condition}) can be rewritten as follows:
	\begin{align}\label{eqn:cor}
	\left\langle\mathbf{x}_i-\mathbf{x}_j, \nabla\tilde{f}(\mathbf{x}_j)\right\rangle \geq 0, \, 1\leq \forall i< \forall j\leq t.
	\end{align}
\end{cor}

\noindent
{\bf Discussion:}
Both Thm.~\ref{thm:fast_sample} and Cor.~\ref{cor:1} imply that, roughly speaking, our solver prefers sampling the parameter space along a path towards a single direction. However, the gradients in conventional backpropagation have little guarantee to satisfy Eq.~(\ref{eqn:fast_condition}) or Eq.~(\ref{eqn:cor}) due to lack of such constraints in learning. On the other hand, momentum~\cite{sutskever2013importance} is a well-known technique in deep learning to dampen oscillations in gradients and accelerate directions of low curvature. Therefore, our solver in Alg.~\ref{alg:BnB-GD} involves momentum to compensate such drawbacks in backpropagation for better approximation of Alg.~\ref{alg:BnB}.

\subsection{Empirical Justification}\label{subsec:Empirical_justification}
In this section, we discuss the feasibility of the assumptions {\em A1-A4} in reducing the computational and storage burden as well as preserving the properties towards global optimization in deep learning.

We utilize MATLAB toolbox for CNNs (MatConvNet)~\cite{vedaldi2015matconvnet} as our testbed, and run our solver BPGrad in Alg.~\ref{alg:BnB-GD} to train the default networks in MatConvNet for MNIST~\cite{lecun1998mnist} and CIFAR-$10$~\cite{krizhevsky2012imagenet}, respectively, using the default parameters without explicit mention. Also we set $L=15$ for MNIST and $L=50$ for CIFAR-$10$ by default. For justification purpose we only run 4 epochs on each dataset, 600 and 500 iterations per epoch for MNIST and CIFAR-$10$, respectively. For more experimental details, please refer to Sec.~\ref{sec:exp}.

Essentially the assumption {\em A1} is made to support the other three to simplify the objective in Eq.~(\ref{eqn:GD-sampler}), and the assumption {\em A2} usually holds in deep learning due to its high dimensionality. Therefore, below we only focus on empirical justification of assumptions {\em A3} and {\em A4}.

\begin{figure}[t]
	\begin{center}
		\includegraphics[width=0.45\linewidth]{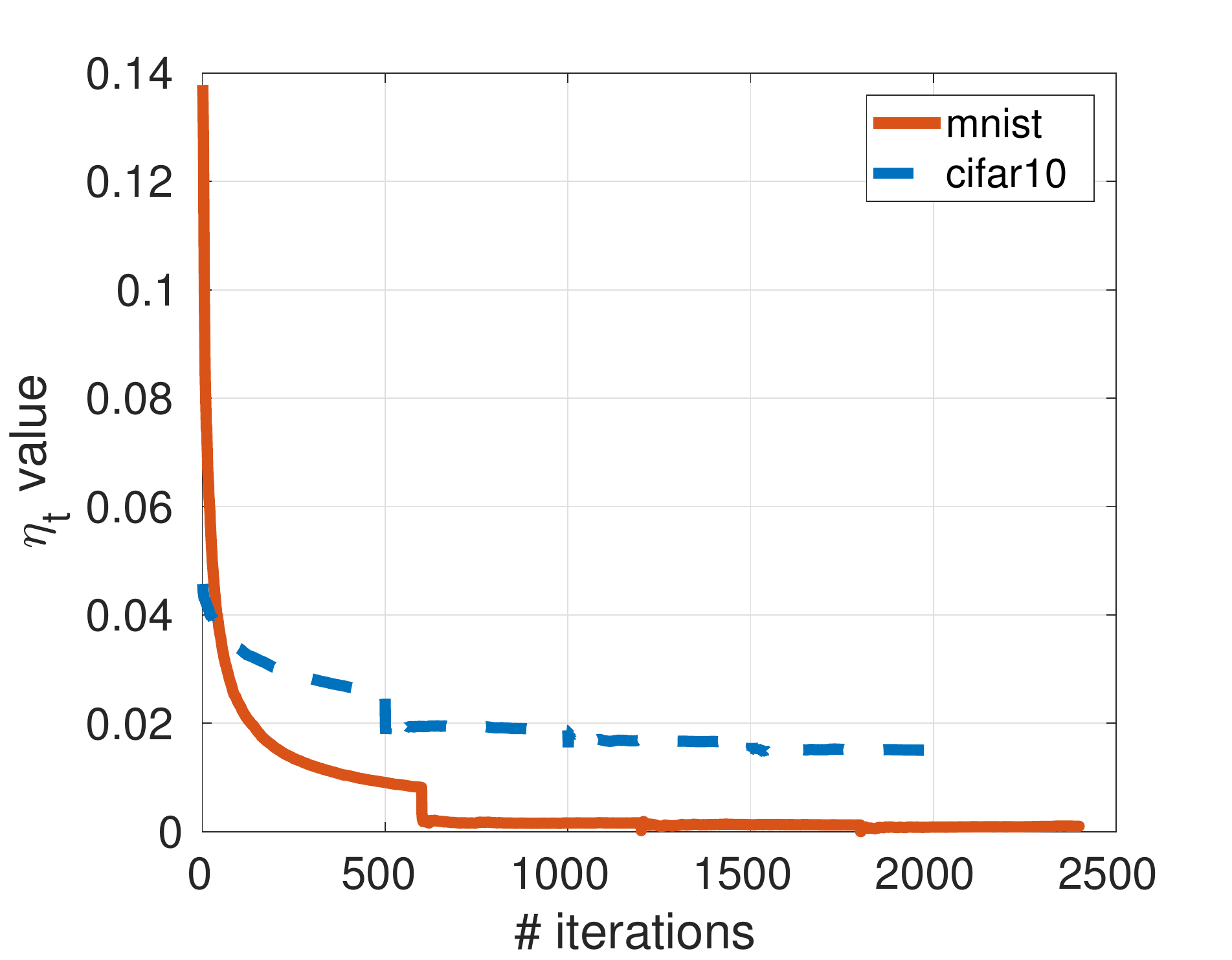}
		\caption{Plots of $\eta_t$ on MNIST and CIFAR-$10$, respectively.}
		\label{fig:eta}
	\end{center} 
\end{figure}

\subsubsection{Feasibility of A3}  
\begin{figure}[t]
	\begin{minipage}[b]{0.49\columnwidth}
		\begin{center}
			\centerline{\includegraphics[width=1.05\columnwidth]{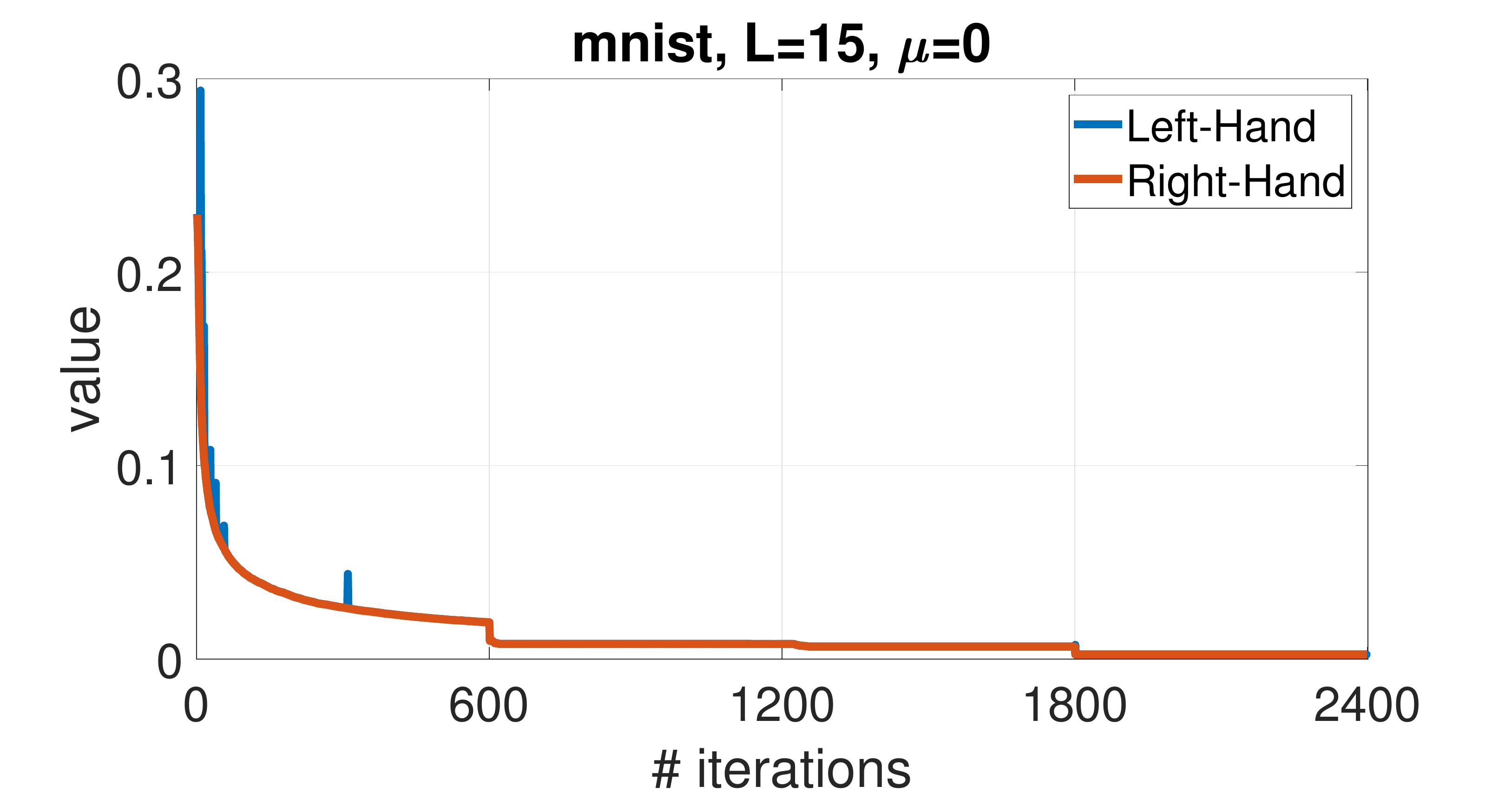}} 
		\end{center}
	\end{minipage} 
	\begin{minipage}[b]{0.49\columnwidth}
		\begin{center}
			\centerline{\includegraphics[width=1.05\columnwidth]{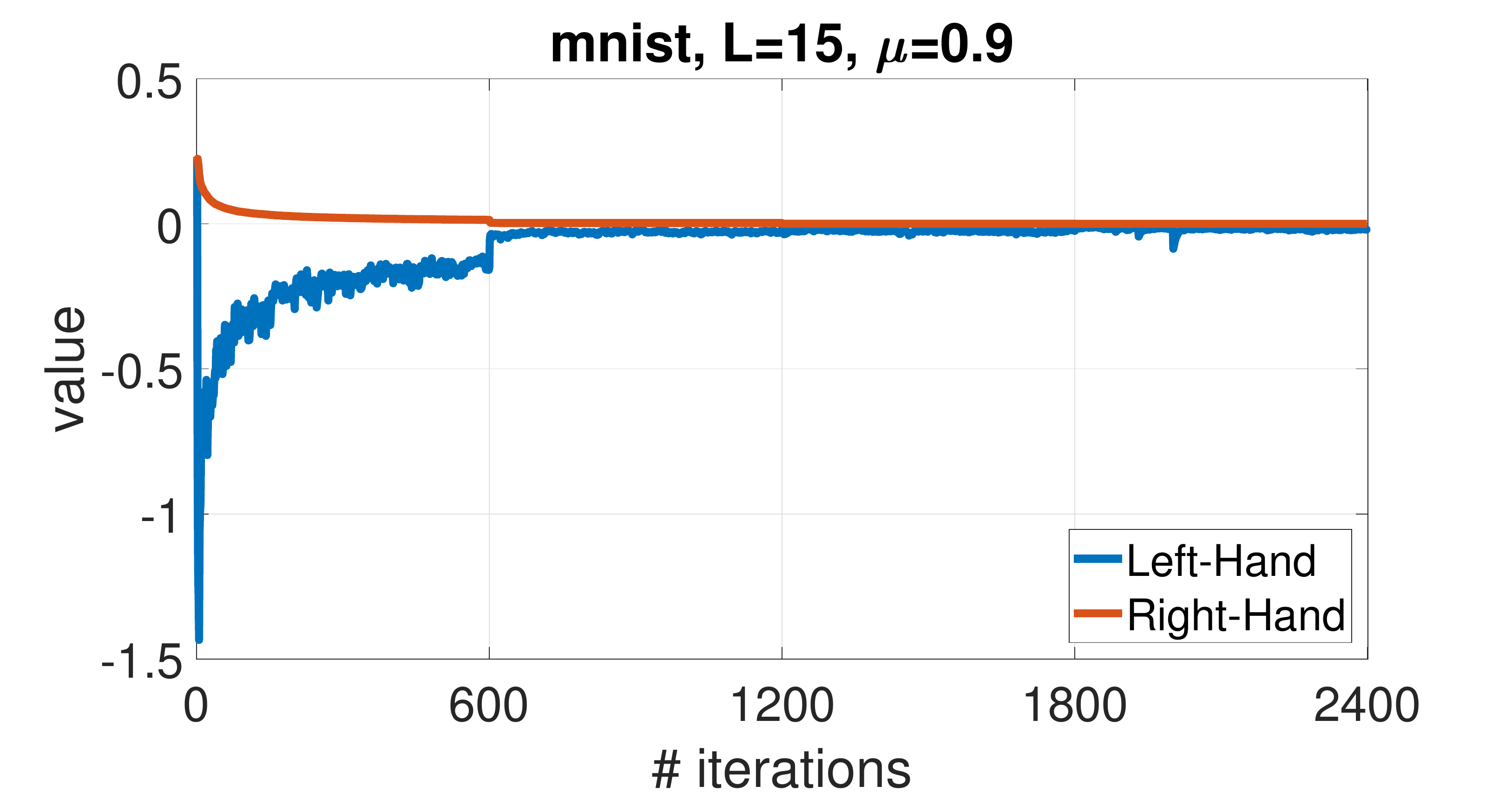}} 
		\end{center}
	\end{minipage} 
	\begin{minipage}[b]{0.49\columnwidth}
		\begin{center}
			\centerline{\includegraphics[width=1.05\columnwidth]{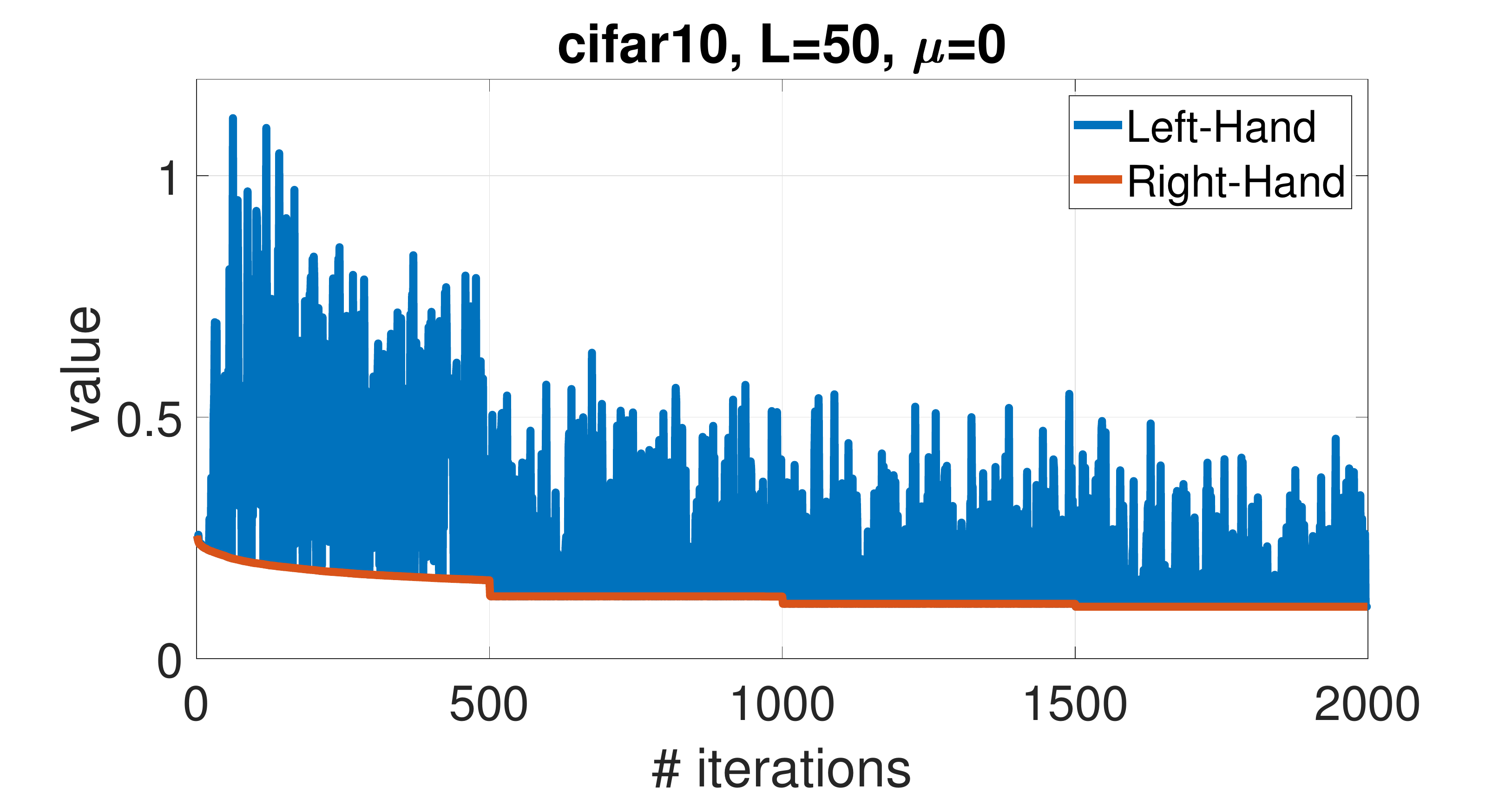}} 
		\end{center}
	\end{minipage} 
	\begin{minipage}[b]{0.49\columnwidth}
		\begin{center}
			\centerline{\includegraphics[width=1.05\columnwidth]{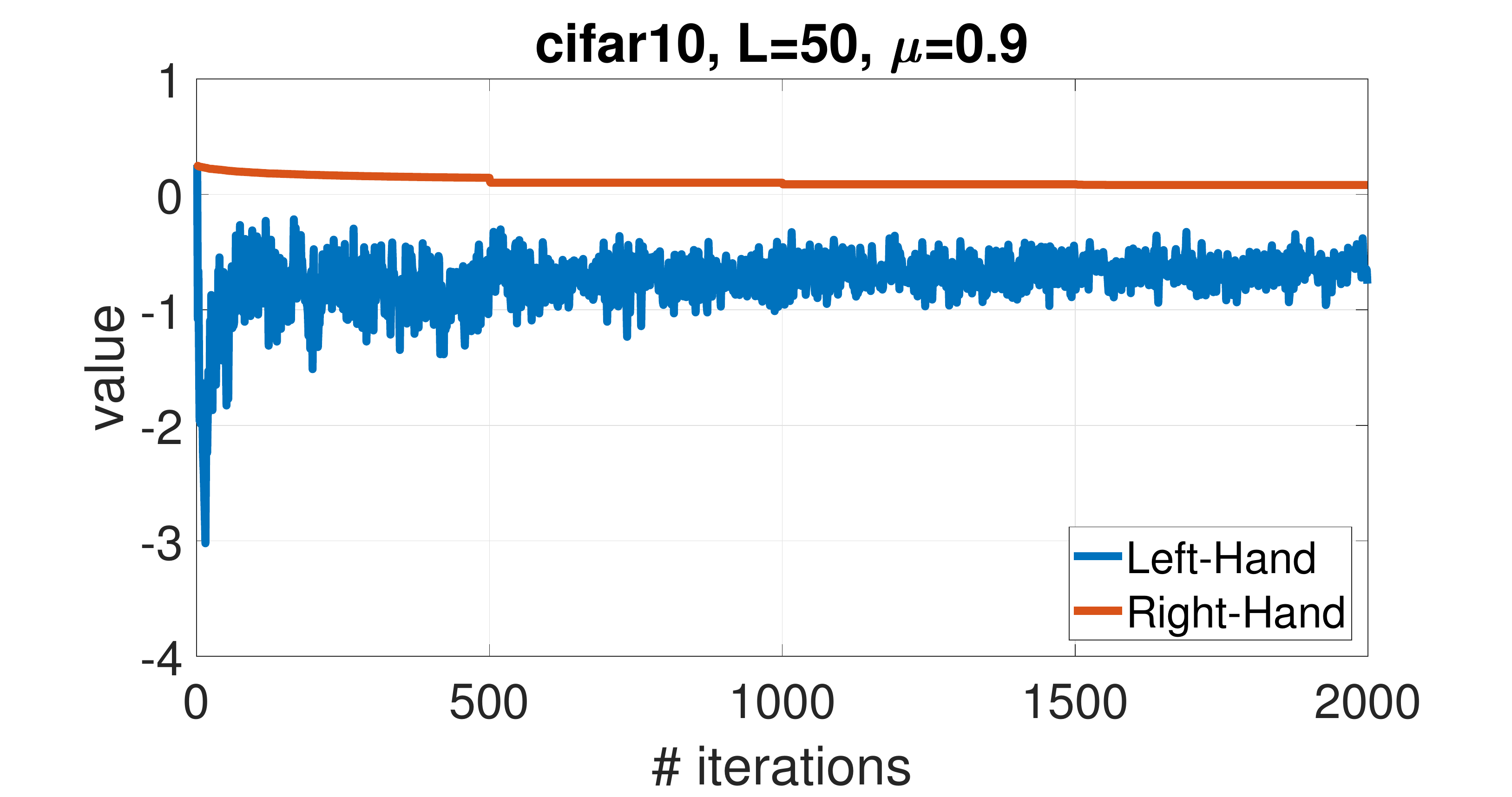}} 
		\end{center}
	\end{minipage}
	\vspace{-8mm}
	\caption{Comparison between LHS and RHS of Eq.~(\ref{eqn:sampling_rule}) based on $\mathbf{x}_t$ returned by Alg.~\ref{alg:BnB-GD} using different values for momentum parameter $\mu$.}
	\label{fig:A4} 
\end{figure}

To justify this, we collect $\eta_t$'s by running Alg.~\ref{alg:BnB-GD} on both datasets, and plot them in Fig.~\ref{fig:eta}. In general, these numbers are indeed sufficiently small for local update based on gradients, and $\eta_t$ decreases with the increase of iterations. This behavior is expected as the objective $f$ is supposed to decrease as well~\textit{w.r.t.} the number of iterations. The value gap at the beginning on the two datasets is mainly induced by different $L$'s.

\begin{figure*}[t]
	\begin{minipage}[b]{0.33\linewidth}
		\begin{center}
			\centerline{\includegraphics[clip=true,width=\linewidth]{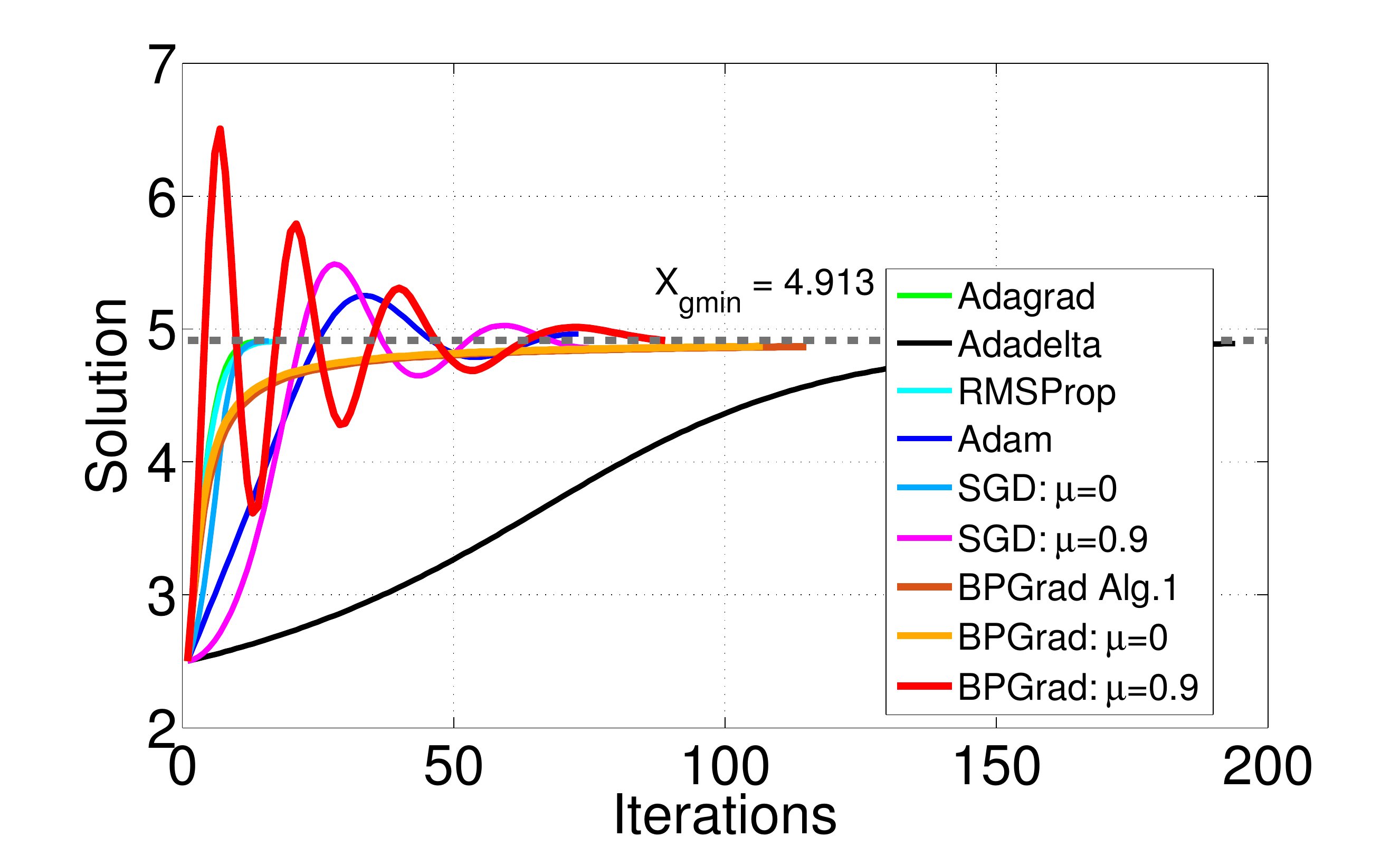}}	
			\centerline{(a) function: $f_1$}	
		\end{center}
	\end{minipage}
	\begin{minipage}[b]{0.33\linewidth}
		\begin{center}
			\centerline{\includegraphics[clip=true,width=\linewidth]{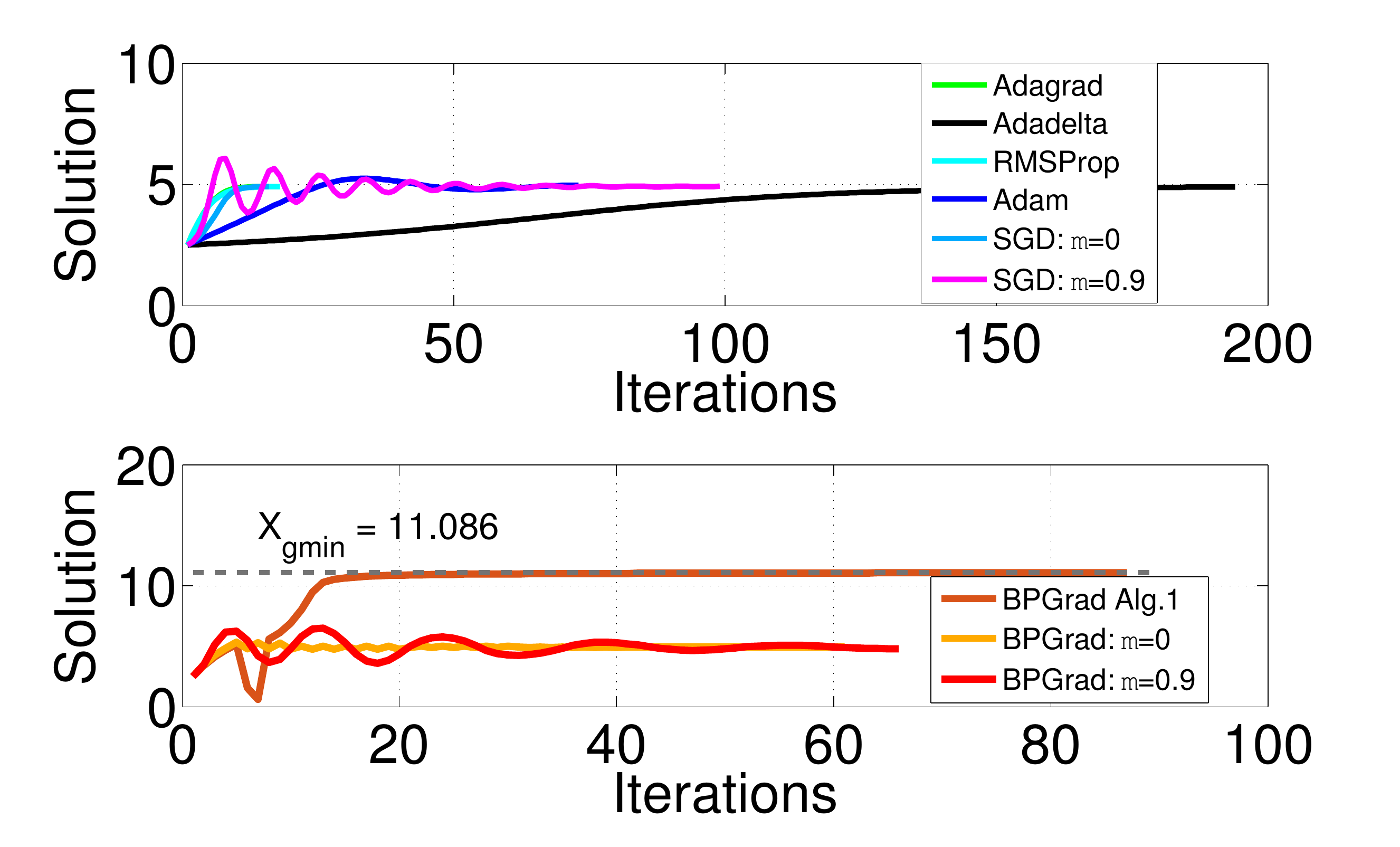}}	
			\centerline{(b) function: $f_2$}	
		\end{center}
	\end{minipage}
	\begin{minipage}[b]{0.32\linewidth}
		\begin{center}
			\centerline{\includegraphics[clip=true,width=\linewidth]{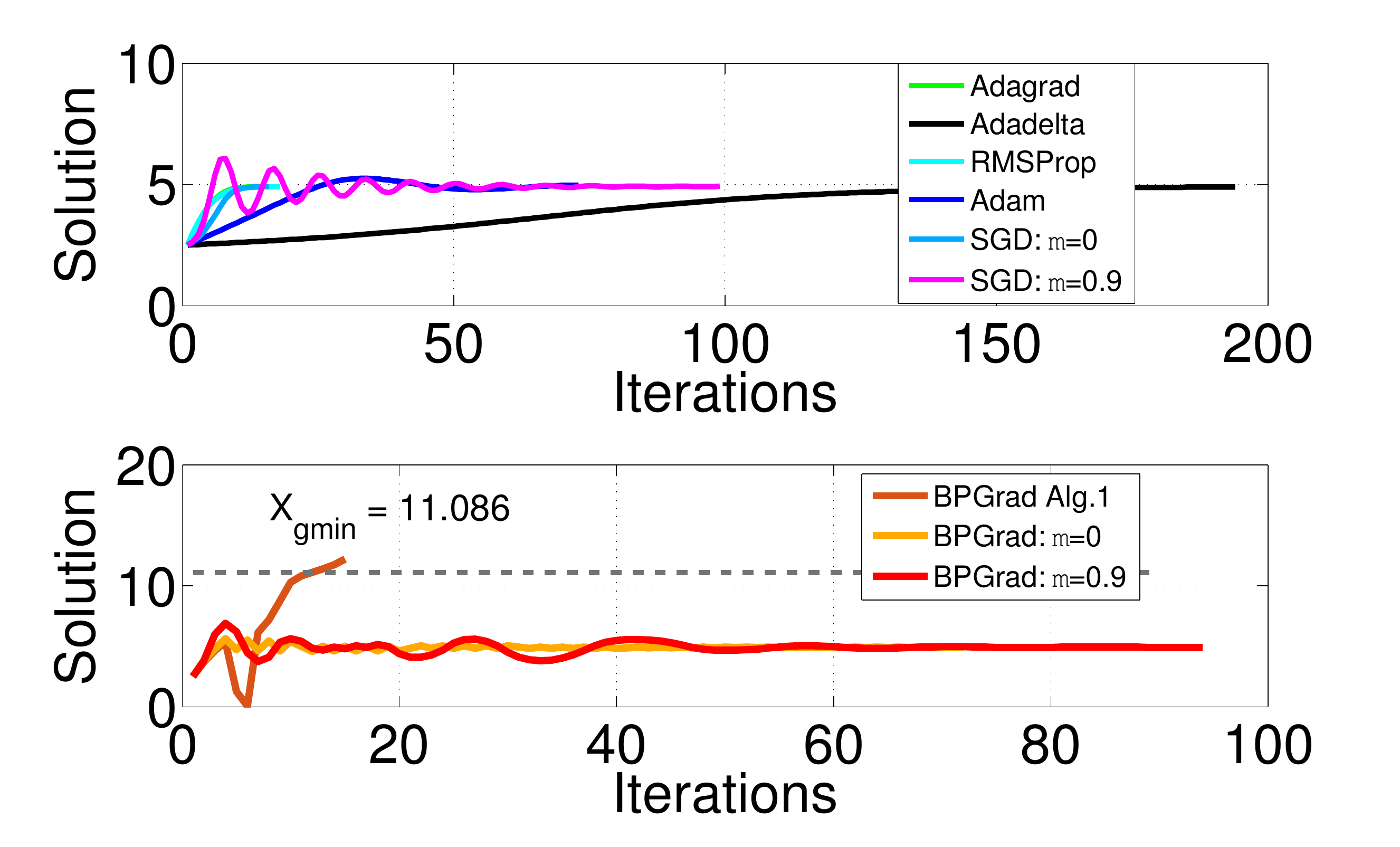}}	
			\centerline{(c) function: $f_3$}	
		\end{center}
	\end{minipage} 
	\vspace{-8mm}
	\caption{Trajectories of different solves on problems with known solutions (gray dashed line denotes the global solution for each function). {\bf (a)} The function $f_1=x\sin(x)+4.815, x\in[0, 8]$ has one global minimum at $X_{gmin}=4.913$ . {\bf (b)} The function $f_2=x\sin(x)+11.05, x\in[0, 4\pi]$ has a local minimum at $X_{gmin}=4.913$, and a global minimum at $X_{gmin}=11.086$. {\bf (c)} The function $f_3=x\sin(x)+15, x\in[0, 4\pi]$ has the same local and global minimums as $f_2$, but it has a larger constant offset than $f_2$.} 
	\label{fig:illustration_in_1d_case}
\end{figure*}

\subsubsection{Feasibility of A4}
To justify this, we show some evidence in Fig.~\ref{fig:A4}, where we plot the left-hand side (LHS) and right-hand side (RHS) of Eq.~(\ref{eqn:sampling_rule}) based on $\mathbf{x}_t$ obtained from the Alg.~\ref{alg:BnB-GD}. As we see in all the subfigures on the right with $\mu=0.9$, the values on RHS are always no smaller than those on LHS correspondingly. In contrast, in the remaining subfigures on the left with $\mu=0$ (i.e.\ vanilla SGD update), the values on RHS are always no bigger than those on LHS correspondingly. These observations appear to be robust across different datasets, and irrelevant to parameter $L$ which determines the radius of balls, i.e.\ step sizes for gradients. The momentum parameter $\mu$, which is related to the directions of gradients for model updating, appears to be the only factor to make the samples of our solver satisfy Eq.~(\ref{eqn:sampling_rule}). This also supports our claims in Thm.~\ref{thm:fast_sample} and Cor.~\ref{cor:1} about the relation between the model update and gradient in order to satisfy Eq.~(\ref{eqn:sampling_rule}). More evidence has been provided by the experiments on MNIST and CIFAR-$10$ datasets in Sec.~\ref{sec:exp}. Based on this evidence, it is safe to say that the assumption {\em A2} holds empirically when using sufficiently large values for momentum $\mu$.

\subsection{Convergence of BPGrad Algorithm and Solver}\label{sec:SGD_vs_BPGrad_convergence}
\subsubsection{One-Dimensional Problems with Known Solutions}
To explore the strength and weakness of the proposed approach in an interpretable domain, we first apply BPGrad in Alg.~\ref{alg:BnB} and its approximate solver in Alg.~\ref{alg:BnB-GD} to nonconvex problems with limited complexity. The problem we consider is to search for the global minimum of the one-dimension sinusoidal function $f(x)=x\sin(x)$ with different constant offsets, which enables us to visualize the trajectories found by each solver. We perform a comparison with Adagrad, Adadelta, RMSProp, Adam, and SGD.

The trajectories are shown in Fig.~\ref{fig:illustration_in_1d_case}. We use a grid-search to determine the best hyper-parameter setting for each solver (details can be found in the Supplementary materials). We report the number of iterations that are needed to converge with a tolerance of $\epsilon=10^{-4}$ in terms of function values. We can observe that all the solvers find the global minimum of function $f_1$. However, only the BPGrad in Alg.~\ref{alg:BnB} locates the global minimum of functions $f_2$ and $f_3$, respectively, while the other solvers are stuck at the local minimum. These observations empirically indicate that BPGrad in Alg.~\ref{alg:BnB} is capable of reaching the global optimum.

\subsubsection{Two-layer Neural Network Optimization}
\begin{figure} 
	\begin{center}
		\includegraphics[width=0.41\linewidth]{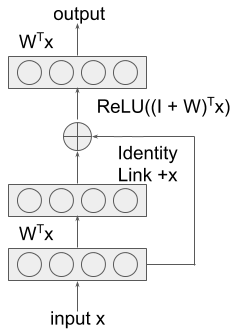}
		\caption{Illustration of two-layer networks.}
		\label{fig:network1}
	\end{center}
\end{figure}
Based on the empirical justification of convergence behavior of our solver in Sec.~\ref{subsec:Empirical_justification}, in this section, we also consider a numerical demonstration of converging to the global optimum using our solver. Recently Li and Yuan in~\cite{NIPS2017_Convergence} proved theoretically that SGD can converge to the global minimum in polynomial time in two-layer neural networks with ReLU activation when the input data and network weight initialization follow Gaussian distributions. 

To demonstrate the convergence of our solver, we implement such a two-layer network in~\cite{NIPS2017_Convergence} as illustrated in Fig.~\ref{fig:network1} with $10,302$ parameters. We train the network using SGD and our BPGrad solver, respectively, with $20$ epochs, batch size of $200$, and momentum of $0.9$. We conduct grid search on learning rate ($lr$) and Lipschitz constant $L$ for SGD and BPGrad, respectively. Then, we measure the Euclidean distance between the two learned network weights. We observe a marginal difference of 0.6 among the 10,302 dimensions. Numerically we can say that both the SGD and our solver converge to the same global minimum.

Given those strong pieces of evidence, we thus hypothesize that, with proper momentum, it is very likely that our solver in Alg.~\ref{alg:BnB-GD} will preserve the theoretical convergence properties of BPGrad algorithm in Alg.~\ref{alg:BnB}, and can locate a solution towards to global optimum in training deep models.

\section{Experiments}\label{sec:exp}
We utilize MatConvNet as our testbed, and employ its demo code as well as the default network architectures for different tasks. Since our solver can automatically determine the learning rates, we compare ours with SGD as well as another four widely used DL solvers with adaptive learning rates, namely Adagrad, Adadelta, RMSProp, and Adam. We use grid-search to determine the best hyper-parameter settings (details can be found in Supplementary materials) for all the solvers and report their best performance. 

\subsection{Estimation of Lipschitz Constant $L$}\label{subsec:Estimate_of_L}
We take the experiments on MNIST and CIFAR-$10$ datasets as examples to show the possibility of automatically tuning or reducing the searching space for manually tuning the parameter. For MNIST dataset, we take LeNet-5 as the network in our experiments. We randomly initialize the parameters (including weights and biases), and randomly feed a mini-batch into the network (\textit{i.e.} feed-forward) to compute the objective value. Specifically, the filter weights are initialized with random numbers following a Gaussian distribution and the biases are initialized to be zero. The training samples are randomly shuffled and a mini-batch of $100$ samples are selected from this shuffled training pool. We repeat this procedure for $600$ times in one epoch on MNIST dataset, leading to $600$ copies of initial network parameters as well as $600$ objectives. Similarly, we repeat it for $500$ times with mini-batch size of $100$ on CIFAR-$10$ dataset with a similar network, leading to the same amount of initial network parameters and objectives.

Based on the definition of Lipschitz continuity in Eq.~(\ref{def:Lipschitz}), we can compute $L$ as follows:
\begin{align}\label{eqn:L}
L = \frac{|f(\mathbf{x}_1)-f(\mathbf{x}_2)|}{\|\mathbf{x}_1-\mathbf{x}_2\|_2}, \forall \mathbf{x}_1 , \mathbf{x}_2 \in \mathcal{X},
\end{align}
where $\mathbf{x_i}$ denotes one copy of initial network parameters, and $f(\mathbf{x_i})$ denotes the corresponding objective value.

\begin{figure}[t]
	\begin{minipage}[b]{0.49\columnwidth}
		\begin{center}
			\centerline{\includegraphics[width=\columnwidth]{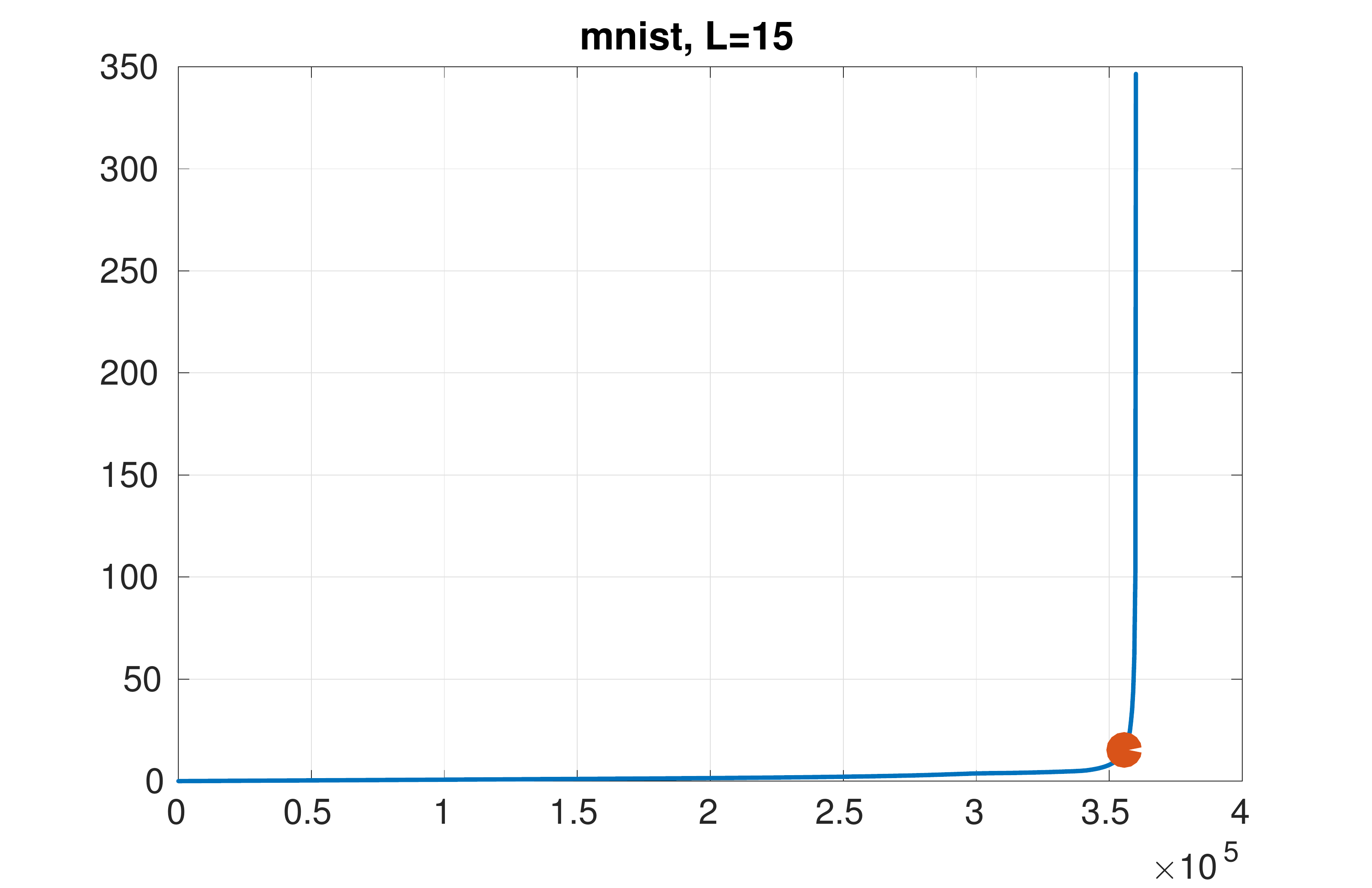}}	
		\end{center}
	\end{minipage}
	\begin{minipage}[b]{0.49\columnwidth}
		\begin{center}
			\centerline{\includegraphics[width=\columnwidth]{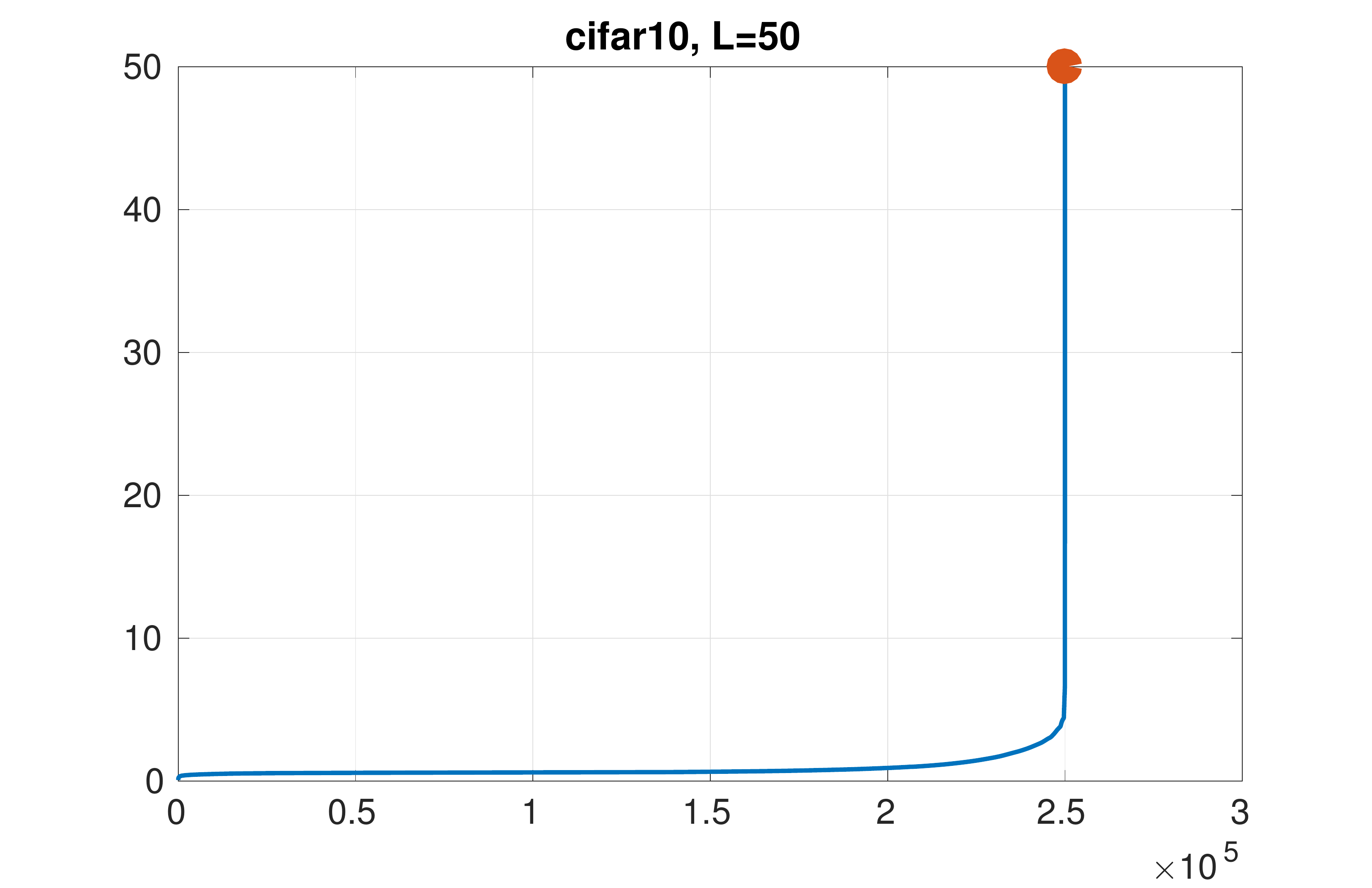}}
		\end{center}
	\end{minipage} 
	\vspace{-8mm}
	\caption{Distributions of computed Lipschitz constant $L$'s in ascending order ($y$-axis denotes the value of $L$). Here the red dots denote the $L$ values that are used in our experiments, \textit{i.e.} $L=15$ on MNIST (left) and $L=50$ on CIFAR-$10$ (right).} 
	\label{fig:dis_L}
\end{figure}

By utilizing all the initial parameters as well as the objectives, we compute $L$ based on Eq.~(\ref{eqn:L}) and plot the distributions of these $L$'s in Fig.~\ref{fig:dis_L}. It is evident that only a tiny portion of computed $L$'s have relatively large values on both datasets. This behavior indicates that the surfaces induced by the objective functions in DL are in general quite smooth (\textit{i.e.} without large jumps in the surface), and only in some regions the curvatures are high, where good models may exist. Therefore, this behavior verifies that our Lipschitz continuity assumption in deep learning can be satisfied {\em w.h.p.} empirically. In practice, we select the value as initial $L$ that surpasses a threshold on the distribution of computed $L$'s.

\begin{figure}[t]
	\begin{minipage}[b]{0.49\columnwidth}
		\begin{center}
			\centerline{\includegraphics[width=\columnwidth,height=0.75\columnwidth]{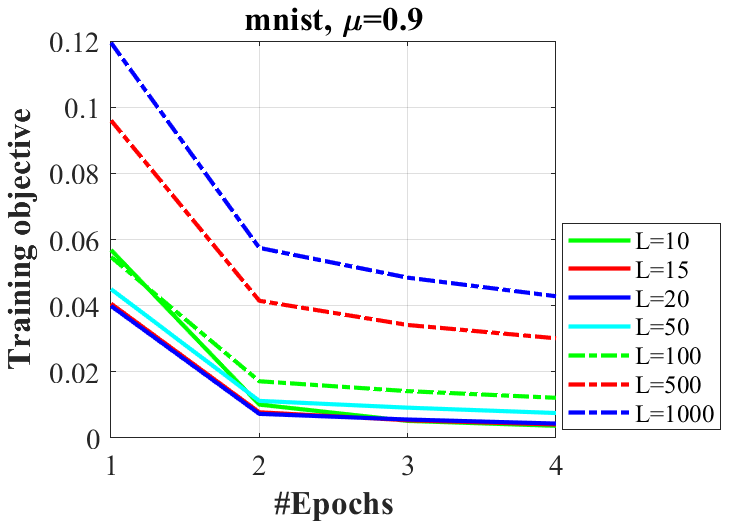}}			
		\end{center}
	\end{minipage}
	\begin{minipage}[b]{0.49\columnwidth}
		\begin{center}
			\centerline{\includegraphics[width=\columnwidth,height=0.75\columnwidth]{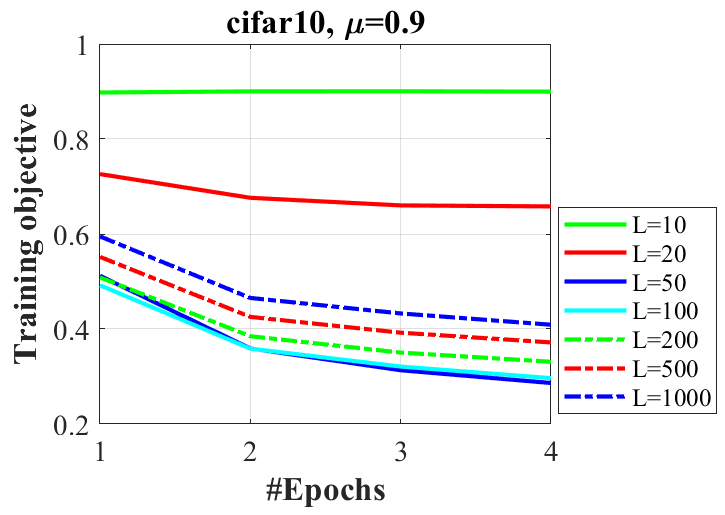}}
		\end{center}
	\end{minipage} 
	\vspace{-8mm}
	\caption{Illustration of robustness of Lipschitz constant $L$ in our solver.}
	\label{fig:L}
\end{figure}

In addition, we observe in the experiments that the parameter $L$ as Lipschitz constant is quite robust \textit{w.r.t.} its performance, indicating that heavily tuning this parameter is unnecessary in practice. To demonstrate the robustness of Lipschitz constant $L$ in our solver, we compare the training objectives of our solver by varying $L$ in Fig.~\ref{fig:L}~(top row). To highlight the differences, we only crop and show the results in the first four epochs, and the remaining results have similar behavior. As we can see on the MNIST dataset, when $L$ varies from $10$ to $100$, the corresponding curves are clustered. Similar clustering is observed as well on the CIFAR-$10$ dataset for $L$ varies from $50$ to $1000$.

Moreover, we notice that the best $L$ in Fig.~\ref{fig:L}, \textit{i.e.} $L=20$ on MNIST and $L=50$ on CIFAR-$10$, respectively, tends to be within the range of the computed $L$'s on each dataset shown in Fig.~\ref{fig:dis_L}. This observation can be used to facilitate the parameter tuning procedure, as we do not necessarily consider any parameter that is far beyond the range, \textit{e.g.} $L=500$ or $L=1000$ on MNIST. Therefore, we set $L=15$ for MNIST and $L=50$ for CIFAR-$10$, respectively, in our solver (see Sec.~\ref{subsubsec:mnist} and Sec.~\ref{subsubsec:cifar10}).

\begin{figure}[t]
	\begin{minipage}[b]{0.49\columnwidth}
		\begin{center}
			\centerline{\includegraphics[width=\columnwidth,height=0.75\columnwidth]{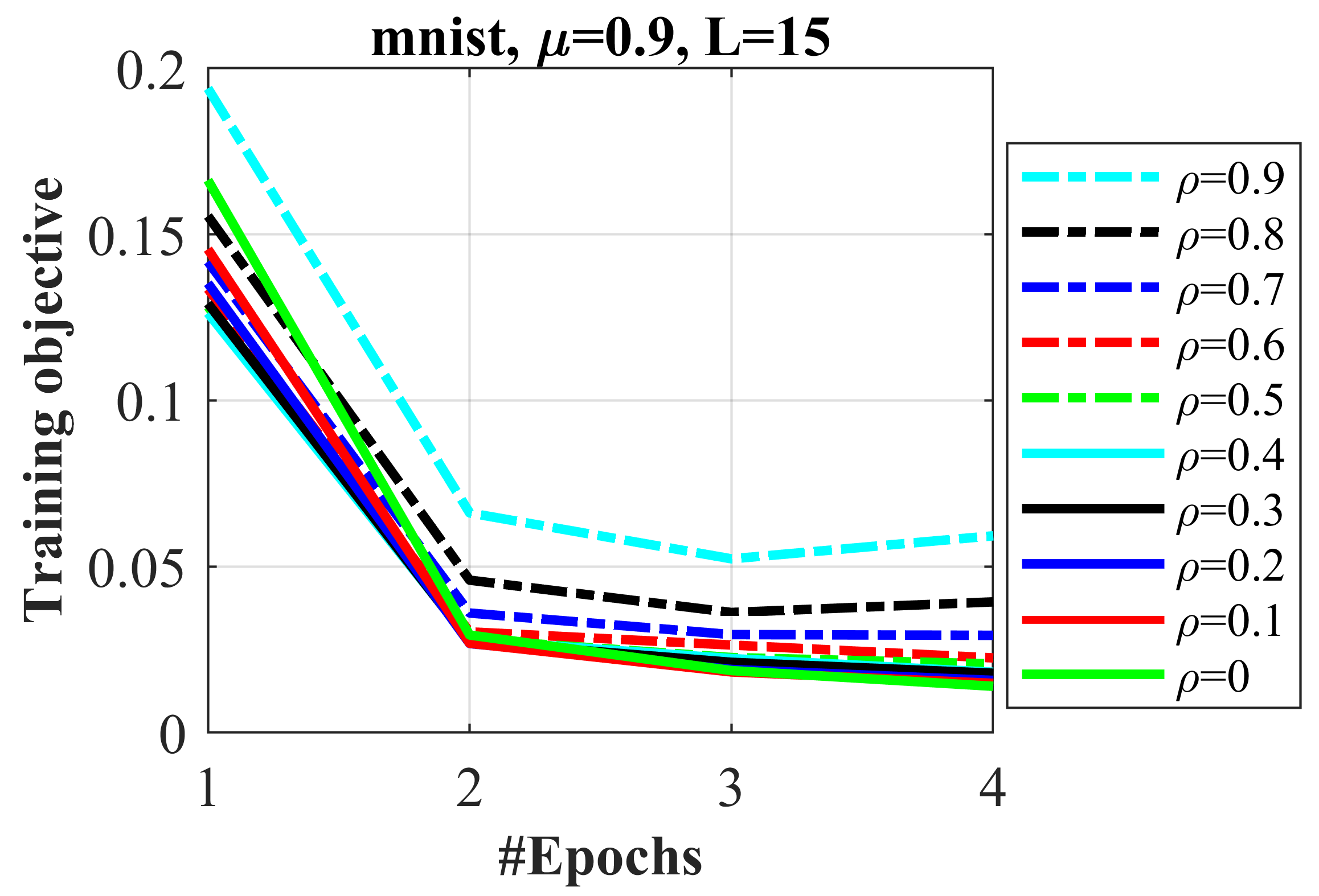}}			
		\end{center}
	\end{minipage}
	\begin{minipage}[b]{0.49\columnwidth}
		\begin{center}
			\centerline{\includegraphics[width=\columnwidth,height=0.75\columnwidth]{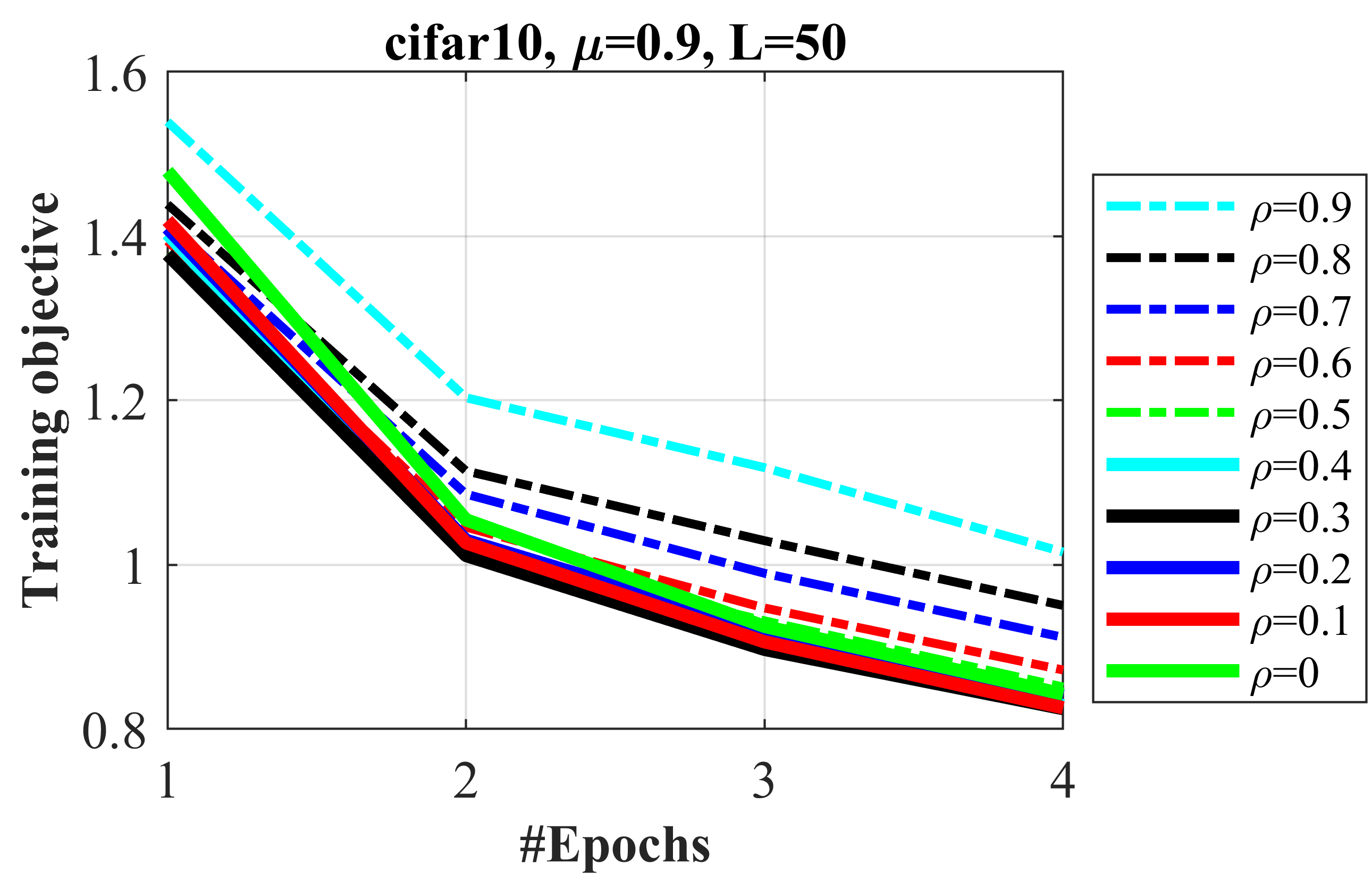}}
		\end{center}
	\end{minipage}	 
	\vspace{-8mm}
	\caption{Illustration of robustness of parameter $\rho$ in our solver.}
	\label{fig:rho}
\end{figure}

\subsection{Effect of $\rho$ on performance}\label{subsec:Effect_of_rho}
Similar robustness is also observed in the experiments for the parameter $\rho$ related to the lower bound estimator in our solver. We compare the training objectives of our solver by varying $\rho$ in a similar setting as $L$ in Fig.~\ref{fig:rho}. For the MNIST dataset, we set $L=15$ and vary $\rho$ from $0$ to $0.9$. As we can see from the result, when $\rho$ varies from $0$ to $0.5$, the corresponding curves are clustered. Similar result is obtained on CIFAR-$10$ for $\rho$ varies from $0$ to $0.5$. Therefore, in the following experiments, we set the initial value of $\rho=0.1$ to make a trade-off between the estimation value of the lower bound and size of the removable parameter space for our BPGrad solver.  

\begin{figure}[t]
	\begin{minipage}[b]{0.49\columnwidth}
		\begin{center}		
			\centerline{\includegraphics[width=\columnwidth,height=0.75\columnwidth]{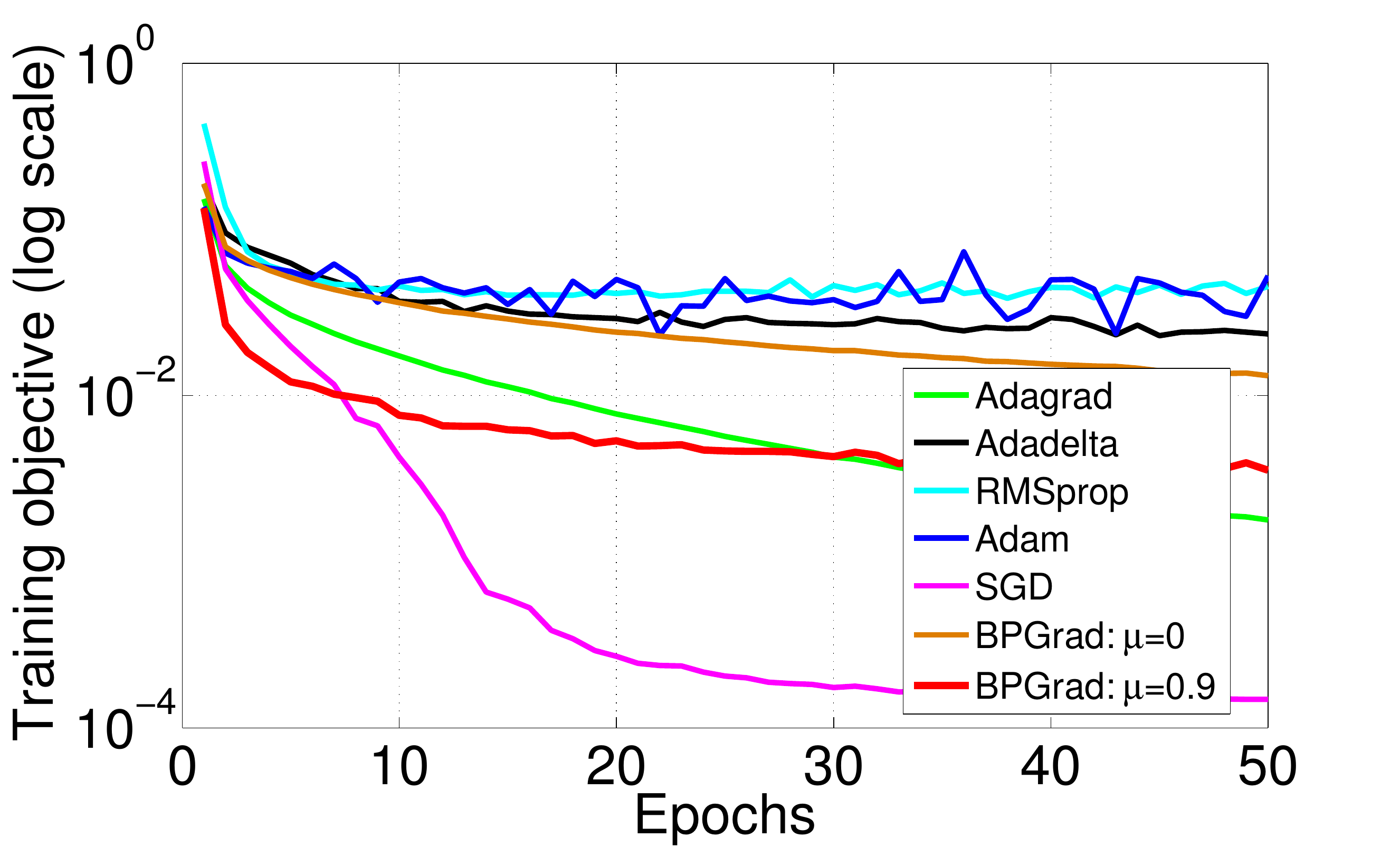}}	
		\end{center}
	\end{minipage}
	\begin{minipage}[b]{0.49\columnwidth}
		\begin{center}
			\centerline{\includegraphics[width=\columnwidth,height=0.75\columnwidth]{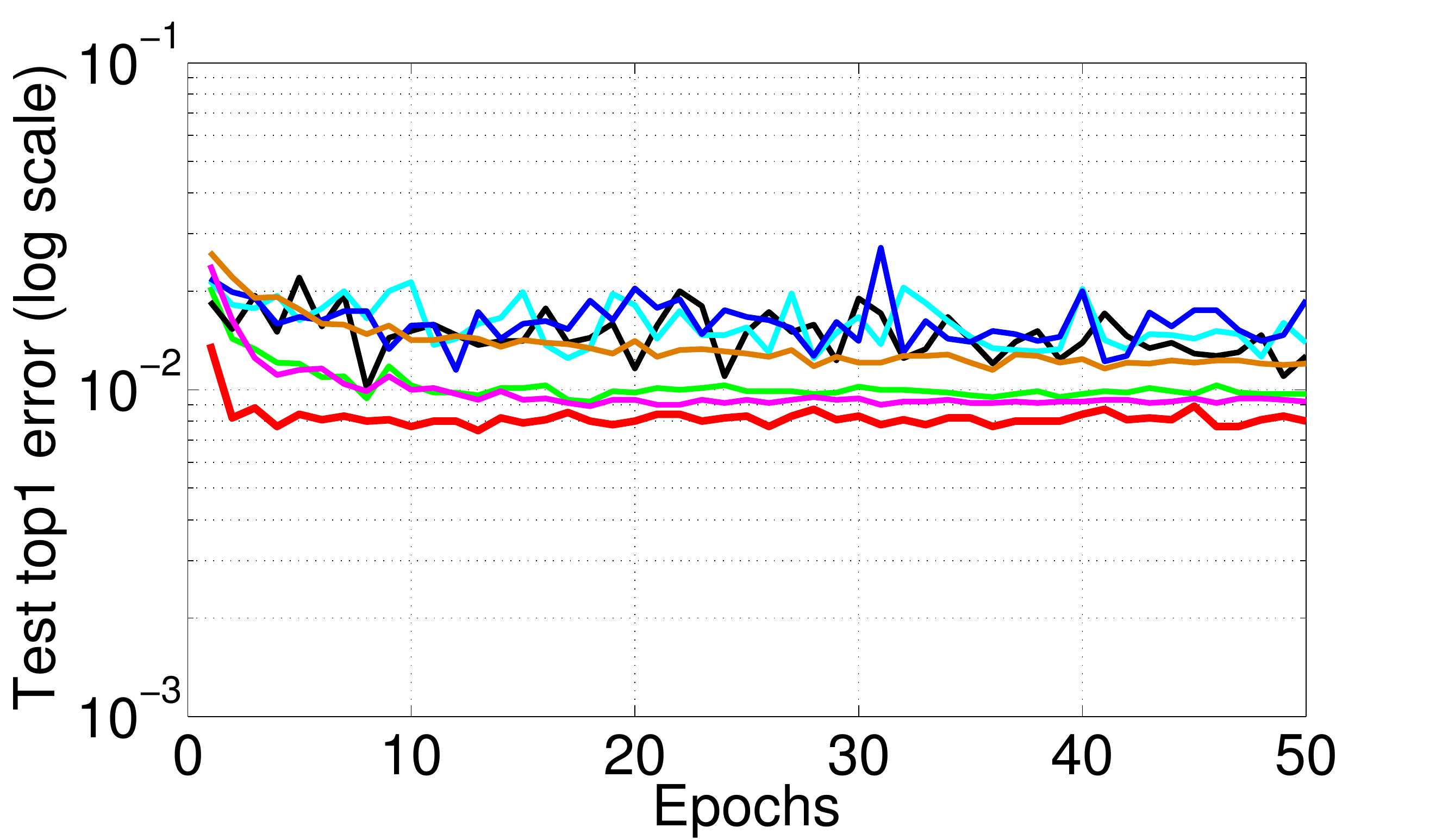}}
		\end{center}
	\end{minipage}	
	\vspace{-8mm} 
	\caption{Comparison on {\bf (left)} training objectives and {\bf (right)} test top-1 errors for object recognition using LeNet-$5$ on MNIST.}
	\label{fig:classification_mnist}
\end{figure}

\subsection{Object Recognition}\label{subsec:object_recognition}
In this section, we explore the use of BPGrad solver in object recognition with different CNNs on four benchmark datasets: MNIST~\cite{lecun1998mnist}, CIFAR-$10$~\cite{krizhevsky2012imagenet} and ImageNet~\cite{russakovsky2015imagenet}. For all the datasets, we follow the default implementation to train the individual CNN model on each dataset.

\subsubsection{MNIST}\label{subsubsec:mnist}
The MNIST dataset consists of handwriting digits $0$ to $9$ which are gray images with a resolution of $28\times28$ pixels. There are $60,000$ training images and $10,000$ testing images in total in $10$ classes labeled from $0$ to $9$. For this dataset, we train an individual LeNet-$5$~\cite{lecun1998gradient} model using each solver. For the details of network architectures please refer to the demo code. Specifically, for all the solvers, we train the network for $50$ epochs with a mini-batch size $100$, weight decay $0.0005$, and momentum $0.9$. In addition, we fix the initial weights for all solvers and the feeding order of mini-batches for a fair comparison. The global learning rate is set to $0.001$ on MNIST for Adagrad, RMSProp, Adam, and SGD. Adadelta does not require a global learning rate.  

The results are shown in Fig.~\ref{fig:classification_mnist}. 
To illustrate the effect of momentum in our solver in terms of performance, here we plot two variants of our solver with $\mu=0$ and $\mu=0.9$, respectively. It is clear that our solver with $\mu=0.9$ works much better than its counterparts, achieving lower training objectives as well as a lower top-$1$ error at test time. This again provides evidence to support the importance of satisfying Eq.~(\ref{eqn:sampling_rule}) in our solver to search for optimal solutions toward global minima.

The proposed Lipschitz continuity assumption can serve as regularization in deep learning. As can be seen in Fig.~\ref{fig:dis_L} on MNIST, the $L=15$ used in Fig.~\ref{fig:classification_mnist} is much smaller than the maximumly computed $L$, making the surface of the approximate function much smoother. This eventually leads to a higher objective than SGD and Adagrad in Fig.~\ref{fig:classification_mnist}. The test error of the BPGrad solver, however, is lower than SGD and Adagrad, owing to the functionality of regularization.

\begin{figure}[t]	  
	\begin{minipage}[b]{0.49\columnwidth}
		\begin{center}		 			
			\centerline{\includegraphics[width=\columnwidth,height=0.75\columnwidth]{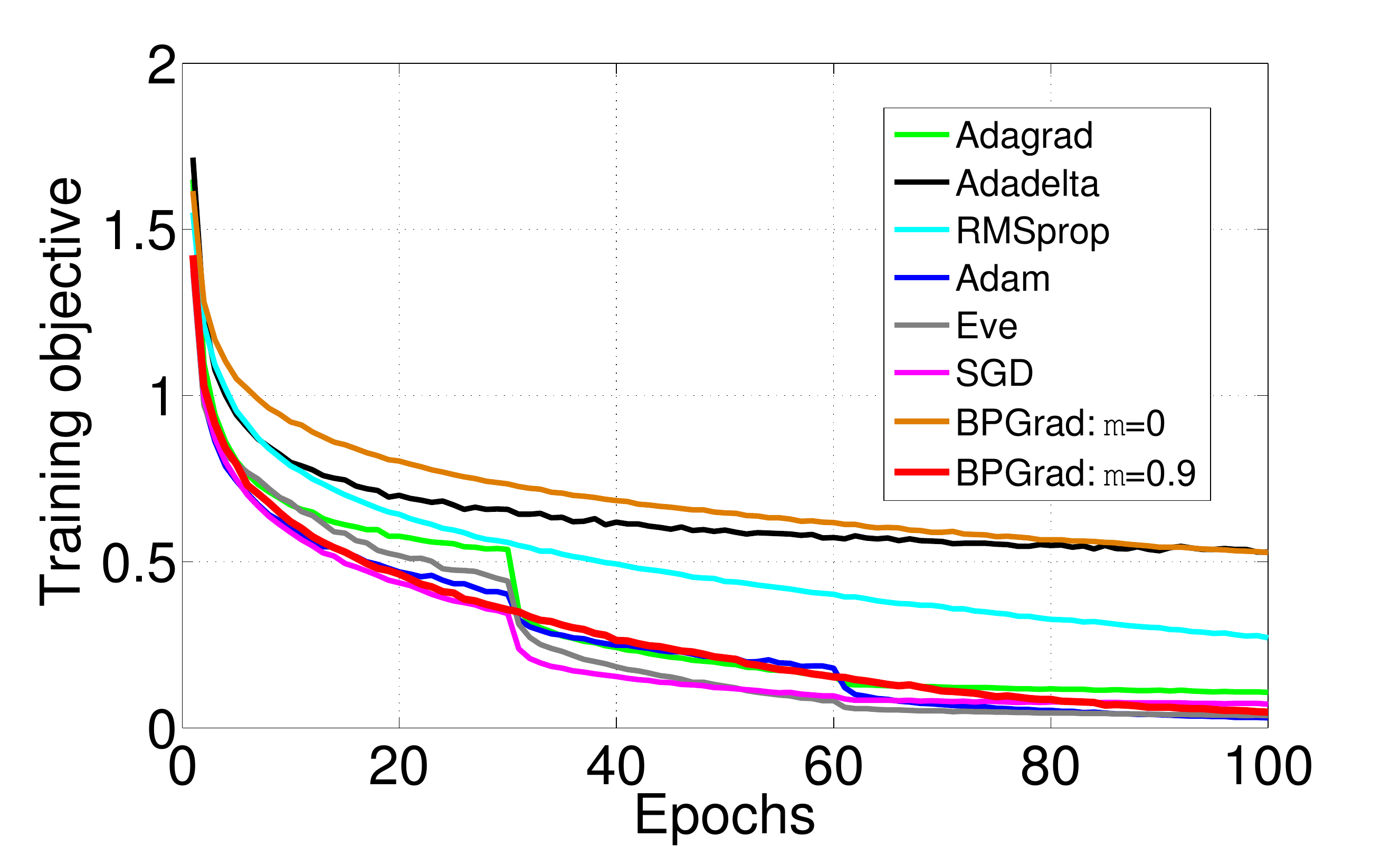}}
		\end{center}
	\end{minipage}
	\begin{minipage}[b]{0.49\columnwidth}
		\begin{center} 
			\centerline{\includegraphics[width=\columnwidth,height=0.75\columnwidth]{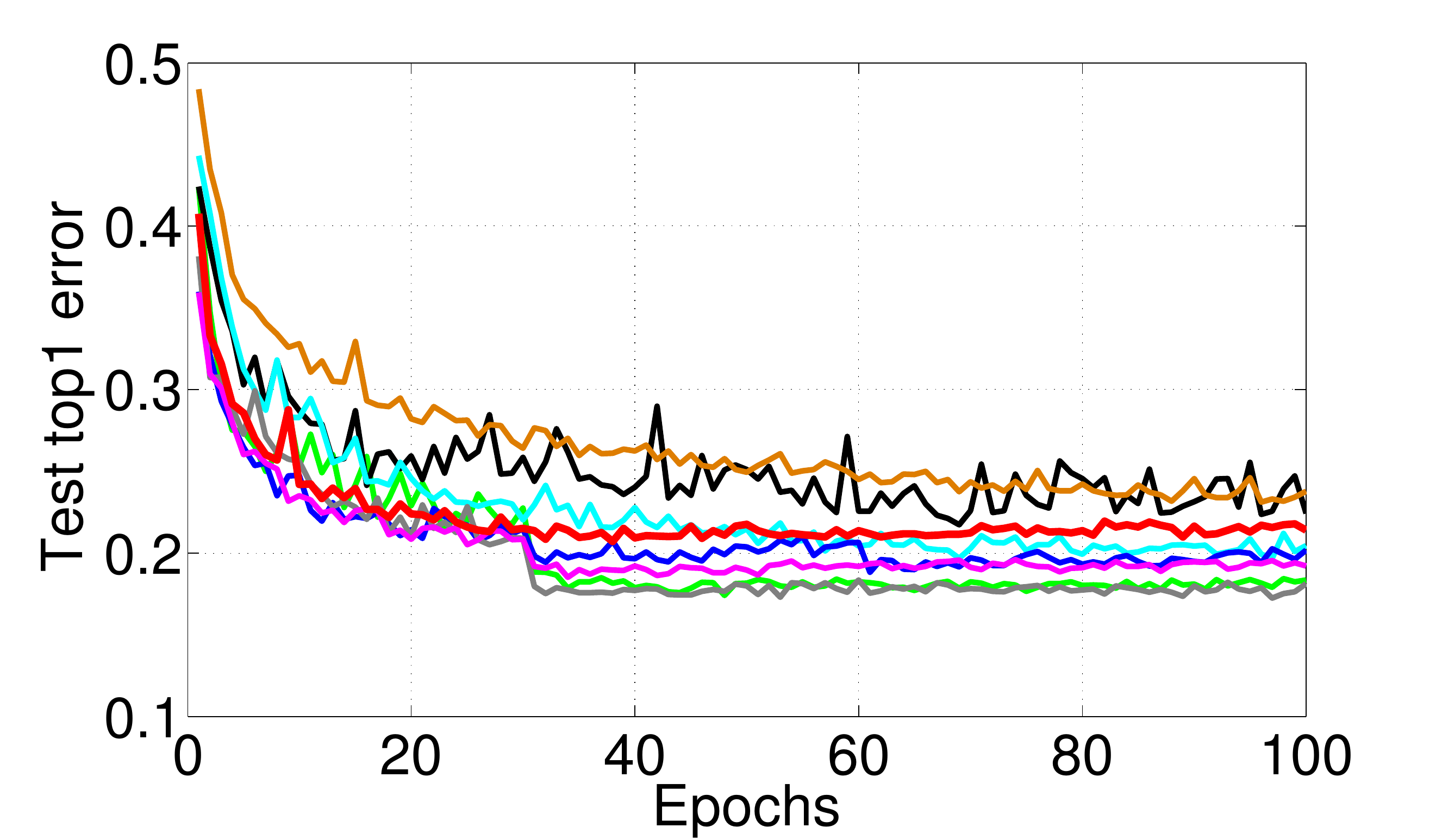}}
		\end{center}
	\end{minipage}	 
	\vspace{-8mm}
	\caption{Comparison on {\bf (left)} training objectives and {\bf (right)} test top-1 errors for object recognition using network similar to LeNet on CIFAR-$10$.} 
	\label{fig:classification_cifar10}
\end{figure}

\begin{table}[t]
	\caption{Training objective and recognition error on Cifar10 dataset in the form of mean$\pm$std using 5 trials (best in bold, 2nd best with underline).} 
	\footnotesize
	\begin{center}    
		\begin{tabular}{|c||c|c|}
			\hline 
			& Train Objective  & Test top-1 error  \\ \hline
			\hline
			Adagrad  & 0.110$\pm$0.0036 &    \textbf{0.181$\pm$0.0041}\\ \hline
			Adadelta &  0.534$\pm$0.0084  &   0.224$\pm$0.0051  \\ \hline
			RMSProp  &  0.302$\pm$0.0021  &  0.205$\pm$0.0017  \\ \hline
			Adam     &  0.113$\pm$0.0072  & 0.192$\pm$0.0029  \\ \hline
			Eve &  0.113$\pm$0.0118 &   0.197$\pm$0.0036 \\ \hline
			SGD      &  \underline{0.075$\pm$0.0036}  &   \underline{0.191$\pm$0.0038}\\ \hline
			BPGrad ($\mu$=0)  &  0.528$\pm$0.0028 &   0.235$\pm$0.0011\\ \hline
			BPGrad ($\mu$=0.9)  &  \textbf{0.045$\pm$0.0008} &   0.215$\pm$0.0034\\ \hline
		\end{tabular}
	\end{center}	
	\label{table:cifar10_run5times} 
\end{table} 

\subsubsection{CIFAR-$10$}\label{subsubsec:cifar10}
The CIFAR-$10$ dataset consists of $10$ object classes of natural images with $50,000$ training images and $10,000$ test images, where the color image resolution is $32\times32$ pixels.  

Similar to LeNet~\cite{lecun1998gradient}, for each solver in this experiment, we train an individual model for $100$ epochs on this dataset, with a mini-batch size $100$, weight decay $0.0005$, and momentum $0.9$. In addition, we fix the initial weights for this network and the feeding order of mini-batches for a fair comparison. The global learning rate is set to $0.001$ for RMSProp; but to $0.01$ for Adagrad, Adam and Eve~\cite{koushik2016improving}, and it is reduced to $0.005$ and $0.001$ at the $31$-st and $61$-st epochs. The initial learning rate for SGD is $0.05$, and it is multiplied by $0.1$ at the $31$-st and $61$-st epochs. The Lipschitz constant L for our solver is set to $50$ for this network.

The results are shown in Fig.~\ref{fig:classification_cifar10}. In addition, we show the results of 5 runs in Table~\ref{table:cifar10_run5times}. Our solver achieves the best performance in terms of training objective, but leads to a slightly inferior top-$1$ error at test time using LeNet. This behavior comes from the effect of regularization on Lipschitz continuity. However, our solver can decrease the objectives much faster than all the competitors in the first few epochs. This observation reflects the superior ability of our solver in determining adaptive learning rates for gradients. In this experiment, we also compare with an extra solver, Eve, which was proposed in related work~\cite{koushik2016improving} that empirically improves Adam with the feedbacks from the objective function. We can observe that our BPGrad solver achieves very competitive performance compared with Eve. Moreover, as reported in recent work~\cite{berrada2019deep}, BPGrad outperforms Adagrad and Adam on CIFAR-$10$ dataset using both wide residual networks~\cite{Zagoruyko2016WRN} and densely connected convolutional networks~\cite{Huang_2017_CVPR}, which further provides solid evidence to support the advantage and robustness of our approach.
\begin{figure}[t]
	\begin{minipage}[b]{0.49\columnwidth}
		\begin{center}
			\centerline{\includegraphics[width=\columnwidth,height=0.75\columnwidth]{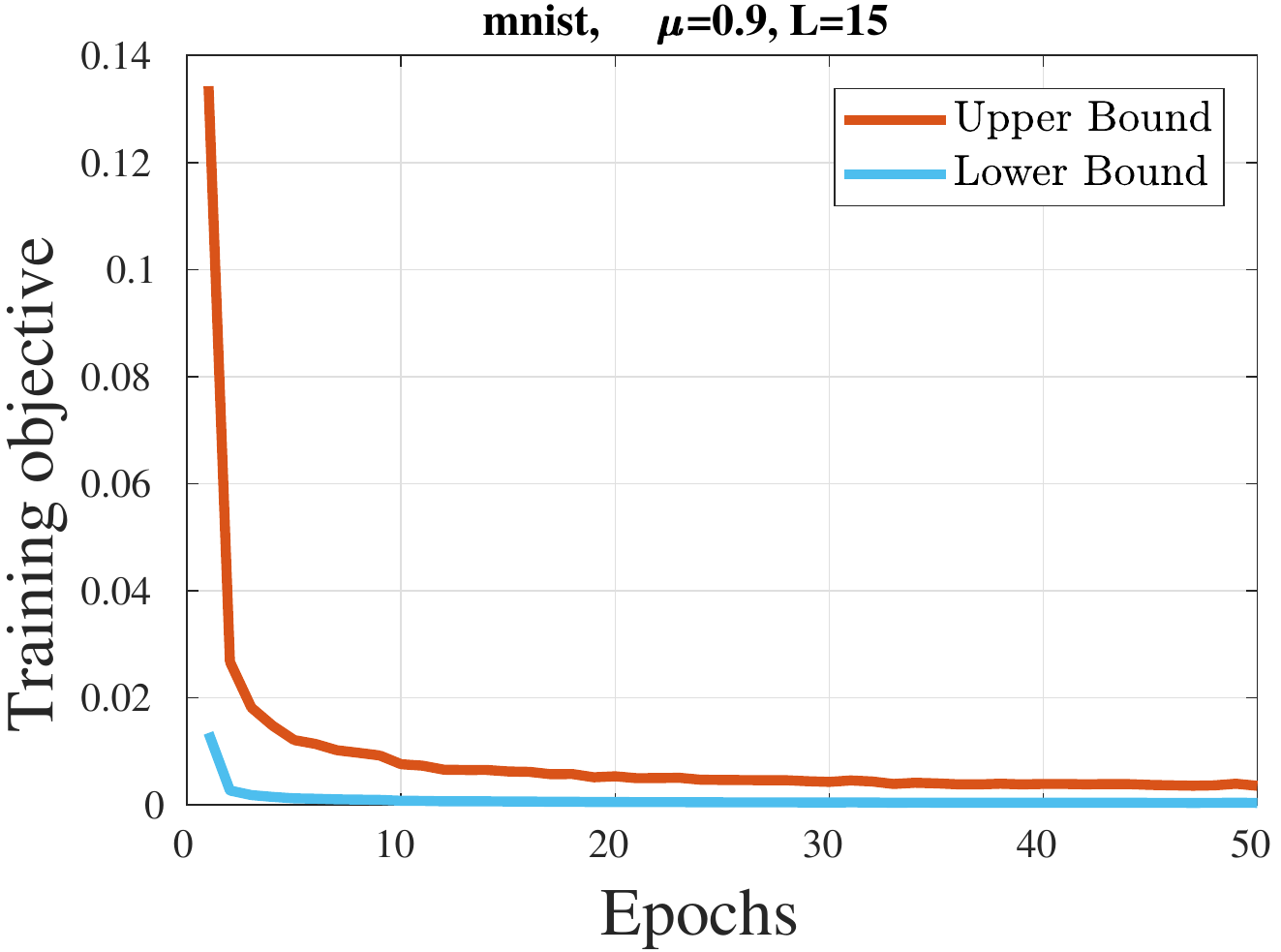}}	
		\end{center}
	\end{minipage}
	\begin{minipage}[b]{0.49\columnwidth}
		\begin{center}
			\centerline{\includegraphics[width=\columnwidth,height=0.75\columnwidth]{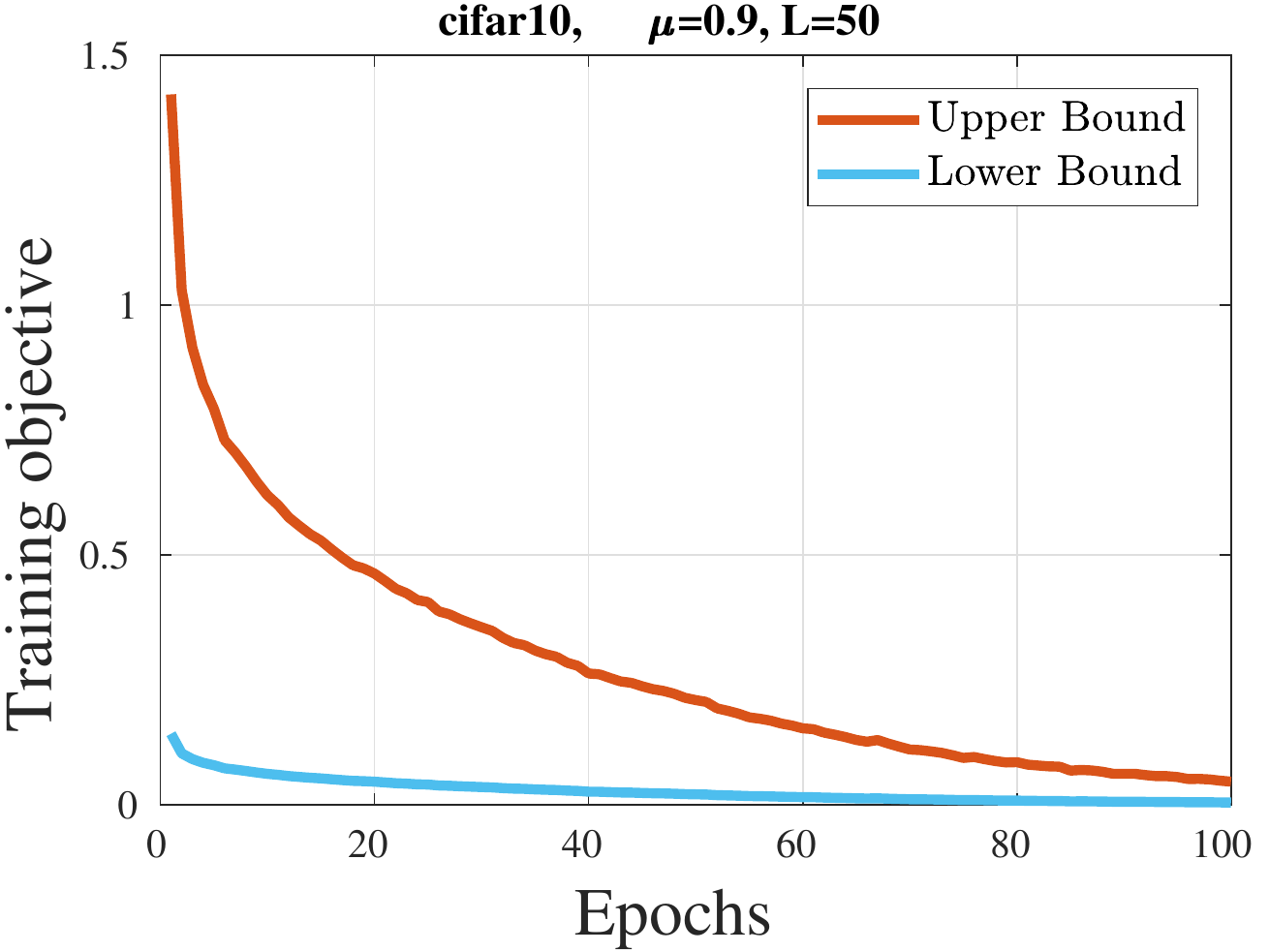}}
		\end{center}
	\end{minipage} 
	\vspace{-8mm}
	\caption{Plots of lower and upper bounds for (\textbf{left}) MNIST and (\textbf{right}) CIFAR-$10$ of our solver using LetNet.}
	\label{fig:LB_UB_mnist_cifar10} 
\end{figure}

Finally, we provide empirical evidence on the convergence of lower and upper bounds estimations on the MNIST and CIFAR-$10$ datasets. As shown in Fig.~\ref{fig:LB_UB_mnist_cifar10}, the global optimum is tightly bounded by our solver in a finite number of iterations during training. The difference between the training objectives of lower and upper bounds is computed to evaluate the convergence in our solver. For MNIST, the gap between the lower and upper bounds is reduced from $0.121$ to $0.003$ in $50$ epochs. Similarly, on CIFAR-$10$, the gap shrinks from $1.280$ to $0.042$. This provides an insight on how our solver is able to find the global solution using branch and pruning strategy. It samples (``branch") a candidate solution along the direction of the local gradient, then checks this branch against the estimated upper and lower bounds for the optimal solution, and removes (``pruning") the candidates that cannot produce a better solution than the best one found so far.

\subsubsection{ImageNet ILSVRC2012}\label{subsubsec:imagenet}
The ImageNet~\cite{russakovsky2015imagenet} dataset contains about $1.28$M training images and $50$K validation images among $1000$ object classes. In this experiment, we employ the numbers of epochs in the demo files, since those values have been fine-tuned for different solvers. Following the demo code, we train the same AlexNet~\cite{krizhevsky2012imagenet} network on the ImageNet dataset from the scratch using different solvers. We perform training for $20$ epochs, with a mini-batch size $256$, weight decay $0.0005$, momentum $0.9$, and default learning rates for the competitors. For our solver we set $L=100$ and $\mu=0.9$. 

\begin{figure}[t]			
	\begin{minipage}[b]{0.49\columnwidth}
		\begin{center}
			\centerline{\includegraphics[width=\columnwidth,height=0.75\columnwidth]{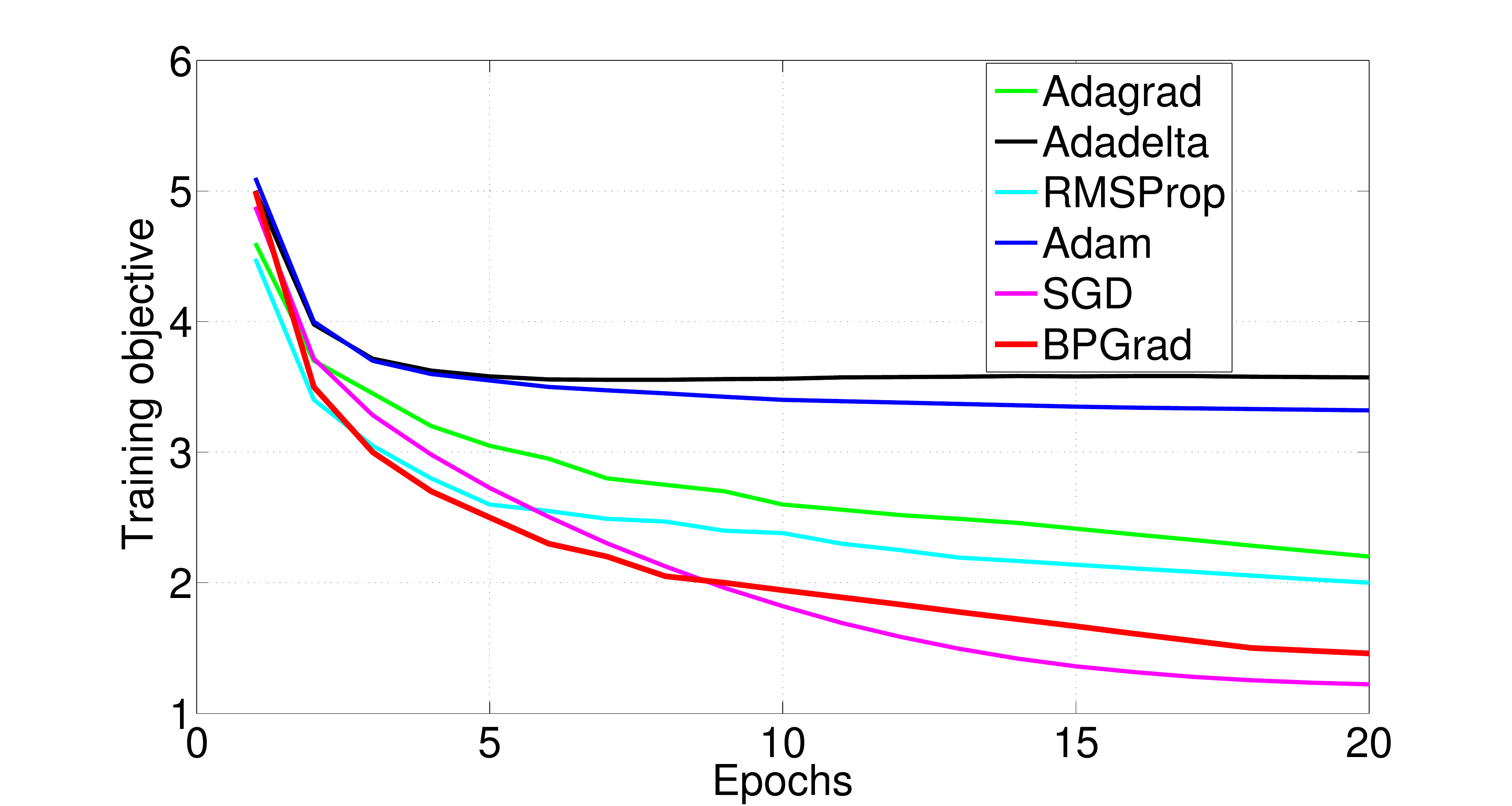}}
		\end{center}
	\end{minipage}	
	\begin{minipage}[b]{0.49\columnwidth}
		\begin{center}		
			\centerline{\includegraphics[width=\columnwidth,height=0.75\columnwidth]{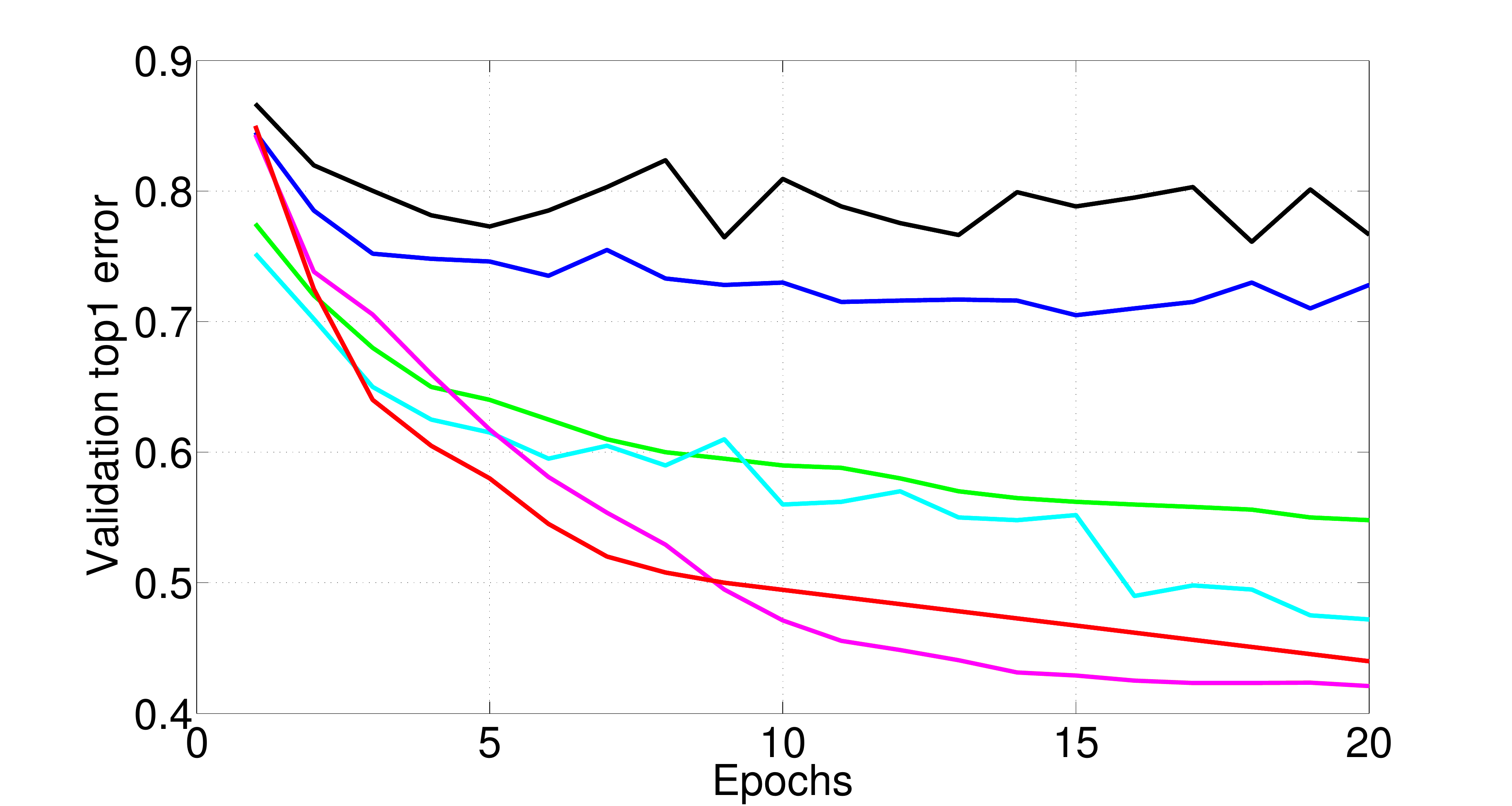}}	
		\end{center}
	\end{minipage}	 
	\vspace{-8mm}
	\caption{Comparison on {\bf (left)} training objectives and {\bf (right)} validation top-1 errors for object recognition using ImageNet ILSVRC2012.}
	\label{fig:imagenet}
\end{figure}

The results are shown in Fig.~\ref{fig:imagenet}. We can observe that BPGrad converges faster than SGD before the $10$-th Epoch at both training and test time, and achieves slightly inferior performance than SGD. However, we observe that BPGrad converges faster than all the other four competitors to achieve the lowest objective as well as the lowest top-$1$ error on the validation dataset. The numbers reported in the submission are indeed averaged over 3 trials. Empirically we found that all the optimizers work stably with similar small standard deviations. We therefore did not report those numbers. Specifically, the top-1 and top-5 error of our result is 3.2\% and 7.2\% lower than the second best adaptive solver, RMSProp, at the $20$-th epoch, respectively, as listed in Table~\ref{table:imagenet}. The state-of-the-art top-1 error using AlexNet on ILSVRC2012 validation data is 42.6\%\footnote{\url{http://www.vlfeat.org/matconvnet/pretrained/}}, while our solver achieves {\bf 42.2\%} top-$1$ error in $50$ epochs, which is 0.4 percent lower than the state-of-the-art performance.

All the above experiments demonstrate the capability of the proposed solver BPGrad in training deep models for large-scale object recognition.

\begin{table}[t]
	\caption{Recognition error ($\%$) rate on ImageNet ILSVRC2012 dataset.}
	\footnotesize
	\begin{center}    
		\begin{tabular}{|c||c|c|c|c|}
			\hline 
			& \multicolumn{2}{|c|}{Top-1 error} & \multicolumn{2}{|c|}{Top-5 error} \\ \hline
			& \shortstack{Training} & \shortstack{Validation} & \shortstack{Training} & \shortstack{Validation} \\ \hline
			\hline
			Adagrad  & 49.0 & 54.8 & 25.5 & 30.2 \\ \hline
			Adadelta & 71.6 & 76.7 & 47.5 & 54.6 \\ \hline
			RMSProp  & 46.0 & 47.2 & 22.7 & 27.5 \\ \hline
			Adam     & 70.0 & 72.8 & 45.1 & 48.2 \\ \hline
			SGD      & \textbf{28.6} & \textbf{42.1} & \textbf{10.5} & \textbf{19.8}\\ \hline
			BPGrad   & \underline{33.0} & \underline{44.0} & \underline{13.2} & \underline{20.3}\\ \hline
		\end{tabular}
	\end{center}	
	\label{table:imagenet}
\end{table} 

\subsection{Object Detection}\label{sub:detection} 
Using the framework and source code of Fast RCNN~\cite{girshick2015fast}, we compare different solvers on the PASCAL VOC2007 dataset~\cite{pascal-voc-2007} with $20$ object classes. The default object proposal approach is selective search~\cite{uijlings2013selective}. For all solvers, we train the network for $12$ epochs using the $5$K images in VOC2007 trainval set and test it using $4.9$K images in VOC2007 test set. We set the weight decay and momentum to $0.0005$ and $0.9$, respectively, and use the default learning rates for the competitors. For our solver, we set $L=100$. 
\setlength{\tabcolsep}{0.85pt}
\begin{table*}[t]
	\renewcommand{\arraystretch}{0.85}	
	\caption{Average precision (AP, \%) of object detection on VOC2007 test dataset.}
	\vspace{-5mm}
	\footnotesize
	\begin{center}
		\begin{tabular}{|c||c|c|c|c|c|c|c|c|c|c|c|c|c|c|c|c|c|c|c|c|c|}
			\hline 
			& aero & bike & bird & boat & bottle & bus & car & cat & chair & cow & table & dog & horse & mbike & persn & plant & sheep & sofa & train & tv & mAP\\ \hline \hline
			Adagrad & 67.5 & 71.5 & 60.7 & 47.1 & 28.3 & 72.7 & 76.7 & 77.0 & 34.3 & 70.2 & 64.0 & 72.0 & 74.2 & 69.5 & 64.9 & 28.8 & 57.4 & 60.5 & 73.1 & 61.1  & 61.7    \\ \hline
			RMSProp & 69.1 & 75.8 & 61.5 & 47.9 & 30.2 & 74.7 & 77.1 & 79.4 & 33.2 & 71.1 & 66.3 & 74.4 & 76.3 & 69.9 & 65.1 & 28.9 & 62.9 & 62.5 & 73.2 & 60.8  & 63.0    \\ \hline
			Adam & 68.9 & \textbf{79.9} & 64.1 & \textbf{56.6} & 37.0 & \underline{77.4} & \underline{77.7} & \underline{82.5} & 38.2 & 71.5 & 64.7 & \underline{77.6} & 77.7 & \textbf{75.0} & \textbf{66.8} & \underline{30.6} & \textbf{65.9} & 65.1 & \underline{74.4} & \textbf{67.9}  & \underline{66.0} \\ \hline
			SGD & \textbf{72.0} & \underline{77.8} & \underline{65.7} & 50.9 & \textbf{40.0} & \textbf{78.1} & \textbf{78.2} & 80.5 & \textbf{41.3} & \underline{73.2} & \underline{66.8} & \textbf{78.5} & \underline{81.8} & \underline{73.6} & \textbf{66.8} & 29.5 & \underline{65.7} & \textbf{69.4} & \textbf{75.0} & 61.9 & \textbf{66.3} \\ \hline
			BPGrad & \underline{69.4} & 77.7 & \textbf{66.4} & \underline{55.1} & \underline{37.2} & 76.1 & \underline{77.7} & \textbf{83.6} & \underline{38.6} & \textbf{73.8} & \textbf{67.4} & 76.0 & \textbf{81.9} & 72.7 & \underline{66.3} & \textbf{31.0} & 64.2 & \underline{66.2} & 73.8 & \underline{64.9} & \underline{66.0} \\ \hline
		\end{tabular}
	\end{center}
	\label{table:detection_fast_rcnn}
\end{table*}

\begin{figure}[t]
	\begin{minipage}[b]{0.49\columnwidth}
		\begin{center} 
			\centerline{\includegraphics[width=\columnwidth,height=0.75\columnwidth]{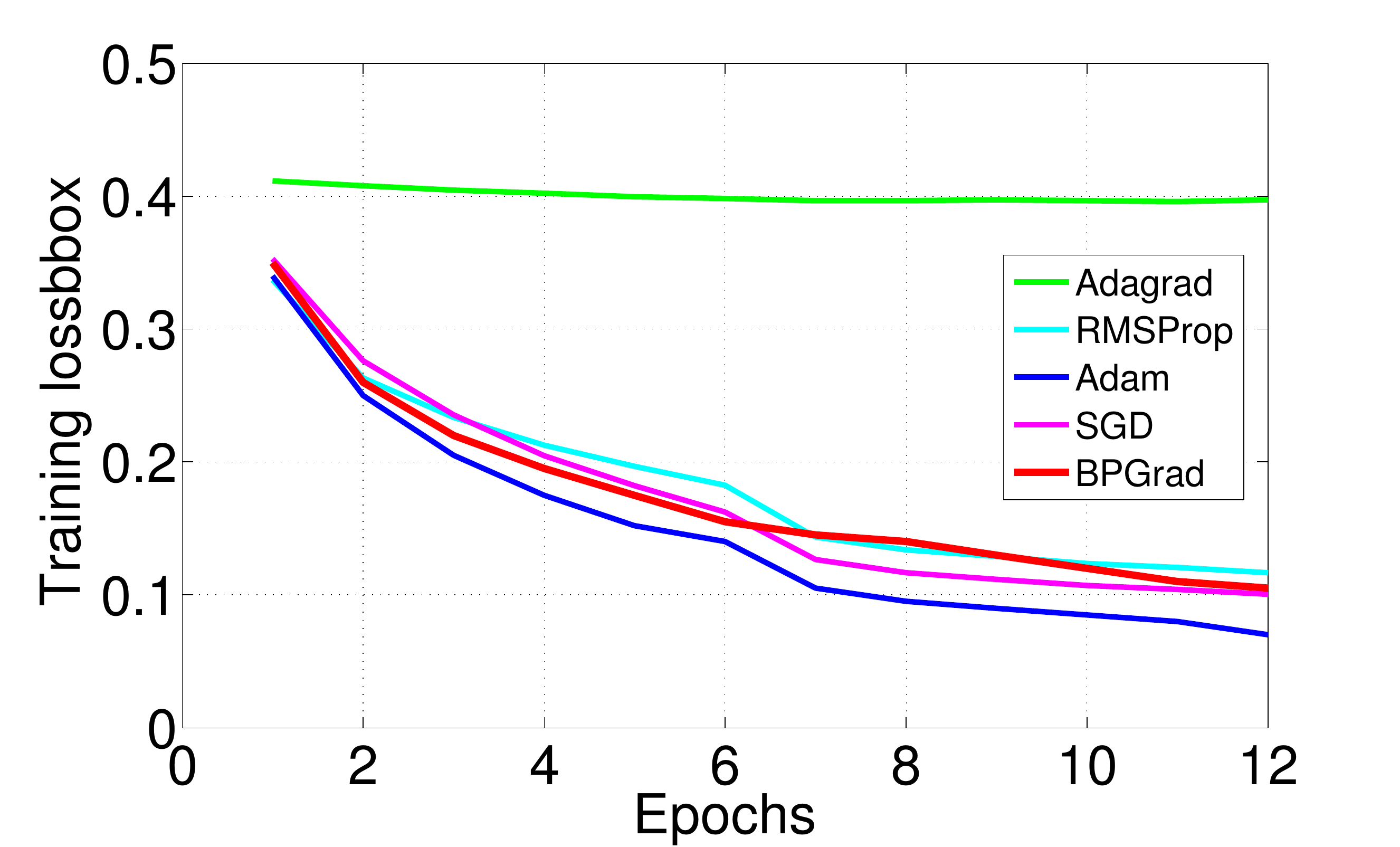}}
		\end{center}
	\end{minipage}	
	\begin{minipage}[b]{0.49\columnwidth}
		\begin{center} 	
		\centerline{\includegraphics[width=\columnwidth,height=0.75\columnwidth]{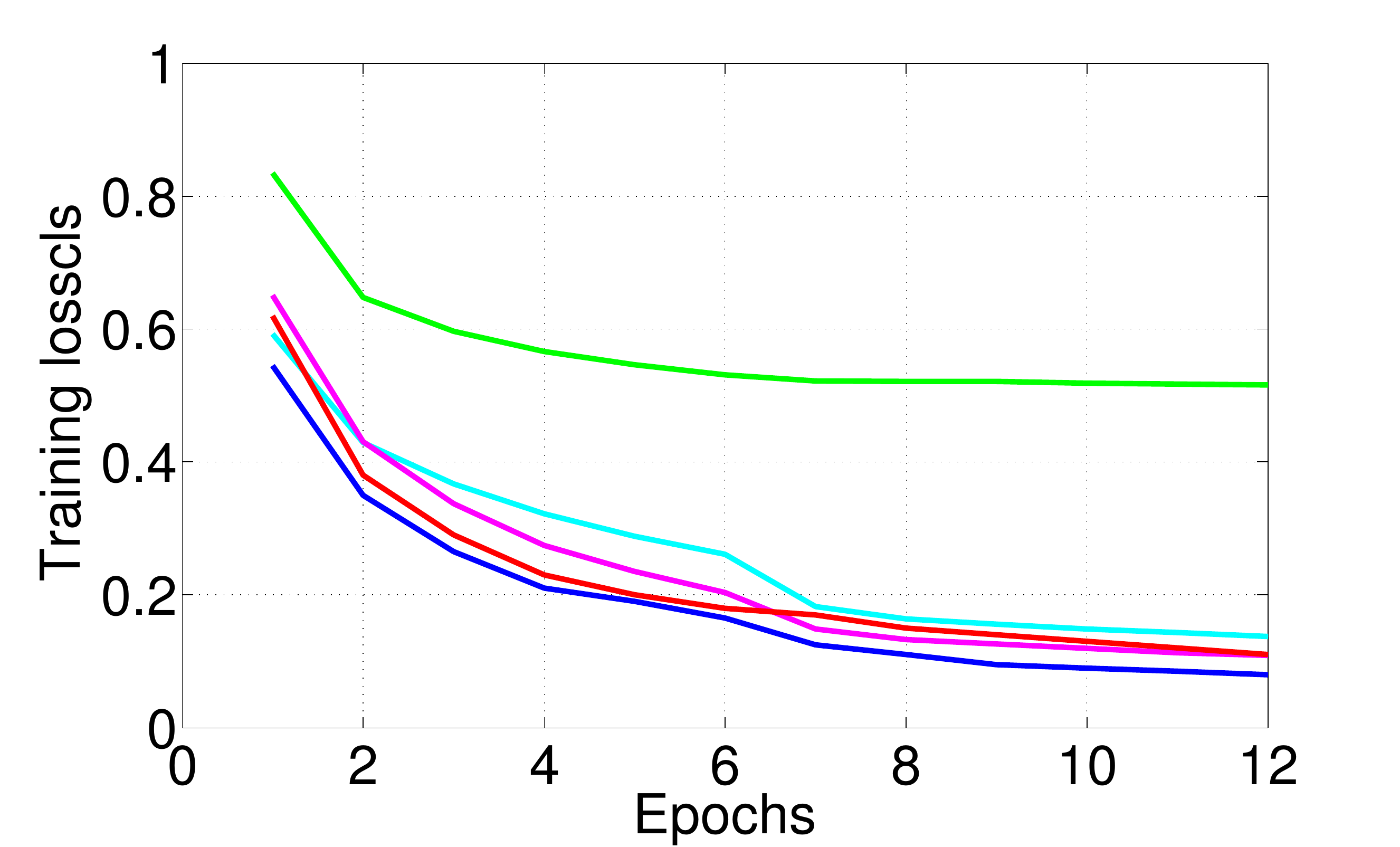}}		
		\end{center}
	\end{minipage} 
	\vspace{-8mm}
	\caption{Loss comparison on VOC2007 trainval dataset for object detection, including {\bf (left)} the regression loss using bounding boxes and {\bf (right)} the classification loss.} 
	\label{fig:fast_rcnn_loss}
\end{figure}
The training loss and average precision at the test time are shown in Fig.~\ref{fig:fast_rcnn_loss} and Table~\ref{table:detection_fast_rcnn}, respectively. Ours achieves slightly higher values in terms of training loss than Adam, however, our solver achieves on par performance with Adam, and SGD at test time on average.

\subsection{Object Segmentation}\label{sub:segmentation} 
Following the work~\cite{long2015fully} for semantic segmentation based on fully convolutional networks (FCN), we train FCN-32s with per-pixel multinomial logistic loss and validate it with the standard metric of mean pixel intersection over union (IU), pixel accuracy, and mean accuracy. For all the solvers, we conduct training for $50$ epochs with momentum $0.9$ and weight decay $0.0005$ on PASCAL VOC2011~\cite{pascal-voc-2011} segmentation set. For Adagrad, RMSProp, Adam, and SGD, we find that the default parameters are able to achieve the best performance. For Adadelta, we tune its parameters with $\epsilon=10^{-9}$. The global learning rate for RMSProp is set to $10^{-5}$ and $10^{-4}$ for Adagrad and Adam, respectively. Adadelta does not require a global learning rate. For our solver, we set $L=500$.   
\begin{figure}[t]
	\begin{minipage}[b]{0.49\columnwidth}
		\begin{center} 
			\centerline{\includegraphics[width=\columnwidth,height=0.75\columnwidth]{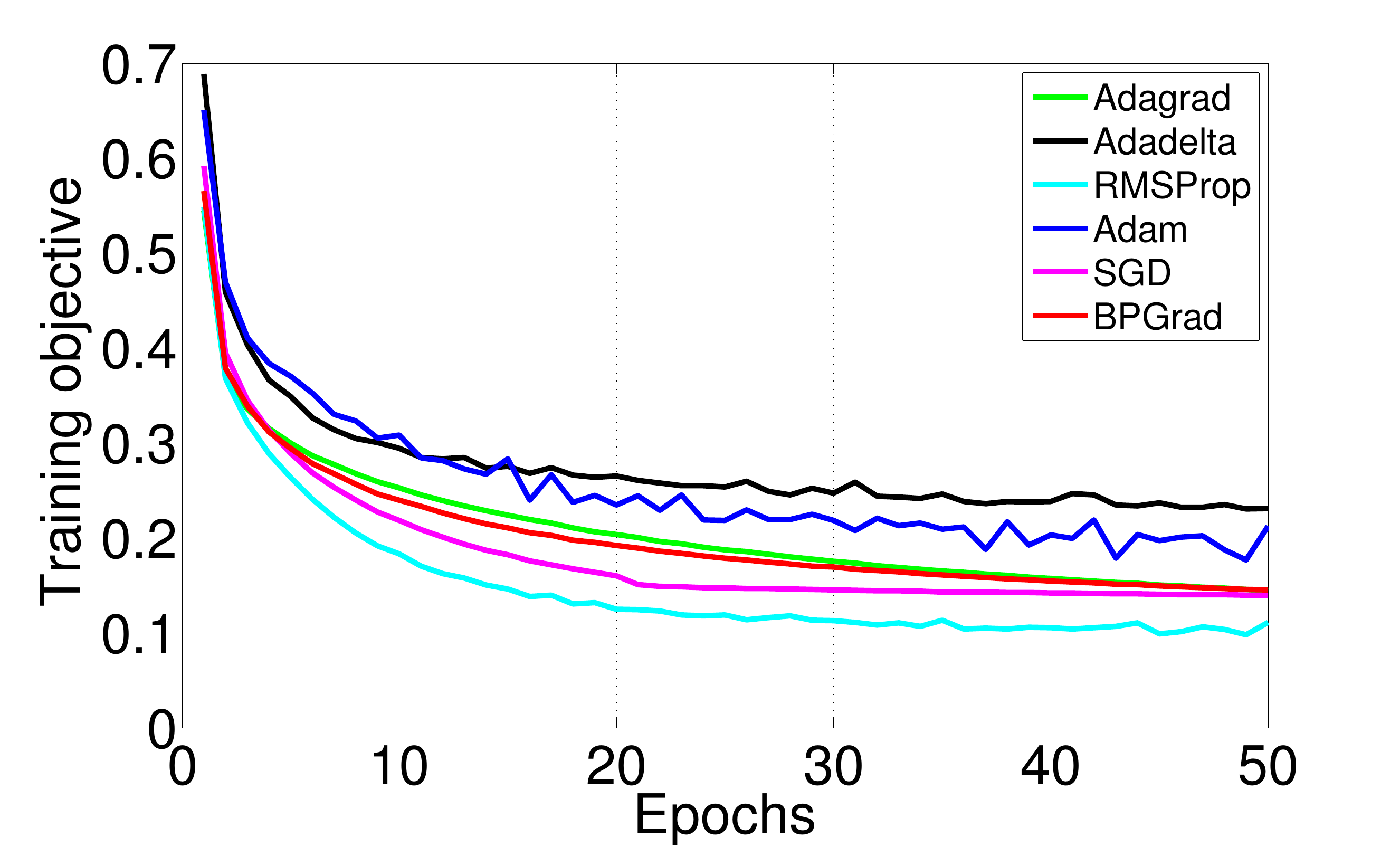}}
		\end{center}
	\end{minipage}	
	\begin{minipage}[b]{0.49\columnwidth}
		\begin{center} 		
			\centerline{\includegraphics[width=\columnwidth,height=0.75\columnwidth]{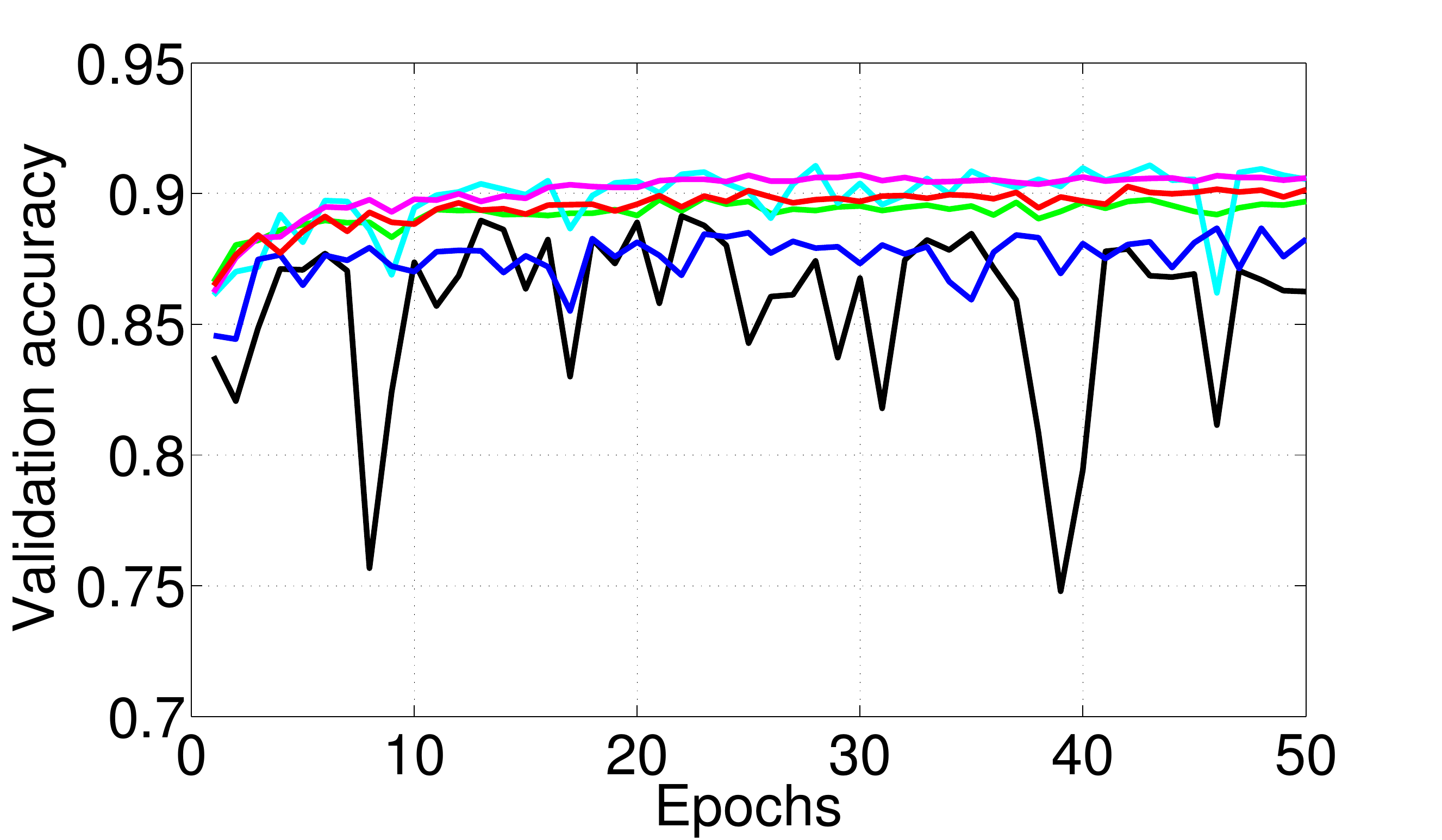}}	
		\end{center}
	\end{minipage}	
	\vspace{-8mm} 
	\caption{Segmentation performance comparison using FCN-32s model on VOC2011 training and validation datasets.}
	\label{fig:fcn_segmentation}
\end{figure}

\setlength{\tabcolsep}{1.8pt}
\begin{table}[t]\small
	\caption{Numerical comparison on semantic segmentation performance using VOC2011 test dataset at the $50$-th epoch in the form of mean$\pm$std using 5 runs (best in bold, 2nd best with underline).}
	\footnotesize
	\begin{center}	
		\begin{tabular}{|c||c|c|c|c|}
			\hline
			& Mean IU    & Pixel Accuracy   & Mean Accuracy & Average \\ \hline \hline
			Adagrad  &0.608$\pm$0.0003      &0.895$\pm$0.0003      &0.773$\pm$0.0011  &0.759 \\ \hline			            
			Adadelta  & 0.456$\pm$0.0703       &0.854$\pm$0.0345       &0.567$\pm$0.0319  & 0.626\\ \hline
			RMSProp  & 0.586$\pm$0.0096      &$\underline{0.899}$$\pm$0.0032      &0.677$\pm$0.0221 & 0.721 \\ \hline
			Adam 	 & 0.499$\pm$0.0262      &0.872$\pm$0.0086      &0.631$\pm$0.0336 & 0.667 \\ \hline		            
			SGD	  &\textbf{0.633}$\pm$0.0010      &\textbf{0.904}$\pm$0.0004      &$\underline{0.786}$$\pm$0.0010   & \textbf{0.774} \\ \hline
			BPGrad    & $\underline{0.623}$$\pm$0.0007 &    $\underline{0.899}$$\pm$0.0003     & \textbf{0.793}$\pm$0.0008	  & $\underline{0.772}$ \\  \hline
		\end{tabular}
	\end{center}		
	\label{table:fcn_segmentation} 
\end{table}

The learning curves on the training and validation datasets are shown in Fig.~\ref{fig:fcn_segmentation}. In addition, the test-time comparison results are shown in Table~\ref{table:fcn_segmentation}. In this experiment, our solver has very similar learning behavior as Adagrad, however, it achieves better performance at test time. The results demonstrate that our solver has the ability to learn robust and deep models for object segmentation. We can also observe from Fig.~\ref{fig:fcn_segmentation} that our solver is reliable as it exhibits smaller fluctuation over epochs in comparison with the competitors. The smaller fluctuation over epochs on the validation dataset demonstrates again the superior reliability of our solver, compared with the competitors. Taking these observations into account, we believe that our solver has the ability to learn robust and deep models for object segmentation.

In summary, we can observe that our BPGrad solver can not only achieve on par performance with SGD, but also eliminating the manual tuning of learning rate. 

\section{Conclusion}\label{sec:conclusion}
In this paper, we have proposed a novel approximation algorithm, {\em BPGrad}, towards searching for the global optimality via branch and pruning based on the Lipschitz continuity assumption. We have theoretically proved that under some conditions our BPGrad algorithm can converge to the global optimality within finite iterations. Empirically, a computationally efficient solver based on BPGrad has been proposed to train deep neural networks. We have demonstrated the superiority of our BPGrad solver to several adaptive deep learning solvers for vision applications of object recognition, detection, and segmentation.

\section*{Acknowledgment}
Dr. Wu and Dr. Wang were supported in part by NSF and USDA under the award no. 2019-67021-28996. This work was done when Dr. Zhang was a researcher at Mitsubishi Electric Research Laboratories (MERL), USA.

\footnotesize 
\bibliography{BPGrad_reference}

\end{document}


\begin{frontmatter}

\title{Branch-and-Pruning Optimization Towards Global Optimality in Deep Learning - Supplementary Document}

\author{Yuanwei Wu, Ziming Zhang, Guanghui Wang}

\end{frontmatter}


\section{Parameter Setting in the Experiments}\label{sec:Appendix_experimental_details}
\subsection{Parameters tuning on NNIST and CIFAR-10} \label{app:params_tuning_mnist_cifar10}
\begin{figure}[t]
	\begin{center}
		\includegraphics[width=.48\linewidth]{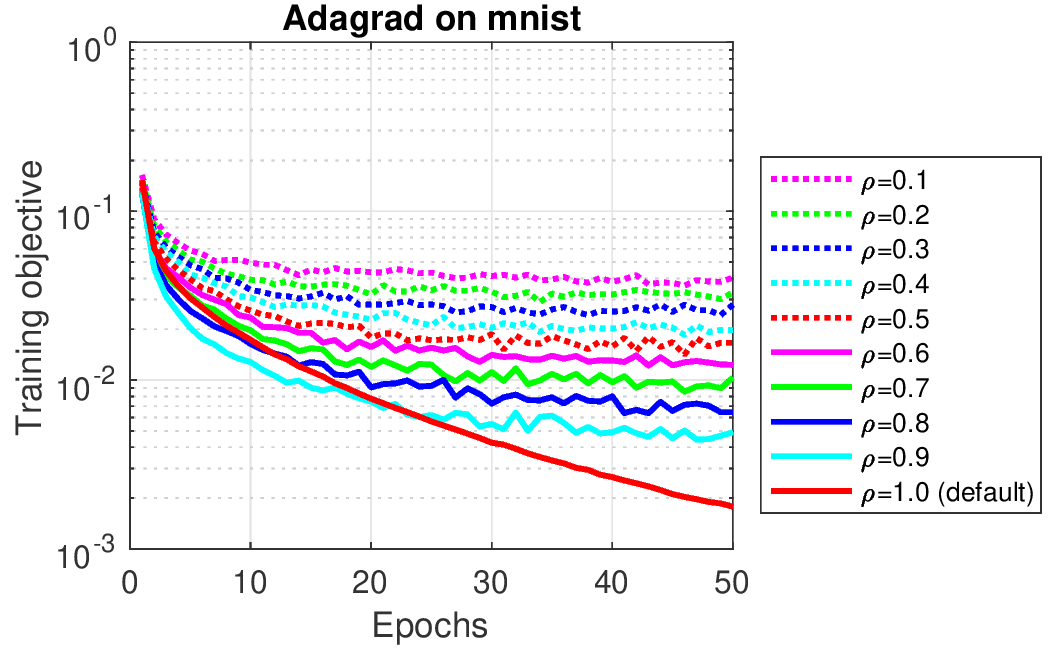}
		\includegraphics[width=.37\linewidth]{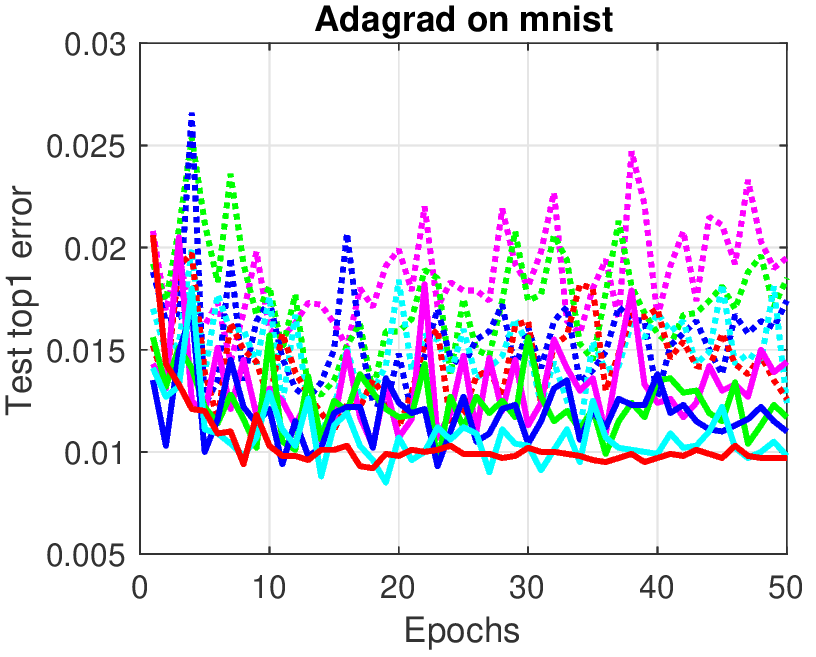}		
		\includegraphics[width=.48\linewidth]{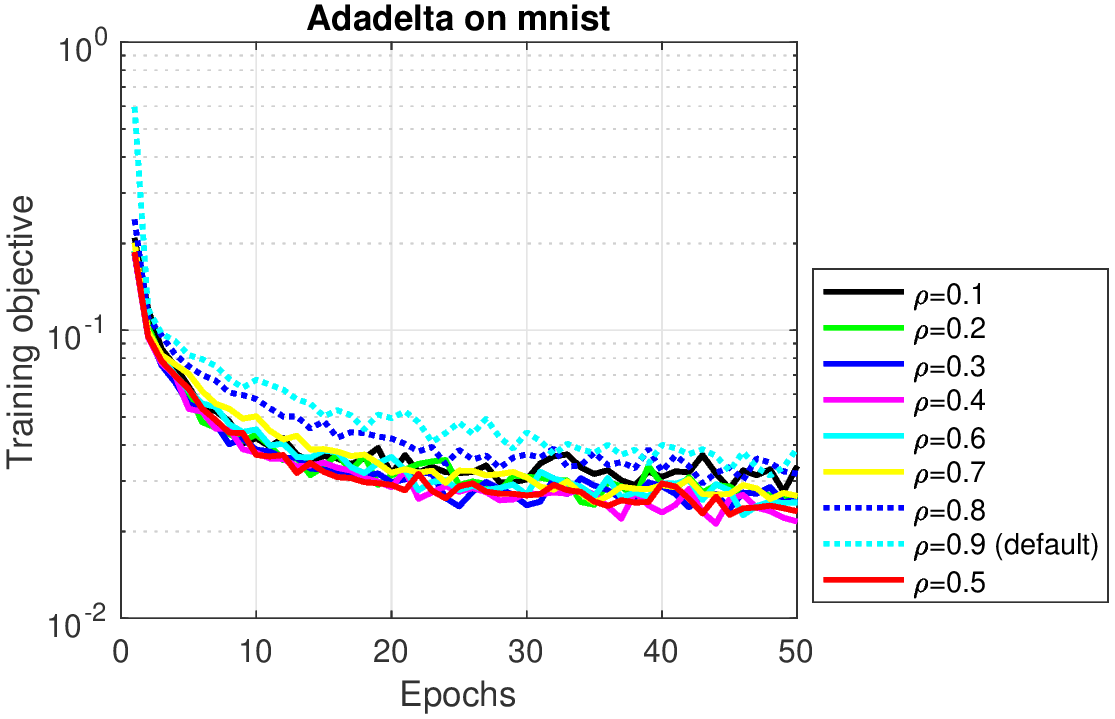}
		\includegraphics[width=.37\linewidth]{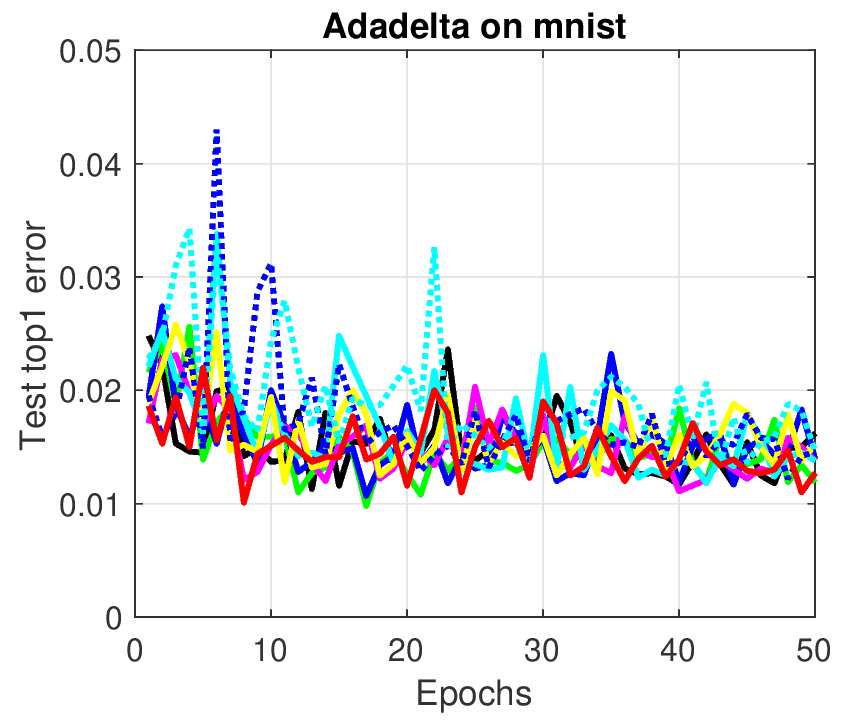}
		\includegraphics[width=.48\linewidth]{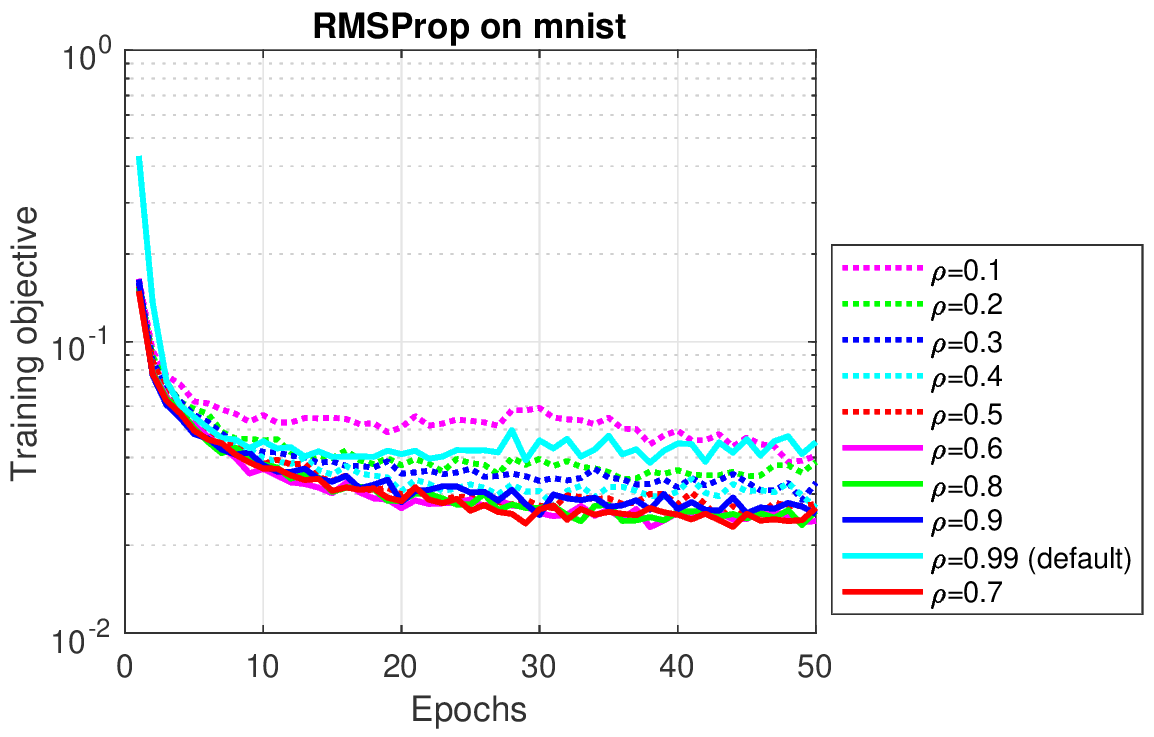}
		\includegraphics[width=.37\linewidth]{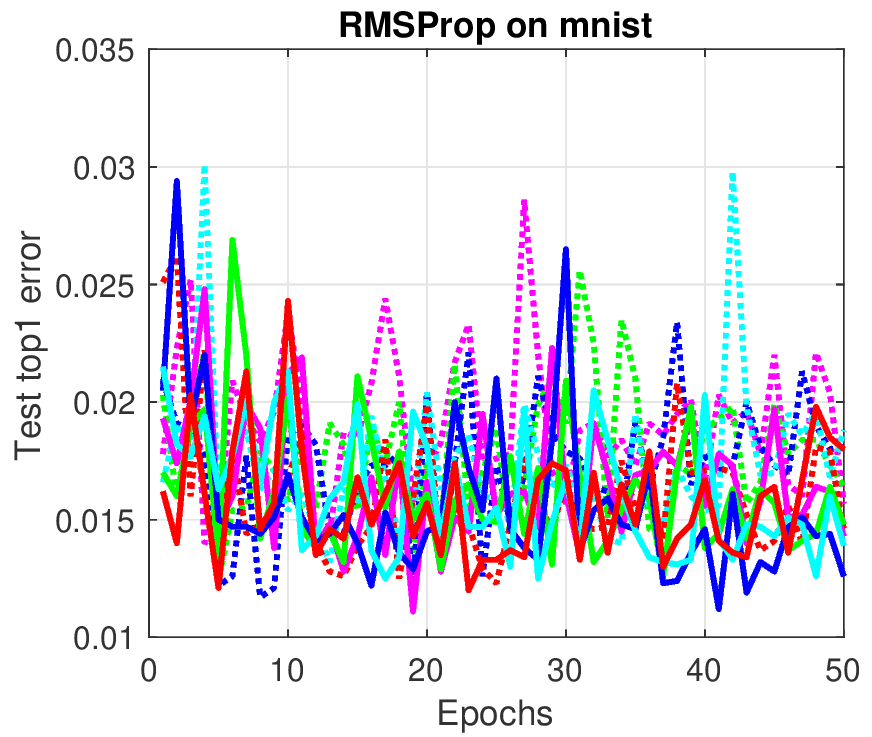}
		\includegraphics[width=.51\linewidth]{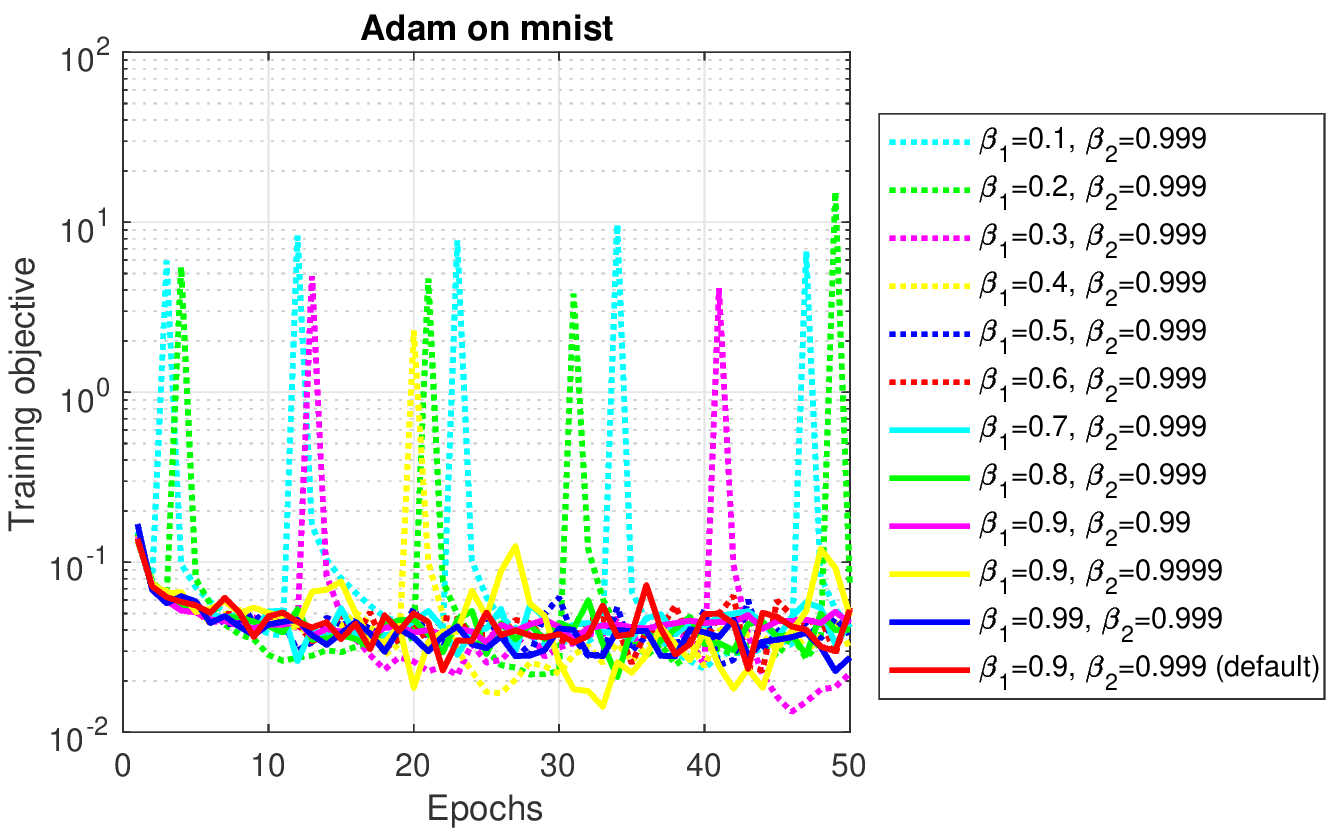}
		\includegraphics[width=.34\linewidth]{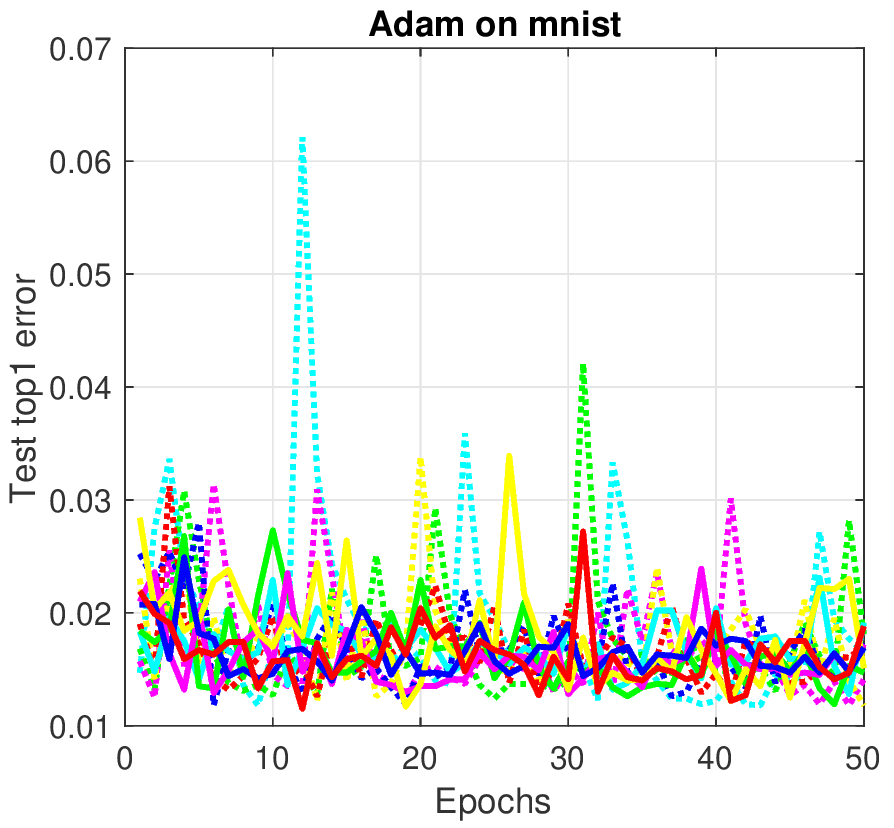}
		\caption{\footnotesize Parameters search on MNIST {\bf (left)} training objectives and {\bf (right)} test top-1 errors for different solvers.}
		\label{fig:params_search_Adam_mnist_train_obj}
	\end{center}
\end{figure}

\begin{figure}[t]
	\begin{center}
		\includegraphics[width=.48\linewidth]{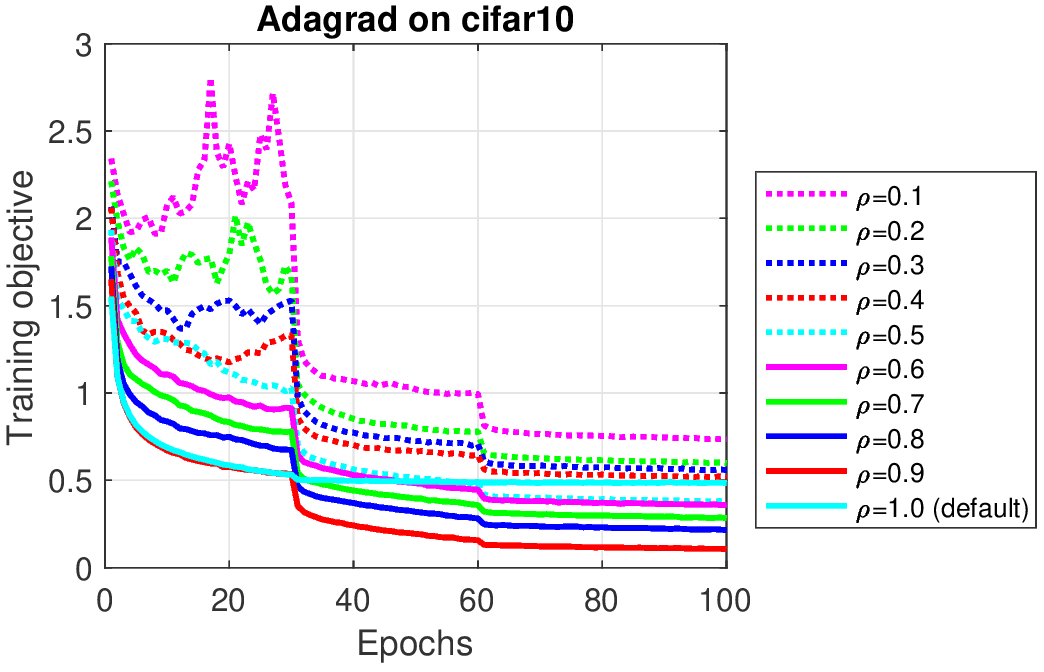}
		\includegraphics[width=.35\linewidth]{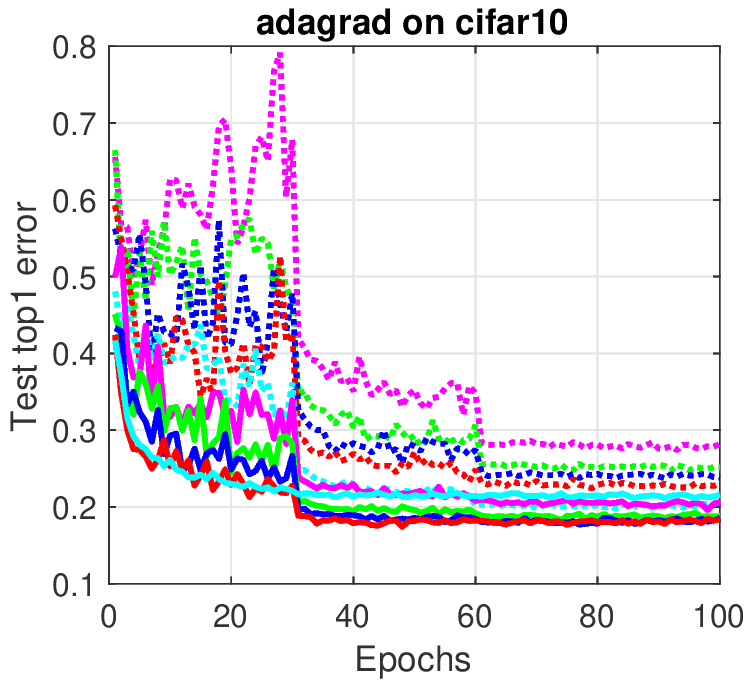}
		\includegraphics[width=.48\linewidth]{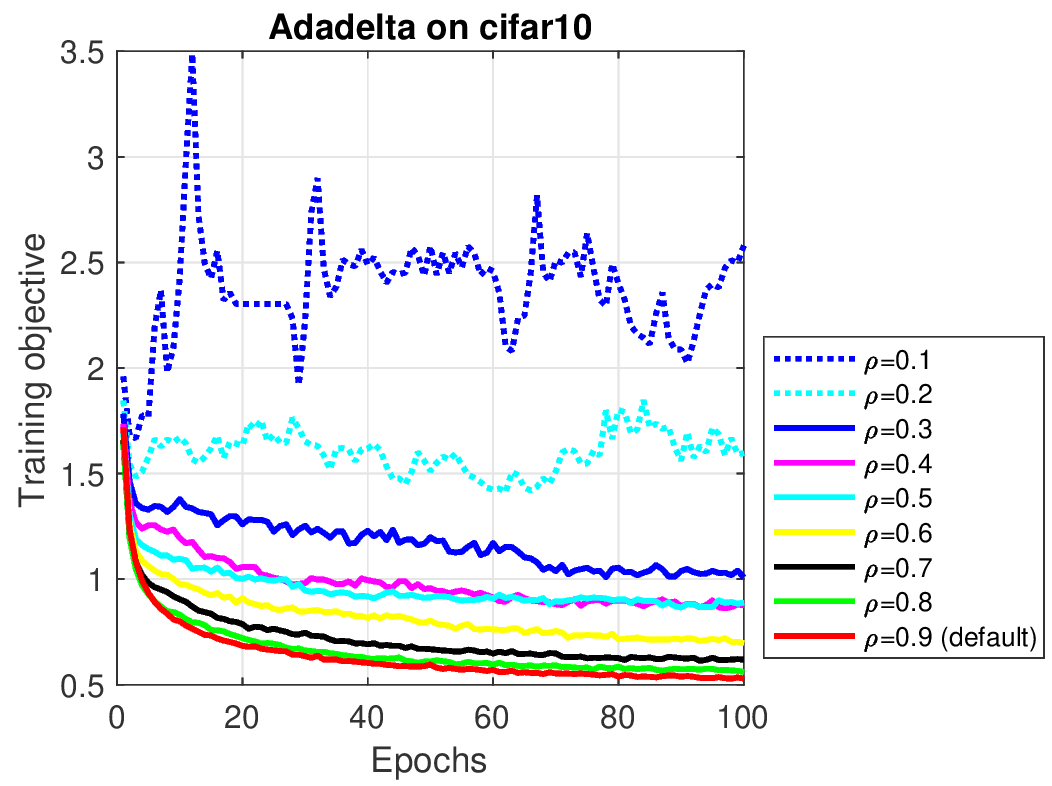}
		\includegraphics[width=.37\linewidth]{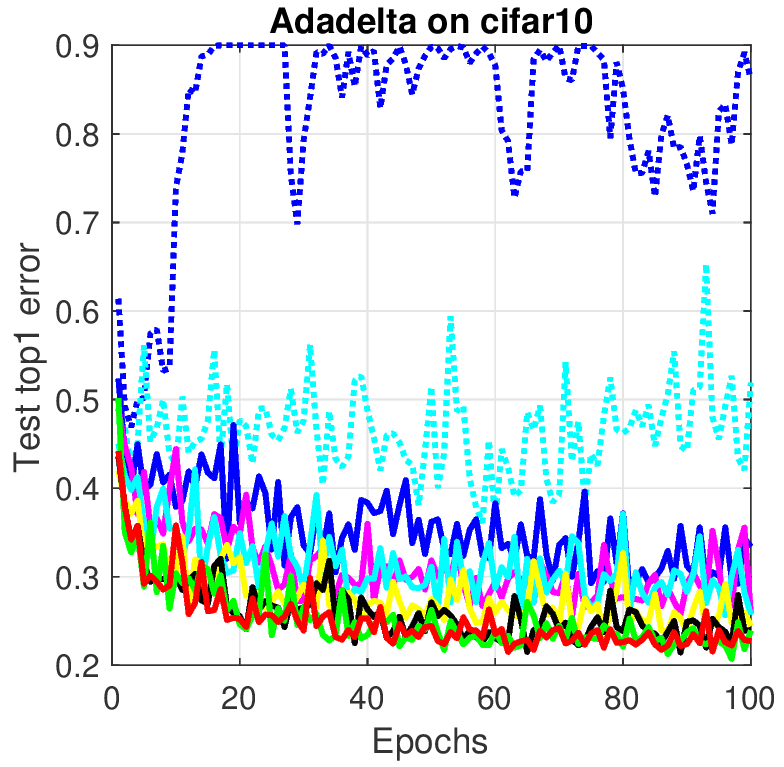}
		\includegraphics[width=.48\linewidth]{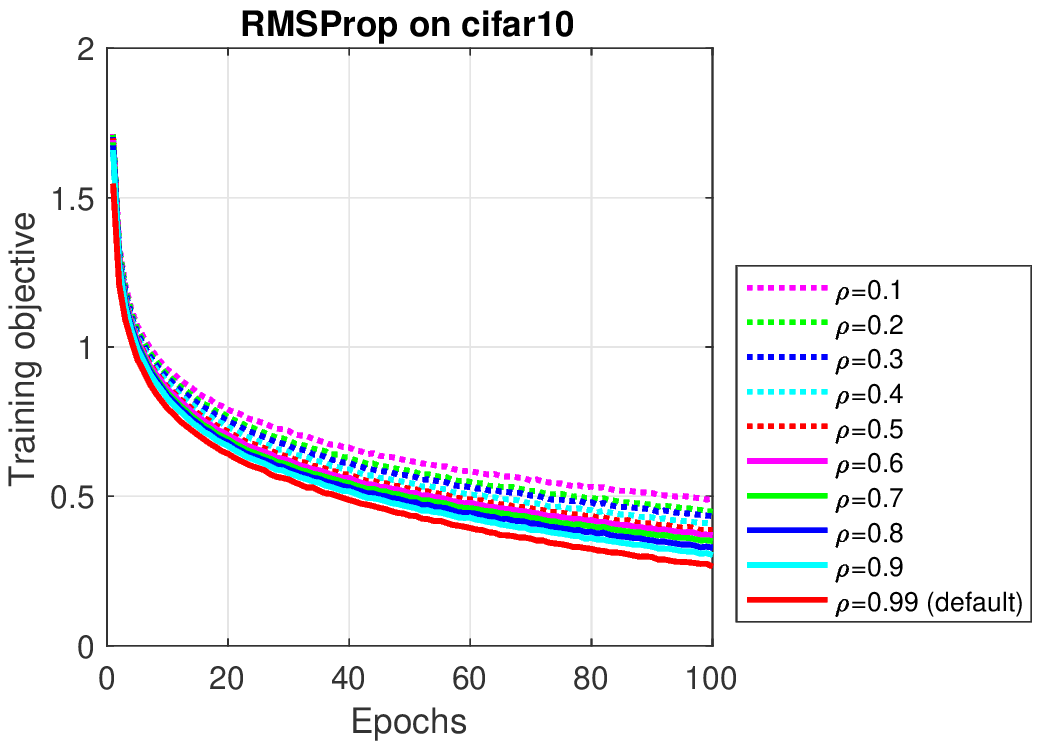}
		\includegraphics[width=.37\linewidth]{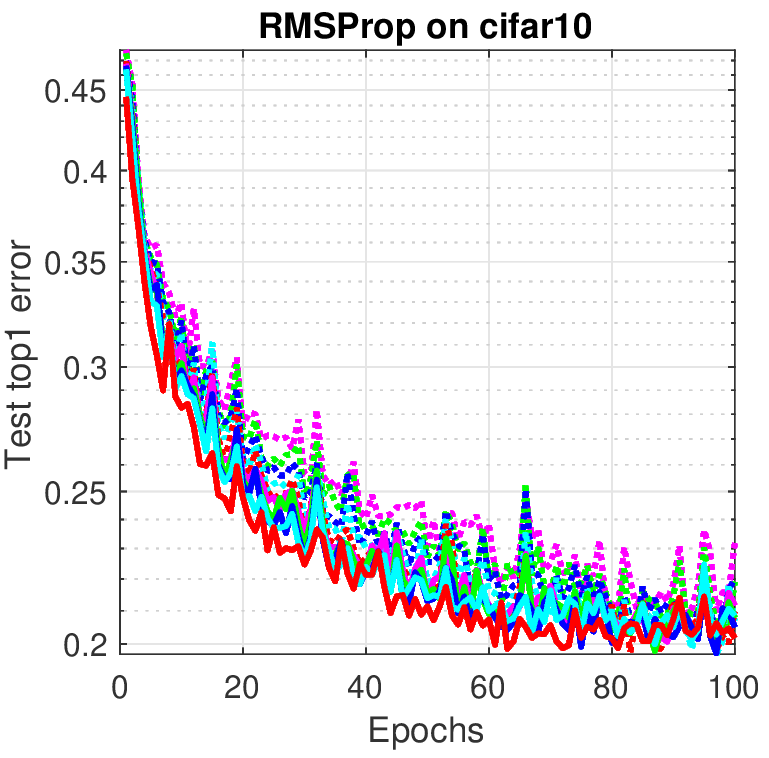}
		\includegraphics[width=.50\linewidth]{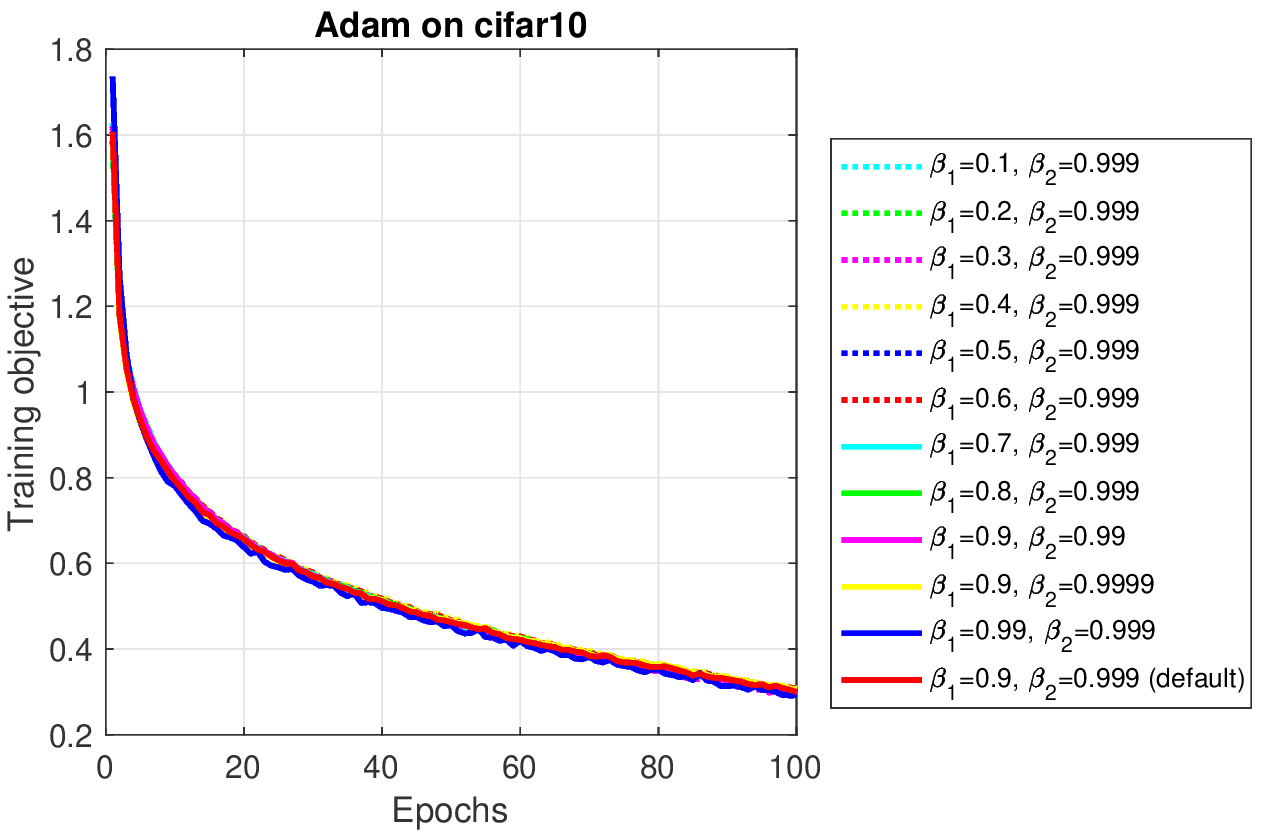}
		\includegraphics[width=.35\linewidth]{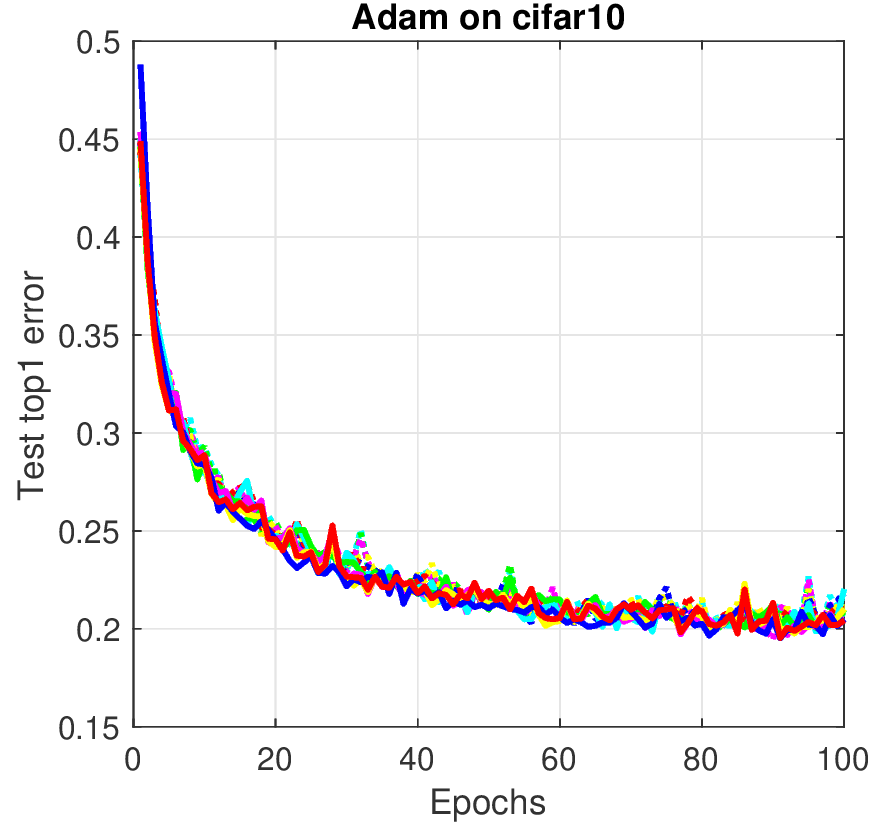}
		\caption{\footnotesize Parameters search on CIFAR-10 {\bf (left)} training objectives and {\bf (right)} test top-1 errors for different solvers.}
		
		\label{fig:params_search_Adam_cifar10_train_obj}
	\end{center}
\end{figure}

We perform grid search of parameters on MNIST and CIFAR-10 for Adagrad, Adadelta, RMSProp and Adam, as shown in Fig.~\ref{fig:params_search_Adam_mnist_train_obj}, and Fig.~\ref{fig:params_search_Adam_cifar10_train_obj}, respectively. We summarize the parameters as follows.
\begin{itemize}
	\item Adagrad: $\epsilon=10^{-10}$, \\$\rho=[0.1, 0.2, 0.3, 0.4, 0.5, 0.6, 0.7, 0.8, 0.9, 1.0~ (\text{default})]$.
	\item Adadelta: $\epsilon=10^{-6}$, \\$\rho=[0.1, 0.2, 0.3, 0.4, 0.5, 0.6, 0.7, 0.8, 0.9 ~(\text{default})]$.
	\item RMSProp: $\epsilon=1e^{-8}$, \\$\rho=[0.1, 0.2, 0.3, 0.4, 0.5, 0.6, 0.7, 0.8, 0.9, 0.99 ~(\text{default})]$.
	\item Adam: $\epsilon=10^{-8}$, \\$\beta_1=[0.1, 0.2, 0.3, 0.4, 0.5, 0.6, 0.7, 0.8, 0.9~(\text{default}), 0.99]$, \\$\beta_2=[0.99, 0.999~(\text{default}), 0.9999]$.
\end{itemize}

\subsection{Parameters tuning for Adam}\label{app:params_tuning_adam}
We perform experiments using Adam with $\beta_1=0.9, \beta_2=0.999~(\text{default})$ and $\beta_1=0.99, \beta_2=0.999$ on ImageNet for recognition and on VOC2011 for segmentation.

\begin{figure}[t]			
	\begin{minipage}[b]{0.49\columnwidth}
		\begin{center}
			\centerline{\includegraphics[width=\columnwidth,height=0.75\columnwidth]{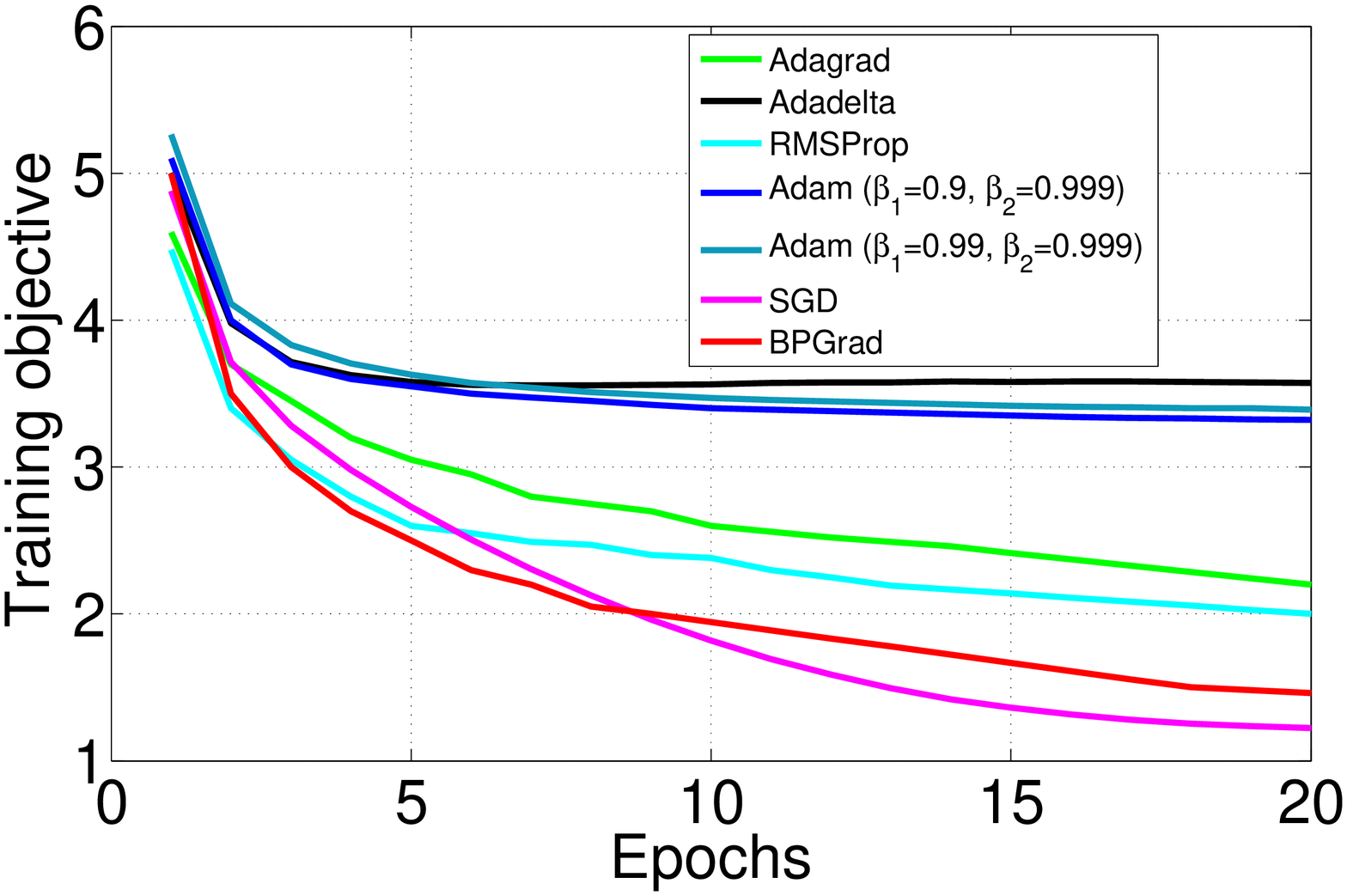}}
		\end{center}
	\end{minipage}	
	\begin{minipage}[b]{0.49\columnwidth}
		\begin{center}
			\centerline{\includegraphics[width=\columnwidth,height=0.75\columnwidth]{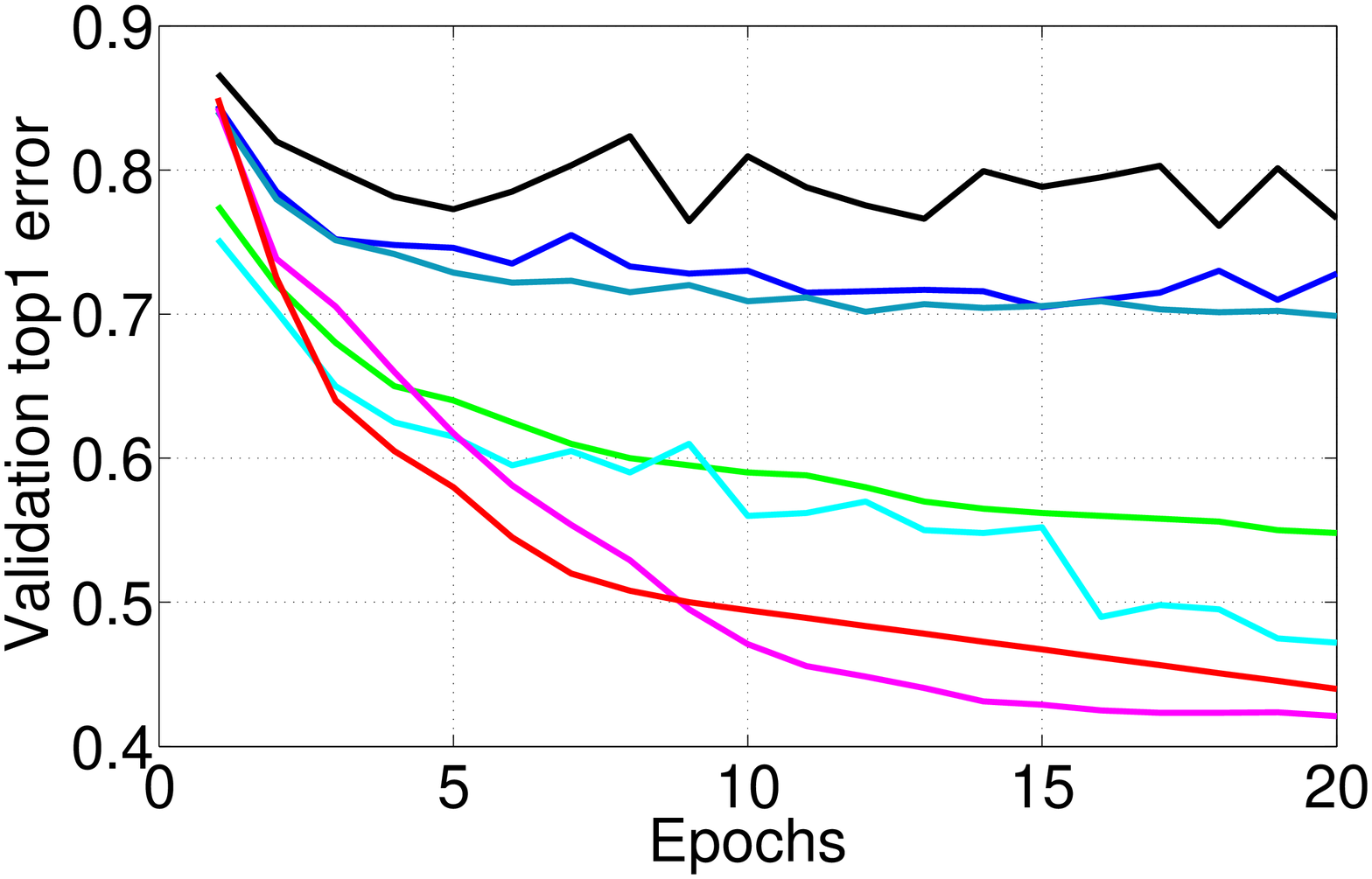}}		
		\end{center}
	\end{minipage}	
	\vspace{-10mm}
	\caption{Comparison on {\bf (left)} training objectives and {\bf (right)} validation top-1 errors for object recognition using ImageNet ILSVRC2012.}\label{fig:imagenet_2_adams}
\end{figure}

\begin{figure}[t]
	\begin{minipage}[b]{0.49\columnwidth}
		\begin{center}
			\centerline{\includegraphics[width=\columnwidth,height=0.75\columnwidth]{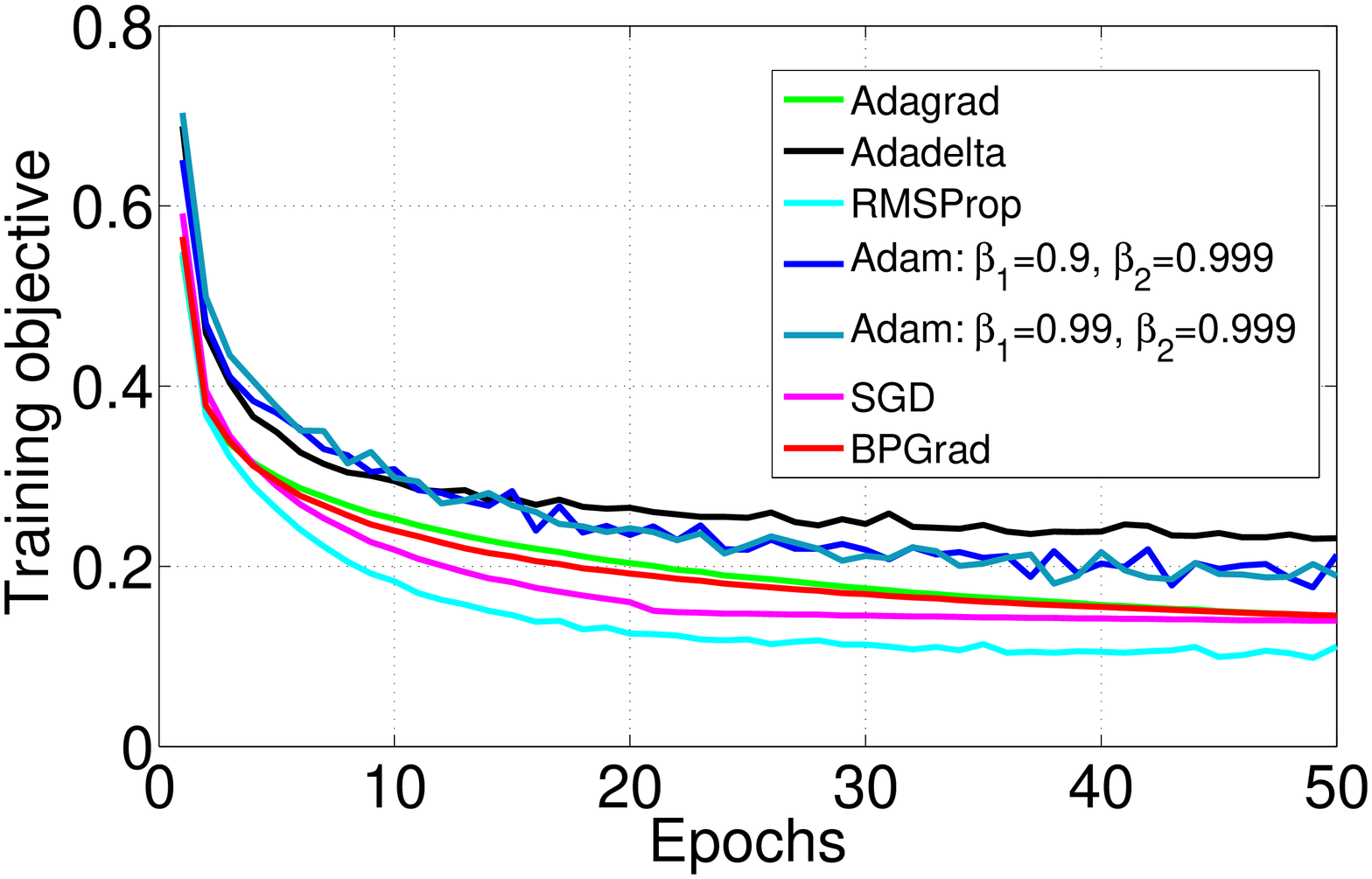}}
		\end{center}
	\end{minipage}	
	\begin{minipage}[b]{0.49\columnwidth}
		\begin{center}
			\centerline{\includegraphics[width=\columnwidth,height=0.75\columnwidth]{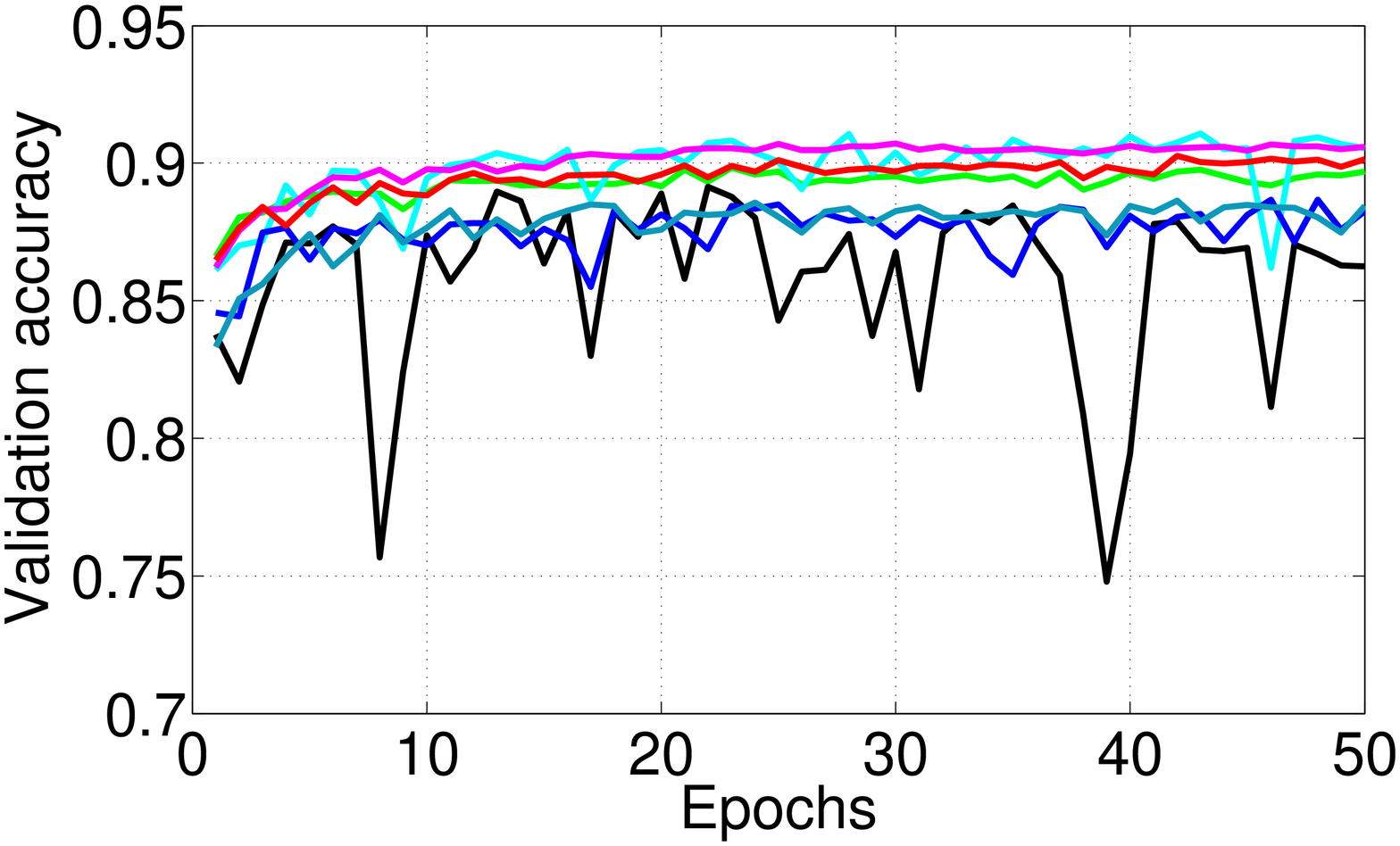}}			
		\end{center}
	\end{minipage}	
	\vspace{-10mm}
	\caption{\footnotesize Segmentation performance comparison using FCN-32s model on VOC2011 training and validation datasets.}\label{fig:fcn_segmentation_2_adams}
\end{figure}

Fig.~\ref{fig:imagenet_2_adams} shows the comparison results on ImageNet. We can observe there is little difference between the two parameter settings. It exhibits similar trend for segmentation as shown Fig.~\ref{fig:fcn_segmentation_2_adams}.

\subsection{Parameters of solvers on 1-D functions}\label{app:params_1d_functions}
\begin{enumerate}
	\item For function $f_1=x\sin(x)+4.815, x\in[0, 8]$:
	\begin{itemize}
		\item Adagrad: $\epsilon=10^{-10}, \rho=0.9, \text{learning rate (lr)=0.5}$,
		\item Adadelta: $\epsilon=10^{-6}, \rho=0.99$,
		\item RMSProp: $\epsilon=10^{-6}, \rho=0.99, \text{lr}=0.05$,
		\item Adam: $\epsilon=10^{-6}, \beta_1=0.9, \beta_2=0.99, \text{lr}=0.1$,
		\item SGD: $\mu=0, \text{lr}=0.1$,
		\item SGD: $\mu=0.9, \text{lr}=0.008$,
		\item BPGrad Alg.~1: $\rho=0.1, L=4\pi$,
		\item BPGrad: $\rho=0.1, L=4\pi, \mu=0$,
		\item BPGrad: $\rho=0.1, L=4\pi, \mu=0.9$.
	\end{itemize}	
	\item For functions $f_2=x\sin(x)+11.05, x\in[0, 4\pi]$ and $f_3=x\sin(x)+15, x\in[0, 4\pi]$:
	\begin{itemize}
		\item Adagrad: $\epsilon=10^{-10}, \rho=0.9, \text{learning rate (lr)=0.5}$,
		\item Adadelta: $\epsilon=10^{-6}, \rho=0.99$,
		\item RMSProp: $\epsilon=10^{-6}, \rho=0.99, \text{lr}=0.05$,
		\item Adam: $\epsilon=10^{-6}, \beta_1=0.9, \beta_2=0.99, \text{lr}=0.1$,
		\item SGD: $\mu=0, \text{lr}=0.1$,
		\item SGD: $\mu=0.9, \text{lr}=0.1$,
		\item BPGrad Alg.~1: $\rho=0.1, L=4\pi$,
		\item BPGrad: $\rho=0.1, L=4\pi, \mu=0$, L is multiplied by 2 every $10$ iterations (similar to regularization using weight decay).
		\item BPGrad: $\rho=0.1, L=4\pi, \mu=0.9$, L is multiplied by 2 every $10$ iterations.
	\end{itemize}	
\end{enumerate}
\subsection{Parameters of solvers on MNIST using LeNet-5}\label{app:params_mnist_lenet}
All the solvers use $\text{weight decay}=0.0005$ and $\text{momentum}=0.9$, unless stated otherwise.
\begin{itemize}
	\item Adagrad: $\epsilon=10^{-10}, \rho=1.0~\text{(default)}$,
	\item Adadelta: $\epsilon=10^{-6}, \rho=0.5$,
	\item RMSProp: $\epsilon=10^{-8}, \rho=0.7$,
	\item Adam: $\epsilon=10^{-8}, \beta_1=0.9, \beta_2=0.999~\text{(default)}$,
	\item SGD: $\text{lr}=[0.001*\text{ones(1, 100)}]$,
	\item BPGrad: $\rho=0.1, L=15, \mu=0$,
	\item BPGrad: $\rho=0.1, L=15, \mu=0.9$.
\end{itemize}

\subsection{Parameters of solvers on CIFAR-10 using LeNet}\label{app:params_cifar10_lenet}
All the solvers use $\text{weight decay}=0.0005$ and $\text{momentum}=0.9$, unless stated otherwise.
\begin{itemize}
	\item Adagrad: $\epsilon=10^{-10}, \rho=0.9$,
	\item Adadelta: $\epsilon=10^{-6}, \rho=0.9~\text{(default)}$,
	\item RMSProp: $\epsilon=10^{-8}, \rho=0.99~\text{(default)}$,
	\item Adam: $\epsilon=10^{-8}, \beta_1=0.9, \beta_2=0.999~\text{(default)}$,
	\item Eve: $\text{lr}=[0.01*\text{ones(1, 30)}, 0.005*\text{ones(1, 30)}, 0.0001*\text{ones(1, 40)}]$,
	\item SGD: $\text{lr}=[0.05*\text{ones(1, 30)}, 0.005*\text{ones(1, 30)}, 0.0005*\text{ones(1, 40)}]$,
	\item BPGrad: $\rho=0.1, L=50, \mu=0$,
	\item BPGrad: $\rho=0.1, L=50, \mu=0.9$.
\end{itemize}

\subsection{Parameters of solvers on ImageNet}\label{app:params_Imagenet}
All the solvers use $\text{weight decay}=0.0005$ and $\text{momentum}=0.9$, unless stated otherwise.
\begin{itemize}
	\item Adagrad: $\epsilon=10^{-10}, \rho=1.0~\text{(default)}$,
	\item Adadelta: $\epsilon=10^{-6}, \rho=0.9~\text{(default)}$,
	\item RMSProp: $\epsilon=10^{-8}, \rho=0.99~\text{(default)}$,
	\item Adam: $\epsilon=10^{-8}, \beta_1=0.9, \beta_2=0.999~\text{(default)}$,
	\item SGD: $\text{lr}=[\log(-1,-4, 20)]$,
	\item BPGrad: $\rho=0.1, L=100, \mu=0.9, N=12$.
\end{itemize} 

\subsection{Parameters of solvers for object detection}\label{app:params_object_detection}
\begin{itemize}
	\item Adagrad: $\epsilon=10^{-10}, \rho=1.0~\text{(default)}$,
	\item RMSProp: $\epsilon=10^{-8}, \rho=0.99~\text{(default)}$,
	\item Adam: $\epsilon=10^{-8}, \beta_1=0.9, \beta_2=0.999~\text{(default)}$,
	\item SGD: $\text{lr}=[0.001*\text{ones(1, 6)}, 0.0001*\text{ones(1, 6)}]$,
	\item BPGrad: $\rho=0.1, L=100, \mu=0.9$.
\end{itemize} 

\subsection{Parameters of solvers for object segmentation}\label{app:params_object_segmentation}
\begin{itemize}
	\item Adagrad: $\epsilon=10^{-10}, \rho=1.0~\text{(default)}$,
	\item Adadelta: $\epsilon=10^{-6}, \rho=0.9~\text{(default)}$,
	\item RMSProp: $\epsilon=10^{-8}, \rho=0.99~\text{(default)}$,
	\item Adam: $\epsilon=10^{-8}, \beta_1=0.9, \beta_2=0.999~\text{(default)}$,
	\item SGD: $\text{lr}=[0.001*\text{ones(1, 20)}, 0.0001*\text{ones(1, 30)}]$,
	\item BPGrad: $\rho=0.1, L=500, \mu=0.9$.
\end{itemize}
